\documentclass[conference]{IEEEtran}
\IEEEoverridecommandlockouts
\usepackage{cite}
\usepackage{amsmath,amssymb,amsfonts}
\usepackage{algorithmic}
\usepackage{graphicx}
\usepackage{textcomp}
\usepackage{xcolor}

\usepackage{multirow}
\usepackage{booktabs}
\usepackage{bm}

\usepackage{subfigure}

\usepackage{subcaption}
\usepackage{enumerate}
\usepackage{color}

\def\BibTeX{{\rm B\kern-.05em{\sc i\kern-.025em b}\kern-.08em
    T\kern-.1667em\lower.7ex\hbox{E}\kern-.125emX}}
\begin{document}

\title{Enhancing Human Motion Prediction via Multi-range Decoupling Decoding with Gating-adjusting Aggregation}

\author{\IEEEauthorblockN{Jiexin Wang\textsuperscript{1}, Wenwen Qiang\textsuperscript{2}, Zhao Yang\textsuperscript{1}, Bing Su\textsuperscript{1,*}}
\IEEEauthorblockA{\textit{\textsuperscript{1} Gaoling School of Artificial Intelligence, Renmin University of China, Beijing, China}}
\IEEEauthorblockA{\textit{\textsuperscript{2} Institute of Software Chinese Academy of Sciences, Beijing, China}}
}

\maketitle

\begin{abstract}
Expressive representation of pose sequences is crucial for accurate motion modeling in human motion prediction (HMP). 
While recent deep learning-based methods have shown promise in learning motion representations, these methods tend to overlook the varying relevance and dependencies between historical information and future moments, with a stronger correlation for short-term predictions and weaker for distant future predictions. This limits the learning of motion representation and then hampers prediction performance. 
In this paper, we propose a novel approach called multi-range decoupling decoding with gating-adjusting aggregation ($MD2GA$), which leverages the temporal correlations to refine motion representation learning. This approach employs a two-stage strategy for HMP. In the first stage, a multi-range decoupling decoding adeptly adjusts feature learning by decoding the shared features into distinct future lengths, where different decoders offer diverse insights into motion patterns. In the second stage, a gating-adjusting aggregation dynamically combines the diverse insights guided by input motion data. 
Extensive experiments 
demonstrate that the proposed method can be easily integrated into other motion prediction methods and enhance their prediction performance. 
\end{abstract}


\section{Introduction}
\label{sec:intro}

Human Motion Prediction (HMP) is a crucial task in the field of artificial intelligence, involving the generation of future pose sequences based on historical motion data. HMP, by enabling machines to understand and anticipate human motion behavior, has garnered significant attention and found diverse applications across various domains, such as human-robot interaction~\cite{unhelkar2018human}, intelligent surveillance~\cite{xu2015learning}, and autonomous driving~\cite{chen20203d}. Accurate predictions of human motion are pivotal in enhancing system performance, safety, and efficiency in these applications. The primary aim of HMP lies in developing a powerful mapping function that can effectively bridge past motion behavior to future motion.


Recent advances in deep learning have significantly influenced HMP. RNN and its variants~\cite{gopalakrishnan2019neural,corona2020context,lee2021video} exploit the temporal characteristics of motion sequences and have shown promising results in capturing complex motion patterns, but these methods suffer from the error accumulation problem.
To address this, the mainstream framework of HMP is to generate all prediction targets at once~\cite{mao2019learning,li2022skeleton,xu2023eqmotion,xu2023auxiliary}. This approach leverages the motion representation learned from the historical data to concurrently decode target poses across future instances, enabling more accurate and reliable predictions.
However, this approach disregards the inconsistent correlation strengths between historical data and different future moments, leading to limitations in feature representation. 
As a result, the generated representations may not be fully expressive and informative.

\begin{figure}[!t]
\centering
\subfigure[GCN.]{\includegraphics[width = 0.15\textwidth]{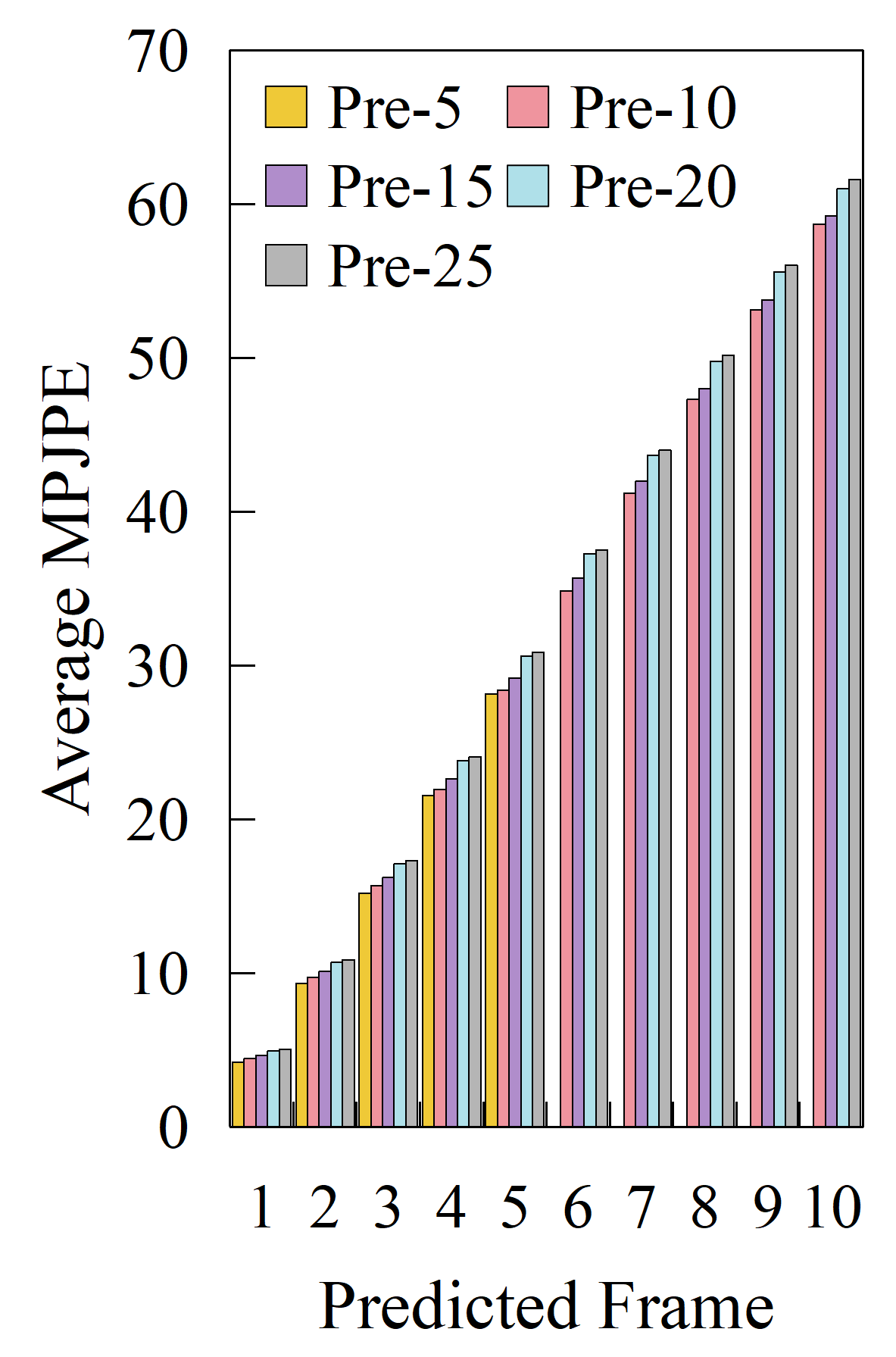}\label{fig:pre_len_gcn}}\hfill
\subfigure[LSTM.]{\includegraphics[width = 0.15\textwidth]{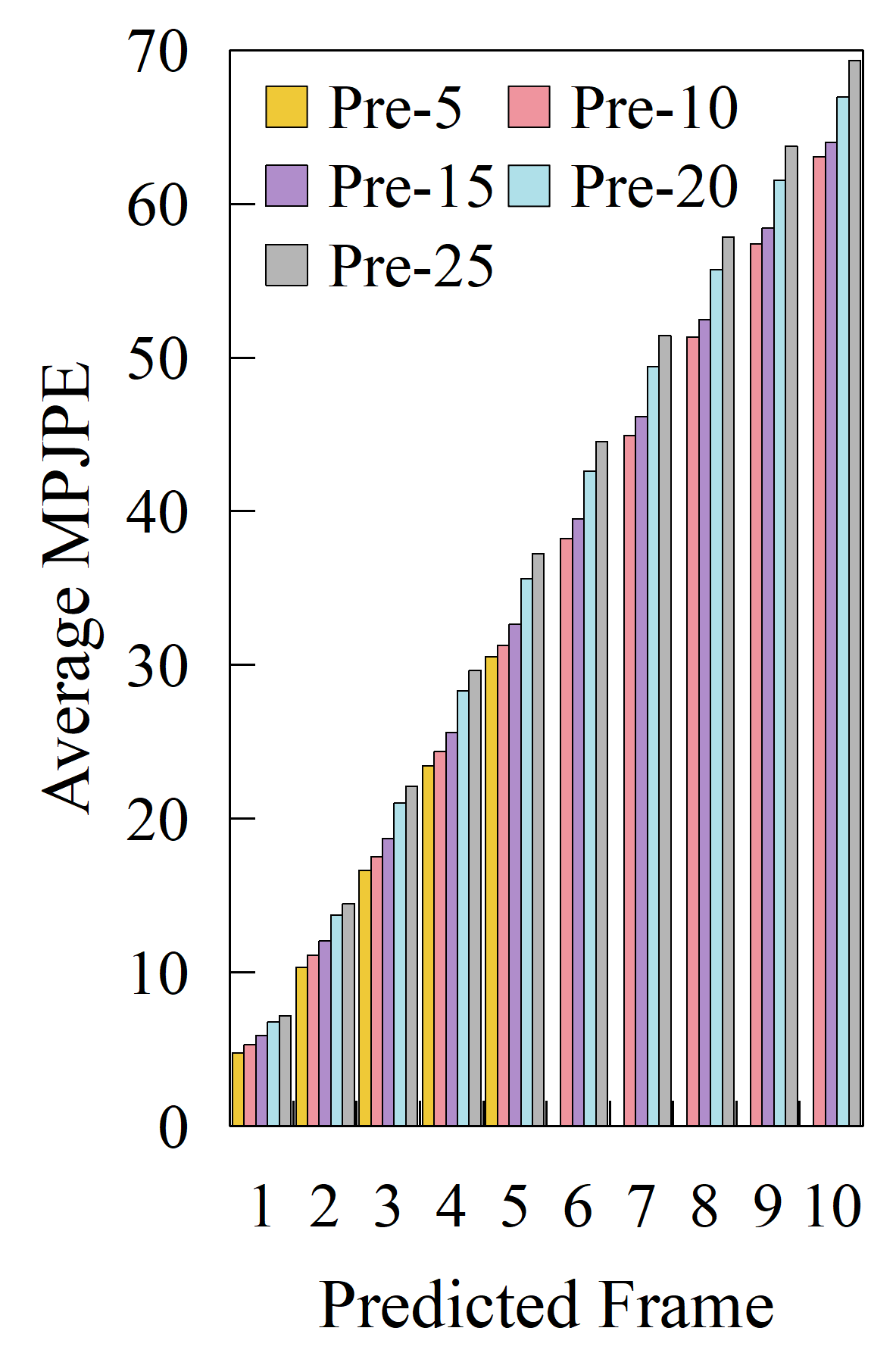}\label{fig:pre_len_lstm}}\hfill
\subfigure[Transformer.]{\includegraphics[width = 0.15\textwidth]{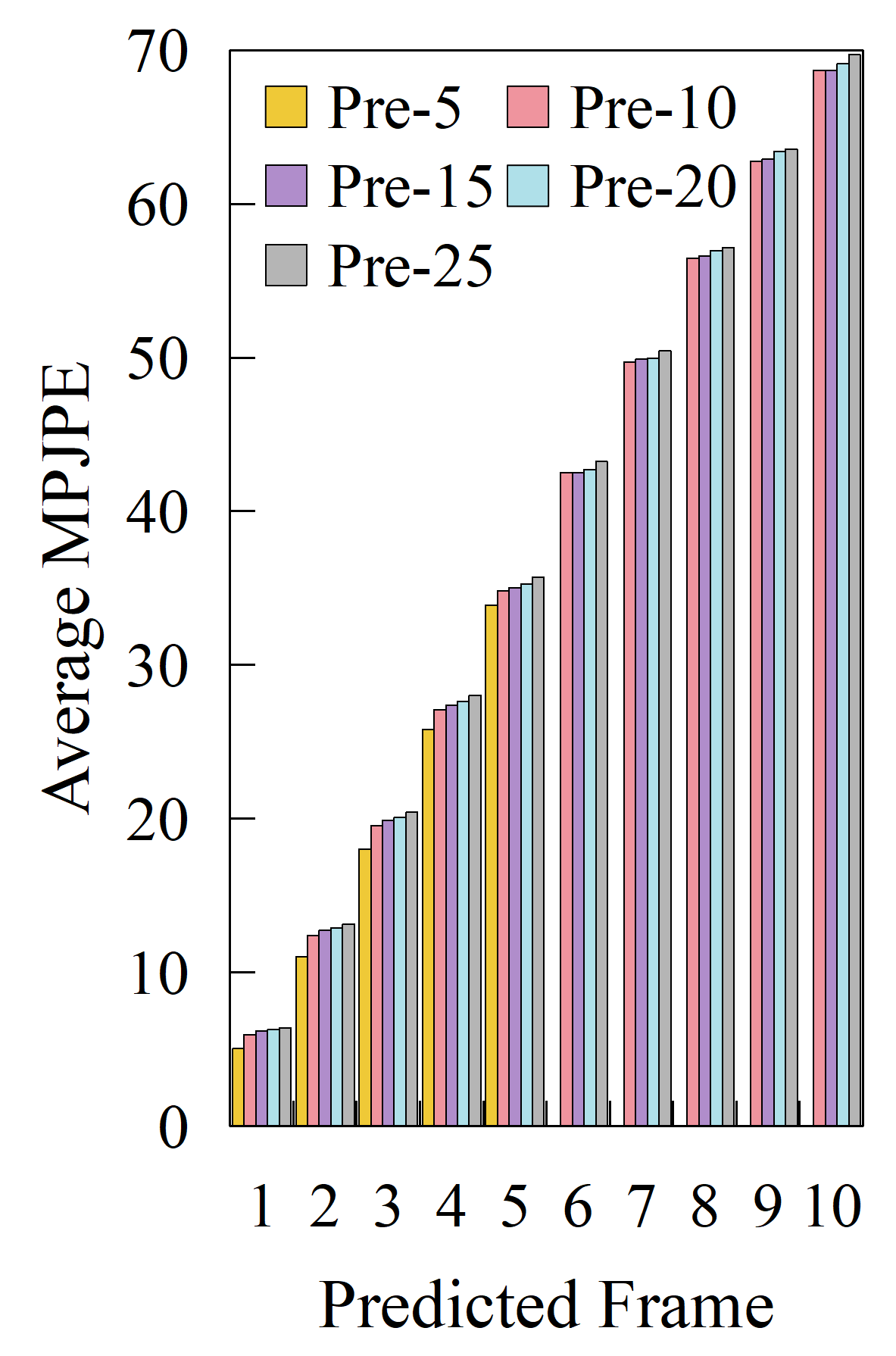}\label{fig:pre_len_transformer}}
\vspace{-0.1in}
\caption{Performance across various future prediction horizons in toy experiments. Experiments are conducted on a GCN/LSTM/Tansformer-based neural network utilizing 10 initial poses as input, with Pre-$x$ denoting the corresponding prediction sequence length ($x$) in the Human 3.6M dataset~\cite{ionescu2013human3}. The figure demonstrates the increasing difficulty of predicting farther into the future, as evidenced by performance degradation from the 1st to the 10th frame. Moreover, it shows that wider prediction horizons lead to poorer short-term prediction performance, as seen in the comparison of the shared prediction horizons between different Pre-$x$ settings.
}
\vspace{-0.2in}
\label{fig:pre_len}
\end{figure}

As depicted in~\figurename{~\ref{fig:pre_len}}, we can view that the short-term predictions consistently outperform the long-term ones across all models, and expanding the prediction horizon leads to degraded short-term prediction performance. The former highlights the learned features exhibit stronger correlations with short-term future predictions but weaker ties with long-term predictions. The latter reveals that the main framework cannot fully exploit the strong correlation of the learned motion representation for the short-term future, such as permutation-invariant and somewhat “anti-order”.
From the representation perspective, models solely focusing on short-term predictions may face overfitting issues, potentially limiting the learning of motion representation. 
In contrast, the long-term prediction task requires a deeper understanding of motion information, urging models to extract more expressive motion representation. Hence, superior short-term predictions under long-term forecasting are expected. 
However, this contradicts the learned motion representation to prioritize overall generation performance, as the objective function equally emphasizes all behaviors at different time steps (e.g., Mean Per Joint Position Error, MPJPE). 
These insights emphasize the significance of exploring techniques that harness temporal correlations to enhance motion representation learning and subsequently improve prediction performance.


Building on the insights derived from the aforementioned observations, we propose an innovative approach that effectively leverages temporal correlations to refine motion representation learning. 
Specifically, we propose a multi-range decoupling decoding architecture that allows for adaptive adjustment of the learned motion representations. Each decoding is tailored to predict actions for distinct future time horizons, accommodating the varying relevance and dependencies. This strategy bolsters the expressiveness of learned motion representations.
Furthermore, to amalgamate insights from these decoding, we propose a gating-adjusting aggregation mechanism. This dynamic mechanism assigns varying weights to outputs from diverse decoders, facilitating an optimal fusion of predictions.
Empowering by above designs, our method can enhance the model's capability to capture fine-grained temporal dependencies, elevating the quality of learned motion representations. Consequently, this leads to more accurate and context-aware human motion predictions. 
Notably, our work is the first attempt to consider using different decoding for different prediction horizons with diverse insights in HMP.
Through extensive experiments on standard HMP benchmarks, we demonstrate the effectiveness and superiority of our method.
The results underscore its potential to surmount the limitations of the mainstream framework, showcasing its ability to pave the way for substantial advancements in HMP.
The contributions of this paper are summarized as follows:
\begin{enumerate}
\item We propose a novel extension to the mainstream framework generating all prediction targets at once. Our approach involves multi-range decoupling decoding, enabling adaptive adjustments of the learned representation and effectively alleviating the challenges of varying correlation strengths across different prediction horizons. 
\item We propose a gating-adjusting aggregation mechanism that adaptively weights the outputs generated by the multi-range decoupling decoding. Our method leverages the associations of diverse decodings with information at different prediction horizons, facilitating the effective capture of nuanced dependencies and enhancing predictive performance.
\item We conduct extensive experiments and integrated our method into many existing methods on standard HMP benchmarks. Results demonstrate the effectiveness and superiority of our method in accurately predicting human motion across diverse scenarios and prediction horizons, and it can be easily integrated into existing methods.
\end{enumerate}
\section{Methodology}
\label{Methodology}

\begin{figure}[!t]
    \centering
    \includegraphics[width=0.99\linewidth]{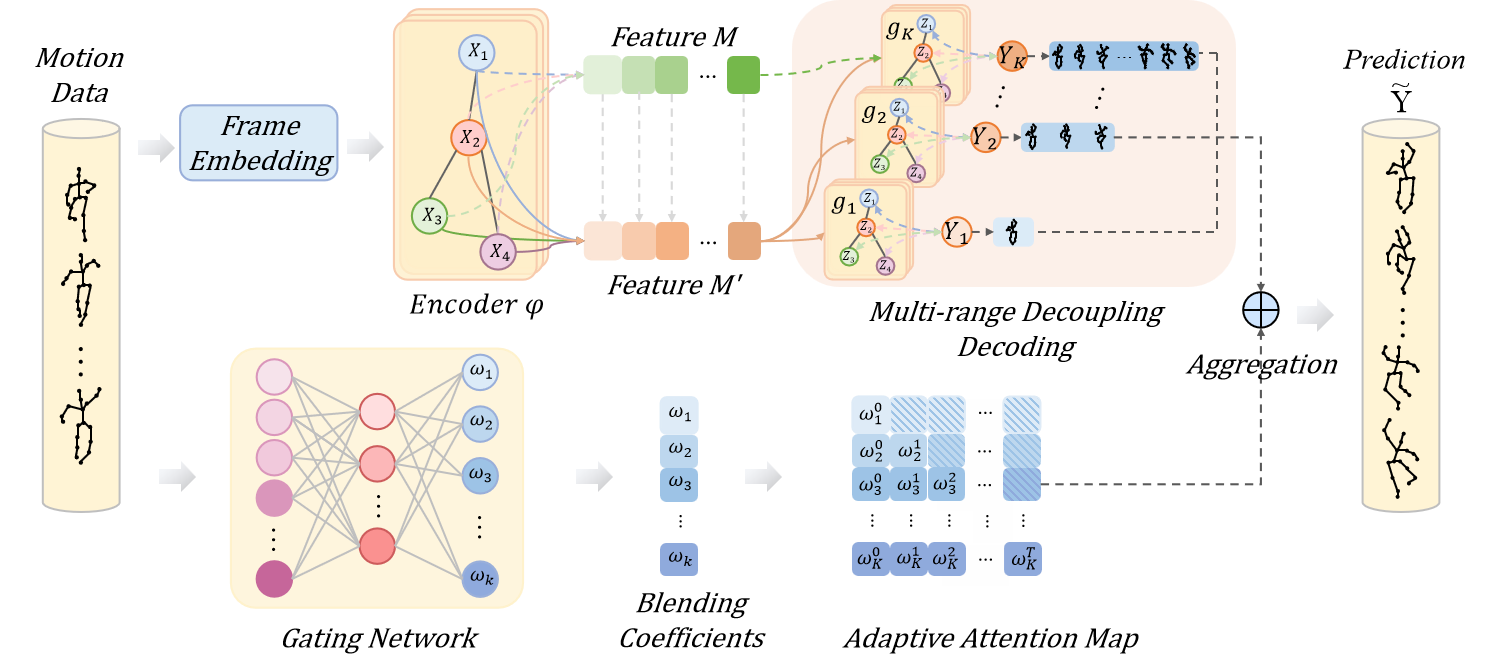}
    \vspace{-0.2in}
    \caption{Illustration of our multi-range decoupling decoding with gating-adjusting aggregation framework. Within our approach, we expand the mainstream framework $\mathcal{F}_\mathrm{{pred}}$ comprising an encoder $\varphi$ ($e.g.$ graph neural networks) and a decoder $g$. We extend $g$ into multi-range decoupling decoding, adjusting adaptively feature learning by decoding shared features into distinct future lengths, transitioning from the motion feature $M$ to $M'$. Furthermore, we propose a dynamic gating-adjusting aggregation mechanism to combine diverse insights derived from the multi-range decoupling decoding.}
    \label{fig:method}
    \vspace{-0.2in}
\end{figure}

\subsection{Problem Formulation}
The human motion prediction problem aims at forecasting the future pose sequences from given historical ones. Mathematically, we represent the given historical poses as $\mathbf{X}=[X_1,\cdots,X_{T_p}] \in \mathbb{R}^{T_p \times J \times D}$, where $T_p$ frames are recorded. Each pose $X_{t} \in \mathbb{R}^{J \times D}$ depicts a human pose with $J$ joints in $D$ dimensions at time $t$. Typically, $D$ is equal to 2 or 3, indicating 2D or 3D case, respectively. Similarly, the target future pose sequence is defined as $\mathbf{Y}=[X_{T_p+1},\cdots,X_{T_p+T_f}] \in \mathbb{R}^{T_f \times J \times D}$. Thus, the HMP problem is formally defined as predicting $\mathbf{Y}$ given $\mathbf{X}$. 
The main challenge of HMP lies in formulating an effective predictor $\mathcal{F}_\mathrm{{pred}}$, which can effectively bridge past body motion $\mathbf{X}$ to the future $\mathbf{Y}$. 
Formally, the problem is represented as:
\begin{equation}
\widetilde{\mathbf{Y}} = \mathcal{F}_\mathrm{{pred}}(\mathbf{X}).
\label{EQ:objective}
\end{equation}
The objective of HMP is that the prediction $\widetilde{\mathbf{Y}}$ are as accurate as possible compared to the ground-truth motions $\mathbf{Y}$. 

\subsection{Paradigm Review}
\textbf{Conventional Approach.} Given $\mathbf{X}$, a prediction module $\mathcal{F}_\mathrm{{pred}}$ consists of a frame embedding to project $X$ into a feature space $\hat{X} = W_{2} ( \sigma ( W_{1}  \mathbf{X} + b_1) ) + b_2$, an encoder $\varphi$ for modeling spatio-temporal dependencies in the skeleton data $M = \varphi(\hat{X})$, and a decoder $g$ for generating the prediction $\widetilde{\mathbf{Y}} = g(M)$. The generic formulation can be summarized as:
\begin{equation}
    \mathcal{F}_\mathrm{{pred}}(\mathbf{X}) = g\left( \varphi (\hat{X})  \right).
    \label{Eq:tradition}
\end{equation}

\textbf{Proposed Approach.}  While existing methods in HMP have explored various techniques for capturing motion patterns~\cite{sofianos2021space,li2022skeleton,xu2023auxiliary}, they often overlook the fact that historical motion information provides distinct reference values for different future moments. 
As a result, it becomes essential not only to focus on the ultimate prediction outcome but also to account for the evolving correlations in timing information.
To this end, we propose an effective and straightforward framework, as depicted  in~\figurename{~\ref{fig:method}}, which extends the decoding progress of the mainstream framework into a two-stage strategy:
\begin{equation}
    \mathcal{F}_\mathrm{{pred}}(\mathbf{X}) = GA \left( MDD \left( \varphi (\hat{X}) \right) \right).
    \label{Eq:our}
\end{equation}
In the decoding stage, we employ a Multi-range Decoupling Decoding ($MDD$) with $K$ decoders, to decode the motion representation into different future time horizons, and each decoder outputs an output. Subsequently, in the following aggregation stage, we introduce a powerful Gating-adjusting Aggregation ($GA$) mechanism that adaptively combines the outputs from $MDD$ based on the input data. In the following sections, we will detail the techniques of these stages.

\subsection{Multi-range Decoupling Decoding} 
The core of our approach centers on multi-range decoupling decoding, which plays a pivotal role in enhancing representation learning. By effectively leveraging inherent temporal cues, our approach fine-tunes the learned features, capturing intricate dependencies that span various future prediction horizons.
Recognizing the impact of temporal correlation on representation quality, we decode the motion representation $M$ to different prediction horizons. 
Specifically, we propose the multi-range decoupling decoding consisting of $K$ decoders $(g_1,\cdots,g_K)$, each responsible for decoding the representation into distinct future horizons. For each decoder $g_k$, $k=1,\cdots,K$, the translation of $M$ results in an output denoted as $Y_k$.  It is important to note that the structures of $g_k$ and $g$ remain consistent, with the only disparity being in the timing dimension of the output. The prediction process can be described as follows:
\begin{equation}
    Y_k = g_k(M) + \mathbf{X}[:L_k] \in \mathbb{R}^{L_k \times J \times D},
    \label{decoder}
\end{equation}
where $L_k$ is the prediction horizon of $g_k$, the prediction horizon is different for different decoders in our method. Importantly, each decoder follows the general decoding framework specified in Eq.\ref{decoder}, with the ability to generate different prediction horizon sequences. The specific design of $L_k$ adheres to the formula outlined below:
\begin{equation}
L_k = \left\lfloor (k-1) \cdot dis \right\rfloor + 1 + T_p,
\label{L_i}
\end{equation}
\begin{equation}
dis = \frac{T_f - 1}{K- 1}, K\geq 2,
\label{L_dis}
\end{equation}
where $dis$ indicates the difference in decoding horizon between neighbouring decoders.
By deploying the multiple decoders within the motion representation $M$, $MDD$ generates $K$ outputs capturing diverse prediction horizons, then extract rich temporal information.

\subsection{Gating-adjusting Aggregation}
After decoding $M$ into a set of $K$ output sequences, the task of effectively fusing these outputs becomes challenging, especially when dealing with diverse action types and variations within human motion. This difficulty is amplified by the range of prediction horizons and diverse motion patterns. Leveraging the observation that the predictive capabilities of learned features exhibit variations across different prediction horizons, we propose a gating-adjusting aggregation mechanism. Our objective is to dynamically establish adaptive relationships to address multi-motion prediction challenges, driven by the inherent characteristics of the motion data.

Inspired by the concept of the mixture of experts, which employs a gating network to autonomously assign task importance~\cite{eigen2013learning,shazeer2017outrageously}, we employ a gating mechanism for prediction blending. This strategy maintains a lightweight network while exclusively affecting feature learning, leaving the feature transfer process unaffected. Concretely, our gating network operates on the predictions of $MDD$, producing a set of blending coefficient parameters as follows:
\begin{equation}
    \omega=Gating(\mathbf{X})=softmax(FC(\mathbf{X})),
    \label{gatingnetwork}
\end{equation}
where $FC$ comprises 3 fully connected layers, and the softmax function serves as the activation function. The resulting $\omega$ represents the blending coefficients as $\omega = \{\omega_1,\cdots,\omega_K\}$.

Considering that the outputs of $MDD$ are inconsistent in terms of prediction horizon, some decoders have outputs that have no effect on some prediction targets. Therefore, we devise a simple attention masking strategy to adaptively adjust the attention scores $\omega$ to fit the mix of outputs at different moments. Specifically, we do this as following:
\begin{equation}
    B_{k,t}=\{
    \begin{array}{rcl}
        1 & & t\leq L_k \\
        0 & & L_k<t\leq T_f,
    \end{array}
\end{equation}
then the adaptive attention are obtained using $\omega$ as:
\begin{equation}
    A_{k,t}= \frac{exp(\omega_k)*B_{k,t}}{\sum_{k=1}^{K} exp(\omega_k)*B_{k,t}},
\end{equation}
where $exp(*)$ denotes the natural exponential function. These adaptive attention scores are then employed to blend candidate predictions, resulting in the adaptive prediction $\widetilde{\mathbf{Y}}=\{P_1,\cdots,P_{T_p+T_f}\}\in \mathbb{R}^{(T_p+T_f) \times J \times D}$:
\begin{equation}
P_t = \sum_{k=1}^{K}Y_{k,t}\cdot A_{k,t}, 1\leq t\leq(T_p+T_f) \wedge A_{k,t}\ne 0.
\end{equation}

\subsection{Loss Function}
Given a predictive model with an initial loss function, denoted as $\mathcal{L}_1$, where $\mathcal{L}_1$ can represent the Mean Per Joint Position Error (MPJPE) or incorporate additional regularization terms. For analysis purposes, let's focus on the scenario where MPJPE serves as the original loss function. Formally, for one single training sample, we have the loss as follows:
\begin{equation}
    \mathcal{L}_1 = \frac{1}{(T_p+T_f)\cdot J} \sum_{t=1}^{T_p+T_f} \sum_{j=1}^{J} \parallel \widetilde{\mathbf{Y}}_{t,j} - \mathbf{Y}_{t,j} \parallel^2 ,
\end{equation}


\begin{equation}
    \mathcal{L}_2 = \sum_{k=1}^{K}\frac{1}{L_k\cdot J} \sum_{t=1}^{L_k} \sum_{j=1}^{J} \parallel Y_{k,t,j} - \mathbf{Y}_{t,j} \parallel^2 ,
\end{equation}
where $\mathcal{L}_1$ constrains predictions from each decoder to correlate with the shared representation across different prediction horizons, and $\mathcal{L}_2$ jointly regulates multiple decoders' outputs.
Then we combine $\mathcal{L}_1$ and $\mathcal{L}_2$ into a final loss function:
\begin{equation}
    \mathcal{L} = \mathcal{L}_1 +  \mathcal{L}_2.
\end{equation}

\section{Experiments}
\label{sec:experiment}

\begin{table}[!t]\scriptsize
\centering
\setlength{\tabcolsep}{6.5pt} 
\caption{Comparisons of average prediction MPJPEs in H3.6M. Results for motion prediction at 80ms, 160ms, 320ms, 400ms, 560ms and 1000ms in the future are shown.
The better results are highlighted in \textbf{bold} within the same method.
}
\label{tab:h3.6m}
\begin{tabular}{lcccccc}
\toprule
 \multirow{1}{*}{Method}  
  & 80ms & 160ms & 320ms & 400ms & 560ms & 1K ms\\ \hline
 Traj-GCN & 12.19  &	24.87 &	50.76 &61.44 &	80.19 &	113.87 \\
 Traj-GCN-M & \textbf{11.17} 	&\textbf{23.85} 	&\textbf{49.77} 	&\textbf{60.60} 	&\textbf{78.50} 	&\textbf{112.36}	 	 	 	 
 \\  \hline
 SPGSN   & 10.74 &22.68 &47.46 &58.64 &79.88 &112.42 \\
 SPGSN-M & \textbf{10.23} &\textbf{22.17} &\textbf{47.07} &\textbf{58.47} &\textbf{78.94} &\textbf{111.69}  \\  \hline
 EqMotion  & 9.45 &21.01 &46.06 &\textbf{57.60} &75.98 &109.75\\ 
 EqMotion-M& \textbf{8.90} &\textbf{20.56} &\textbf{45.96} &57.89 &\textbf{75.82} &\textbf{109.51} \\ \hline
 STBMPT   & 10.73 &23.70 &49.72 &61.21 &82.26 &113.85 \\
 STBMPT-M & \textbf{10.36} &\textbf{23.11} &\textbf{49.01} &\textbf{60.70} &\textbf{80.94} &\textbf{113.06} \\ \hline
 STBMPS   & 9.56 &21.80 &47.18 &58.64 &80.75 &113.98 \\
 STBMPS-M & \textbf{9.37} &\textbf{21.55} &\textbf{47.00} &\textbf{58.45} &\textbf{79.40} &\textbf{112.62} \\  
\bottomrule
\end{tabular}
\vspace{-0.1in}
\end{table}

\begin{table}[!t]\scriptsize
\centering
\setlength{\tabcolsep}{6.5pt} 
\caption{Comparisons of average MPJPEs across all actions in the CMU-Mocap dataset.
}
\label{tab:cmu}
\begin{tabular}{lccccccc}
\toprule
Method  & 80ms & 160ms & 320ms & 400ms & 560ms & 1K ms\\ \hline 
Traj-GCN   & 11.34 &19.96 &37.64 &46.57 &62.15 &96.91 \\
Traj-GCN-M &  \textbf{10.62} &\textbf{18.76} &\textbf{35.87} &\textbf{44.54} &\textbf{59.83} &\textbf{96.64} \\ \hline
SPGSN      &  10.83 &19.72 &36.84 &44.94 &59.37 &88.00 \\
SPGSN-M    &  \textbf{9.24} &\textbf{16.80} &\textbf{31.99} &\textbf{39.50} &\textbf{53.38} &\textbf{83.07} \\ \hline
STBMPT   & 9.10 &16.47 &31.72 &39.49 &53.73 &79.48 \\
STBMPT-M & \textbf{8.37 } &\textbf{15.35} &\textbf{15.35} &\textbf{37.93} &\textbf{51.99} &\textbf{79.37} \\ \hline 
STBMPS   & 8.76 &16.54 &33.77 &42.43 &57.99 &86.70 \\
STBMPS-M & \textbf{8.07} &\textbf{15.13} &\textbf{30.87} &\textbf{38.83} &\textbf{53.55} &\textbf{82.41}  \\ 
\bottomrule
\end{tabular}
\vspace{-0.2in}
\end{table}

\subsection{Datasets and Baselines}

\textbf{Datasets}. We evaluate our proposed method on the Human3.6m (H3.6M), CMU Motion Capture (CMU-Mocap), and 3D Poses in the Wild (3DPW) datasets.  The detailed descriptions of these datasets are provided in Appendix.


\textbf{Baselines.} 
To establish baselines for comparison, we use several state-of-the-art and open-source baselines, including Traj-GCN~\cite{mao2019learning}, SPGSN~\cite{li2022skeleton}, Eqmotion~\cite{xu2023eqmotion}, STBMP~\cite{wang2023spatio} and AuxFormer~\cite{xu2023auxiliary}.
We distinguish the baselines integrated with our method by adding the suffix ``-M'', $e.g.$, SPGSN-M represents SPGSN integrated with ours. 
For all experiments, we follow the training setup of the original method without any modification to assess the performance of our method. 

\textbf{Evaluation Metric.}  
Mean Per Joint Position Error (MPJPE) is the most widely used evaluation metric in HMP. 
A lower MPJPE value indicates better predictive performance.


\subsection{Comparison with Prediction Performance}
To evaluate the effectiveness of our method, we carry out five experiments for each combination between the original and proposed method across all datasets. 
The reported results are the average scores obtained from these experiments.
STBMPT and STBMPS denote the temporal and spatial branches of STBMP, respectively, without incorporating incremental information. Furthermore, we also use T-SNE~\cite{van2008visualizing} to illustrate the motion feature for qualitative evaluation. 

\textbf{Short-term motion prediction.}
Short-term motion prediction aims to predict future poses within 400 milliseconds. 
\tablename{~\ref{tab:h3.6m}} shows the average MPJPEs of 15 actions in H3.6M, 
all baselines achieve an average reduction of over 1.3 points in MPJPE after the integration of our approach.
The results of CMU-Mocap are presented in \tablename{~\ref{tab:cmu}},
all baselines exhibited an average decrease of more than 4.7 points in MPJPEs. 
In \tablename{~\ref{tab:3dpw}}, the results on 3DPW further confirm that our method consistently enhances model performance across all baselines and prediction lengths. More detailed results can be found in Appendix \tablename{~\ref{tab:h3.6M—shortterm-appendix}} and \ref{tab:CMU-Mocap-appendix}. Besides, \figurename{~\ref{fig:rand_seed_h36m}} to~\ref{fig:rand_seed_3dpw}  in the Appendix show that the models with our method all outperform the performance of the original and exhibit smaller error bars across various architectures, which highlights the superiority and stability of our approach.

\begin{table}[!t]\scriptsize
\centering
\setlength{\tabcolsep}{6.5pt} %
\caption{The average prediction MPJPEs across the test set of 3DPW at various prediction time steps.}
\label{tab:3dpw}
\begin{tabular}{lccccccccccc}
\toprule
Method & 100ms & 200ms & 400ms & 600ms & 800ms & 1K ms \\ \hline 
Traj-GCN  & 19.91 &40.12  &73.74  &93.19  &104.38  &111.72 \\
Traj-GCN-M& \textbf{18.04} &\textbf{36.93} &\textbf{70.50}  &\textbf{90.76} &\textbf{103.23} &\textbf{111.27} \\ \hline
SPGSN     & 19.71 	&40.35 	&74.85	&94.62 &105.30 &112.04 \\
SPGSN-M &  \textbf{18.54} &\textbf{38.44} &\textbf{72.68} &\textbf{92.87}  &\textbf{104.44}  &\textbf{111.20} \\ \hline 
STBMPT    & 19.99 	&41.19 &76.48  	&97.41 	&110.45	&119.79 \\
STBMPT-M  & \textbf{17.85 }&\textbf{37.15 } &\textbf{71.67 } &\textbf{94.17 }  &\textbf{108.24} &\textbf{117.06} \\ \hline
STBMPS    & 16.39 	&36.33	&75.50	&101.70 &118.18 &128.85 \\
STBMPS-M  & \textbf{16.00 }&\textbf{35.49} &\textbf{73.17} &\textbf{98.32}  &\textbf{114.10} &\textbf{123.61} \\ 
\bottomrule
\end{tabular}
\vspace{-0.2in}
\end{table}


\begin{figure}[!t]
  \centering
  \subfigure[STBMPT.]{\includegraphics[width=0.35\linewidth]{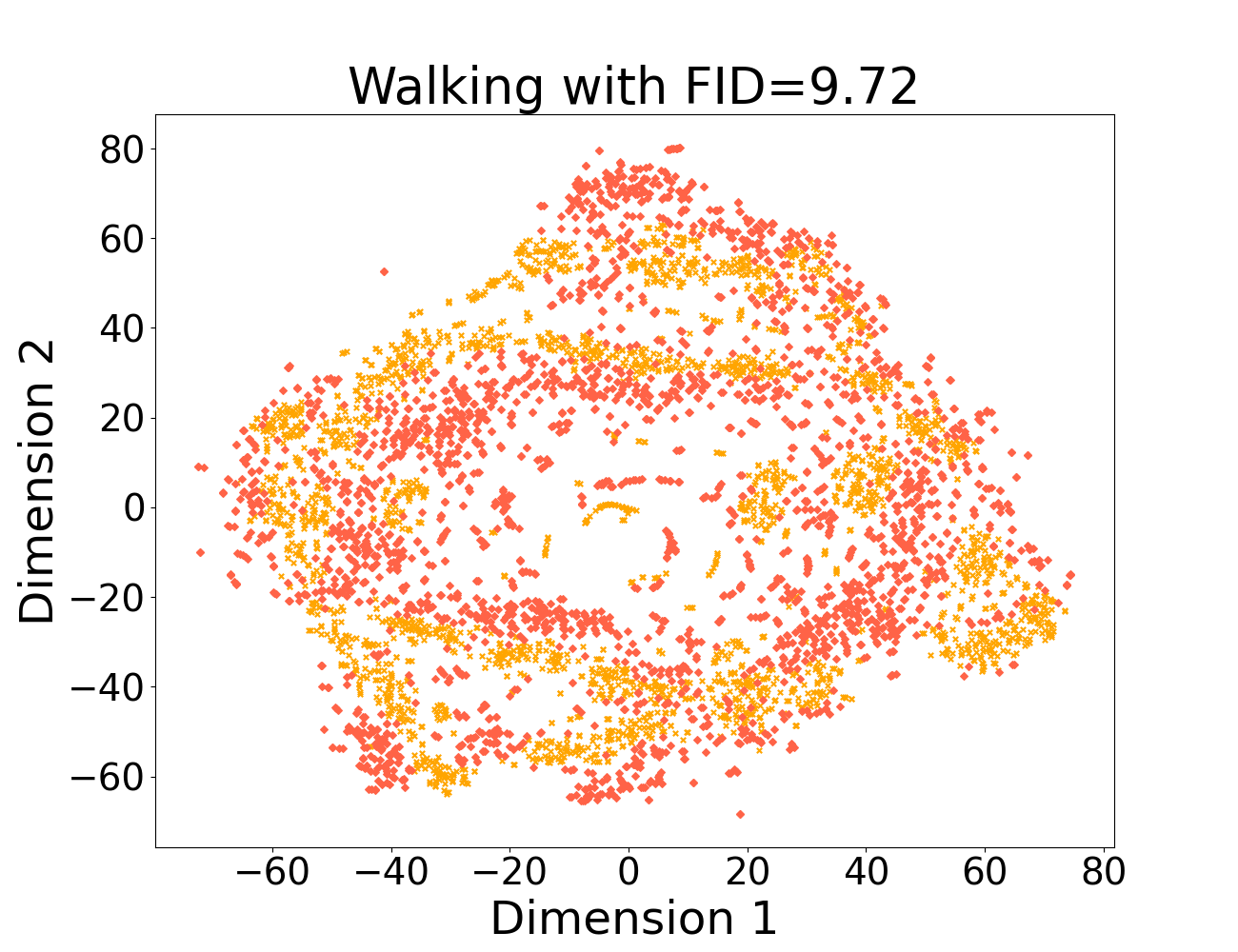} \includegraphics[width=0.35\linewidth]{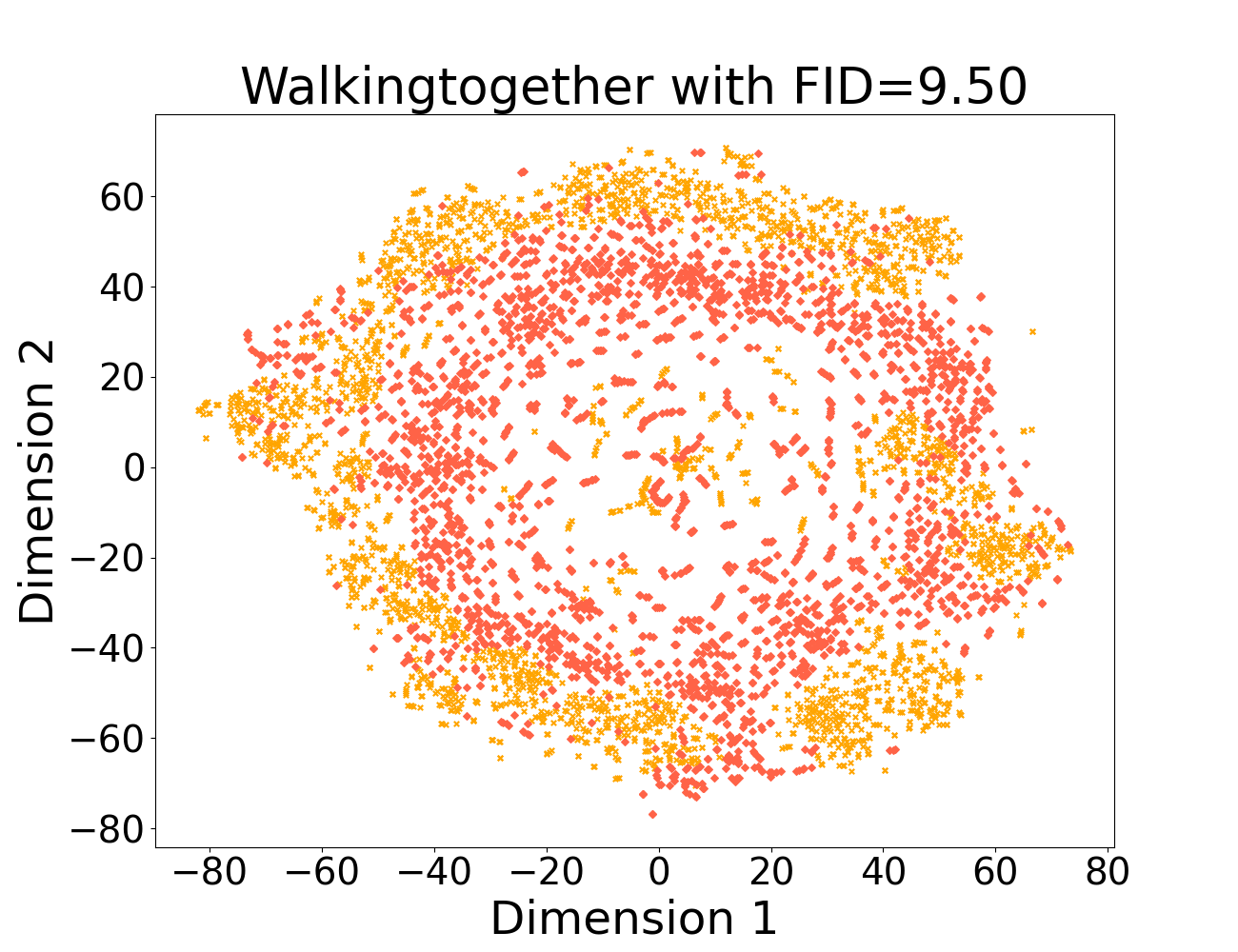}}
  \subfigure[STBMPT integrated with our method.]{\includegraphics[width=0.35\linewidth]{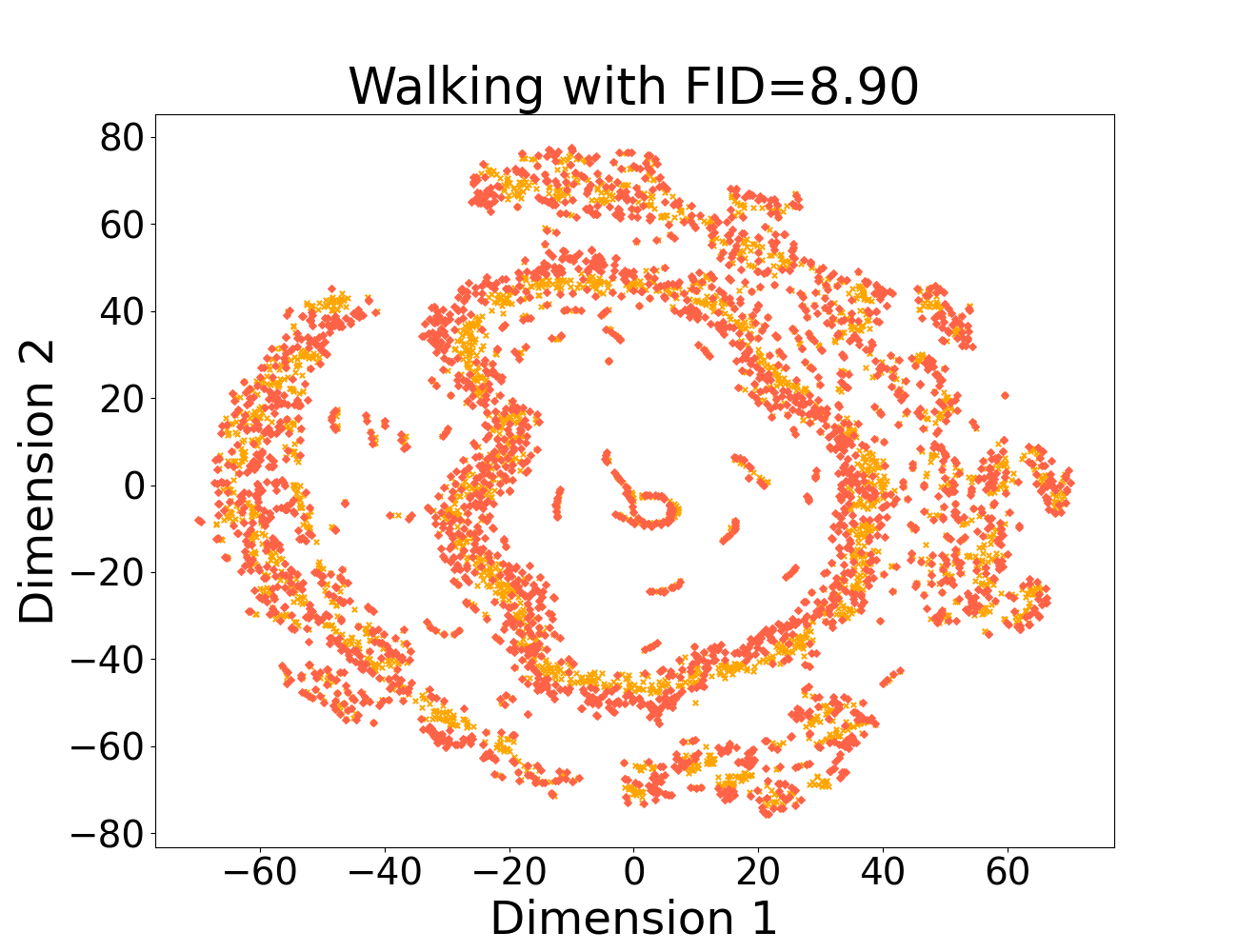} \includegraphics[width=0.35\linewidth]{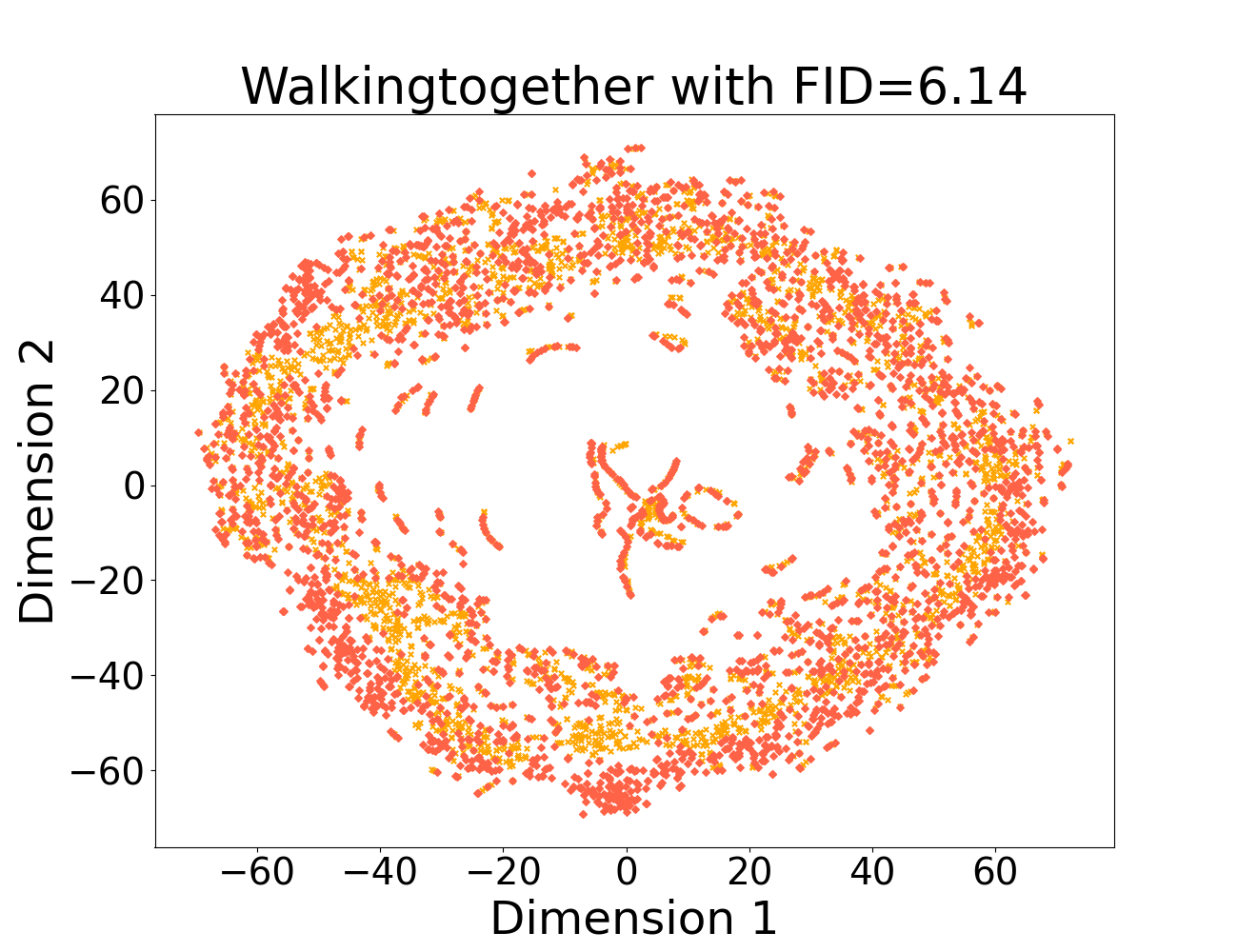}}
  \vspace{-0.15in}
   \caption{T-SNE visualization of predicted human motion. The red represents ground truth, and the yellow depicts the motion features predicted by the model. 
   }
   \vspace{-0.3in}
   \label{fig:feature_compare}
\end{figure}

\textbf{Long-term motion prediction.}
Long-term motion prediction involves forecasting poses over durations exceeding 400 milliseconds or more.
In H3.6M and CMU-Mocap datasets, we report two long-term prediction levels: 560ms and 1K ms. In the case of the 3DPW dataset, we have a richer variety of validation. As shown in~\tablename{~\ref{tab:h3.6m}} and~\tablename{~\ref{tab:cmu}}, it's important to note that in both H3.6M and CMU-Mocap datasets, our method consistently improves the baseline model's performance across all baselines and prediction lengths. 
As shown in~\tablename{~\ref{tab:3dpw}}, in 3DPW, our module also demonstrates improvements in most cases.
Specific action prediction results are shown in the Appendix \tablename{~\ref{tab:h3.6M—longterm-appendix}} to \ref{tab:CMU-Mocap-appendix} and \figurename{~\ref{fig:rand_seed_h36m}} to \ref{fig:rand_seed_3dpw}. Our method enhances the predictive performance for most action types, even in long-term prediction scenarios.
These results underscore the robustness of our proposed approach. 


\textbf{Visualization results.}
To better assess the superiority of our method, we provide visual experimental results in \figurename~\ref{fig:feature_compare}. Using T-SNE~\cite{van2008visualizing} for dimensionality reduction, we visualize predicted and ground-truth action features. Variations in the ground-truth feature distributions are due to the different motion representations captured by different models. The top and bottom halves of the figure show results before and after integrating our method, respectively, for "Walking" and "Walking together" actions. 
After incorporating our method, the predicted features become closer to the ground-truth features. We measured feature distribution disparity using Frechet Inception Distance (FID), with smaller values indicating better alignment. Our results show a notable decrease in FID for both actions. More results are provided in the Appendix. 


\begin{table}[!t]\scriptsize
\setlength{\tabcolsep}{2.5pt}
\centering
\caption{Ablation analysis of our method's architecture. The average MPJPEs across all actions in H3.6M are reported. }
\label{tab:ab_architecture}
\begin{tabular}{lcccccc}
\toprule
Method & Traj-GCN & EqMotion & SPGSN &STBMPT &STBMPS &Average  \\ \hline
 Single decoder  & 37.37 & 33.53 &34.88 &36.34 &34.29 &35.27 \\
 $w/o$ $\mathcal{L}_1$     & 36.43 & 33.36 & 34.37 &35.95 &34.19 &34.46 \\ 
 $w/o$ $\mathcal{L}_2$     & 36.72 & 33.60 & 35.18 &36.24 &34.43 &35.23 \\ 
 $w/o$ $GA$      & 36.38 & 33.53 & 34.51 & 35.92 & 34.14 &34.90 \\ 
 $P$             & 3\textbf{6.35} &\textbf{33.33} & \textbf{34.48} &\textbf{35.79} &\textbf{34.09} & \textbf{34.81} \\ \bottomrule
\end{tabular}
\end{table}

\begin{figure}[!t]
\vspace{-0.1in}
\centering
\begin{minipage}{0.48\linewidth}
\centering
\setlength{\abovecaptionskip}{0.28cm}
\includegraphics[width=0.99\linewidth]{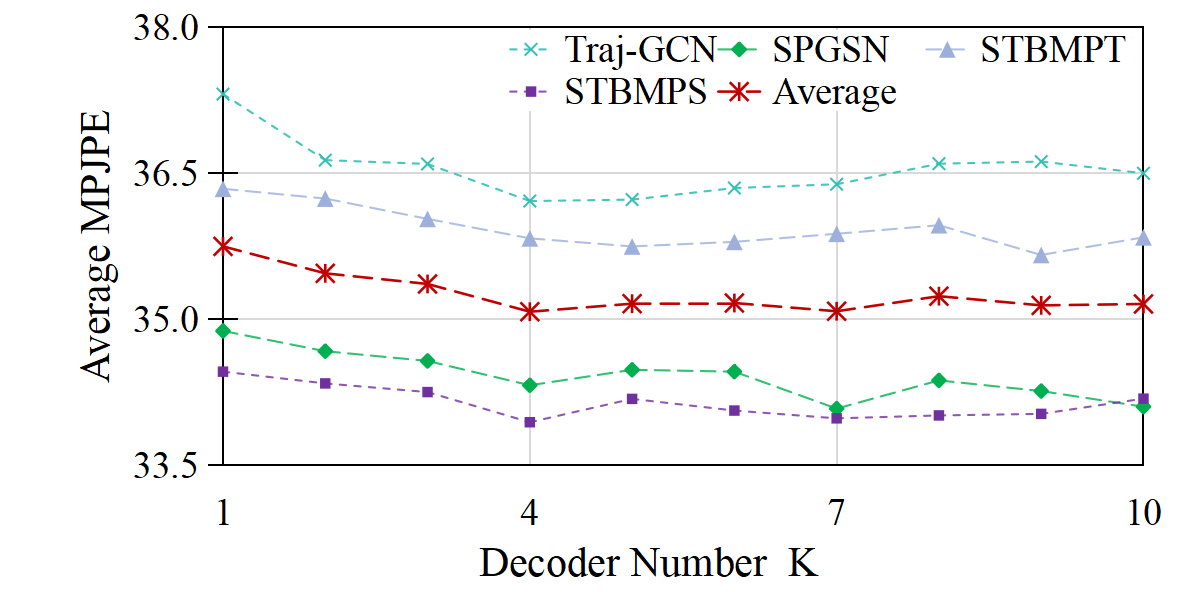}
\vspace{-0.2in}
\caption{Ablation on the number of decoding. Average MPJPE of short-term motion prediction is reported.}
    \label{fig:result_decodernumber}
\vspace{-0.1in}
\end{minipage}
\begin{minipage}{0.48\linewidth}
\centering
\setlength{\abovecaptionskip}{0.28cm}
\includegraphics[width=0.99\linewidth]{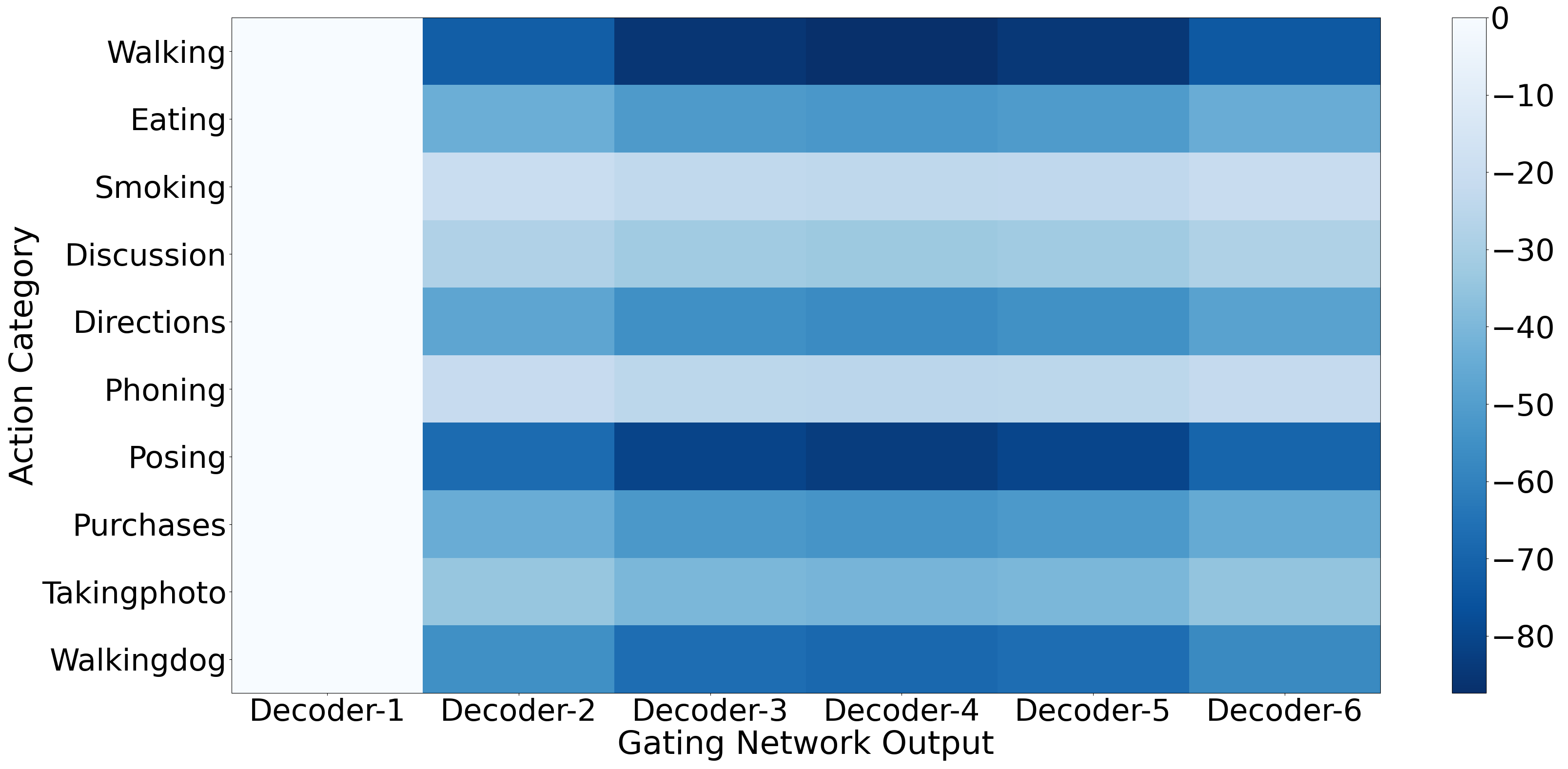}
\vspace{-0.2in}
\caption{Qualitative visualization of the adaptive attention across different actions in SPGSN with our method.}
    \label{fig:result_gatingnetwork}
\vspace{-0.1in}
\end{minipage}
\vspace{-0.1in}
\end{figure}

\subsection{Model Analysis}

\textbf{Ablation study.}
To gain deeper insights into our approach, we conduct experiment in H3.6M, analyzing the impact of the number of decoders, $MDD$, and $GA$.
\tablename~\ref{tab:ab_architecture} shows comprehensive ablation experiments on different variants of the full model. "Single decoder" signifies the utilization of a sole decoder ($g$) during both the training and testing phases in our approach. To validate $\mathcal{L}_1$, $\mathcal{L}_2$, and $GA$ roles, we individually removed them. Removal of any component noticeably reduces performance compared to the full model, highlighting the collective efficacy of these components.

\textbf{Number of decoders.}
As demonstrated in~\figurename~\ref{fig:result_decodernumber},  
applying multi-range decoupling decoding focusing on different prediction horizons notably improves the performance of the original model. Average MPJPE initially decreases with an increasing decoder count, followed by a stable trend. Despite variations of improved performance, the improvement is consistently significant, with $K \geq 2$ consistently outperforming $K=1$. Long-term performance under different $K$ setting are provided in the Appendix.
Experimentally, considering a finer or coarser division of horizons has little impact, we fix $K=6$ to 
avoid potential outlier effects. Additionally, the dynamic adjustment of decoding prediction influenced by $GA$ plays a pivotal role in shaping the final prediction. This phenomenon is visually illustrated in~\figurename~\ref{fig:result_gatingnetwork} via a heatmap visualization, where we perform a logarithmic transformation of the raw attention scores to improve the visualization of the heatmap. Lighter colors indicate a greater impact weight of corresponding decoder outputs. Notably, diverse action data offer distinct reference values for predicting varying future moments.



\begin{table}[!t]\scriptsize
\setlength{\tabcolsep}{4pt}
\centering
\caption{Comparison of different prediction horizon in H3.6M. The average MPJPEs across all actions at 80ms, 160ms, 320ms and 400ms in H3.6M are reported. }
\label{tab:ab_pre}
\begin{tabular}{lcccccc}
\toprule
Method & Traj-GCN &SPGSN &STBMPT &STBMPS &Average  \\ \hline
Ori. & 37.31 & 34.88 & 36.34 & 34.29 &35.71 \\
Ours ($L_k = T_f/K$) & 36.73 & 34.72& 35.98 &34.44 & 35.47\\
Ours ($L_k = T_p + T_f$) & \textbf{36.23} & 34.63& 36.33 &34.47 & 35.42\\
Ours & 36.35 & \textbf{34.48} & \textbf{35.79} & \textbf{34.09} & \textbf{35.18}\\
 \bottomrule
\end{tabular}
\vspace{-0.1in}
\end{table}

\textbf{Design of the prediction horizon.}
To study the proposed prediction horizon, we compare our design to two common variants. First, all decoders predict the entire target prediction horizon,  namely $L_k = T_p + T_f$. Second, different decoders predict distinct horizons and concatenate the outputs, namely $L_k = T_f/K$. \tablename~\ref{tab:ab_pre} shows the results of $MD2GA$ with these designs in H3.6M.  It is evident that our method and the variants outperform the original model in most cases, confirming the efficacy of multi-decoding. Moreover, we see that our method surpasses the two variants in most cases. There are likely two reasons for this state of affairs: 1) with $L_k = T_p + T_f$, different decoders have the same learning goal, ignoring varying correlation strengths between historical data and different future moments; 2) with $L_k = T_f/K$, each decoder only consider fragment predictions, overlooking the characteristic that the human motion itself is coherent in time dimension. These results further verify the efficacy of our proposed design in enhancing predictive performance.

\begin{table}[!t]\scriptsize
\setlength{\tabcolsep}{6pt} 
\centering
\caption{Time and model size comparisons. The average MPJPE at 80ms, 160ms, 320ms, and 400ms is also reported.}
\label{Time and mode size comparisons}
\begin{tabular}{lcccccccccc}
\toprule
 Method & Train(Per batch) & Test(Per batch) & Model Size &MPJPE \\ \hline 
 Traj-GCN  & 52ms & 27ms &2.56M & 37.31\\
 Traj-GCN-M& 64ms & 35ms &2.87M & \textbf{36.35}\\ \hline 
 SPGSN     & 144ms & 68ms & 5.67M  & 34.88\\  
 SPGSN-M   & 162ms & 73ms & 5.98M  & \textbf{34.48}\\ \hline %
 EqMotion  & 70ms & 132ms & 0.65M & 33.53\\ 
 EqMotion-M& 82ms & 160ms & 0.82M& \textbf{33.33}\\ \hline
 STBMPT    & 15ms & 11ms & 3.43M & 36.34\\ 
 STBMPT-M  & 37ms & 14ms & 4.10M & \textbf{35.79}\\ \hline
 STBMPS    & 15ms & 10ms & 3.92M & 34.29\\ 
 STBMPS-M  & 39ms & 16ms & 4.47M & \textbf{34.09}\\ 
 \bottomrule
\end{tabular}
\vspace{-0.2in}
\end{table}

\textbf{Time and Model Size Comparisons.}
In~\tablename~\ref{Time and mode size comparisons}, we evaluate the computational efficiency of our method in terms of prediction time and model size in H3.6M. The models with our method consistently exhibit superior performance with slightly increased computational costs. 
A notable advantage of our method lies in its simple architecture, without any complex module design or arithmetic mechanism.

\begin{table}[!t]\scriptsize
\setlength{\tabcolsep}{6pt}
\centering
\caption{Performance with auxiliary training task. 
The average MPJPEs across all actions in H3.6M are reported. 
}
\label{tab:auxi_task}
\begin{tabular}{lcccccc}
\toprule
Method &Prediction &80ms &160ms &320ms &400ms &Average  \\ \hline
\multirow{3}{*}{Traj-GCN} & $Ori.$ & 12.19 &24.87 &50.76 &61.44 &37.31 \\
 &$Y_K$ & 11.27 &24.04 &49.58 &60.62 &36.37\\ 
 &$P$ &\textbf{11.17} &\textbf{23.85} &\textbf{49.77} &\textbf{60.60} &\textbf{36.35} \\ \hline
 \multirow{3}{*}{SPGSN} & $Ori.$ &10.74 &22.68 &47.46 &58.64 &34.88 \\
 &$Y_K$ & 10.26 &22.28 &47.19 &58.57 &34.58\\ 
 &$P$ & \textbf{10.23} &\textbf{22.17} &\textbf{47.07} &\textbf{58.47} &\textbf{34.48} \\   \hline
 \multirow{3}{*}{EqMotion} & $Ori.$ & 9.45 &21.01 &46.06 &57.60 &33.53 \\ 
  &$Y_K$ & \textbf{8.88} &20.63 & 46.14 &58.03 & 33.42 \\   
  &$P$   & 8.90 &\textbf{20.56} &\textbf{45.96} &\textbf{57.89} & \textbf{33.33} \\ \hline
 \multirow{3}{*}{STBMPS} & $Ori.$ & 9.56 &21.80 &47.18 &58.64 &34.29 \\
 &$Y_K$ & 9.43 &21.56 &47.03 &58.46 &34.12\\
 &$P$ & \textbf{9.37} &\textbf{21.55} &\textbf{47.00} &\textbf{58.45} &\textbf{34.09} \\ 
 \bottomrule
\end{tabular}
\vspace{-0.1in}
\end{table}

\begin{table}[!t]\scriptsize
\centering
\setlength{\tabcolsep}{3.0pt} 
\caption{Comparisons of average prediction MPJPEs. }
\label{tab:_aux}
\begin{tabular}{lccccccc}
\toprule
\multirow{1}{*}{Dataset}&  \multirow{1}{*}{Mothod}  & 80ms & 160ms & 320ms & 400ms & 560ms & 1K ms\\ \hline
\multirow{2}{*}{H3.6M} & AuxFormer   & 10.00 &21.91 &46.92 &58.34 &78.98 &111.51  \\
& AuxFormer-M & \textbf{9.19} &\textbf{20.96} 	&\textbf{45.79} &\textbf{57.41} &\textbf{77.39} 	&\textbf{109.92}  \\  \hline
\multirow{2}{*}{CMU-Mocap} & AuxFormer   &  7.63 	&13.97 	&28.24 	&35.72 	&52.23 	&79.08\\
& AuxFormer-M &\textbf{7.48} &\textbf{13.78}	&\textbf{28.03}	&\textbf{35.74}	&\textbf{51.40}	&\textbf{78.48} \\ 
\bottomrule
\end{tabular}
\vspace{-0.2in}
\end{table}

\textbf{Efficacy of Our Method as a Training Auxiliary Task}. 
In~\tablename~\ref{tab:auxi_task}, we report the performance of the original model ($Ori.$), as well as the performance of the original decoder ($Y_K$) and the multi-range decoupling decoding ($P$) after integrating our method. We found that even when our method is only involved in the training phase and omitted during testing, it enhances the predictive performance of the original model's decoding (comparison between $Ori.$ and $Y_K$). Furthermore, $Y_K$'s results are close to $P$, and this does not impact the testing efficiency of the original model.
To further validate our method's effectiveness as an auxiliary task, we conduct experiments using the Transformer-based method AuxFormer. As shown in \tablename~\ref{tab:_aux}, our method enhances the predictive performance, demonstrating its potential applicability across various models.
Moreover, the effectiveness of this auxiliary task also indicates that our proposed method can effectively improve the quality of motion feature learning.


\section{Conclusion}
\label{sec:conclusion}
In this work, we propose a novel extension to the mainstream human motion prediction framework by leveraging temporal correlations.
Our approach utilizes multi-range decoupling decoding to enhance the motion feature learning, while a gating-adjusting aggregation mechanism dynamically combines predictions, both tailored to the motion data. 
Notably, our approach maintains simplicity without sacrificing effectiveness, ensuring practicality and efficiency. It is seamlessly integrated into various prediction methods. 
Extensive experiments conducted on three benchmarks demonstrate the effectiveness and superiority of our method in leveraging nuanced temporal correlations over the mainstream framework. Our approach provides a valuable alternative for motion prediction, holding promise for integration into diverse methods. 



\bibliographystyle{IEEEbib}
\bibliography{icme2025references}

\begin{thebibliography}{10}

\bibitem{unhelkar2018human}
Vaibhav~V Unhelkar, Przemyslaw~A Lasota, Quirin Tyroller, Rares-Darius Buhai, Laurie Marceau, Barbara Deml, and Julie~A Shah,
\newblock ``Human-aware robotic assistant for collaborative assembly: Integrating human motion prediction with planning in time,''
\newblock {\em IEEE Robotics and Automation Letters}, vol. 3, no. 3, pp. 2394--2401, 2018.

\bibitem{xu2015learning}
Dan Xu, Elisa Ricci, Yan Yan, Jingkuan Song, and Nicu Sebe,
\newblock ``Learning deep representations of appearance and motion for anomalous event detection,''
\newblock {\em arXiv preprint arXiv:1510.01553}, 2015.

\bibitem{chen20203d}
Siheng Chen, Baoan Liu, Chen Feng, Carlos Vallespi-Gonzalez, and Carl Wellington,
\newblock ``3d point cloud processing and learning for autonomous driving: Impacting map creation, localization, and perception,''
\newblock {\em IEEE Signal Processing Magazine}, vol. 38, no. 1, pp. 68--86, 2020.

\bibitem{gopalakrishnan2019neural}
Anand Gopalakrishnan, Ankur Mali, Dan Kifer, Lee Giles, and Alexander~G Ororbia,
\newblock ``A neural temporal model for human motion prediction,''
\newblock in {\em Proceedings of the IEEE/CVF Conference on Computer Vision and Pattern Recognition}, 2019, pp. 12116--12125.

\bibitem{corona2020context}
Enric Corona, Albert Pumarola, Guillem Alenya, and Francesc Moreno-Noguer,
\newblock ``Context-aware human motion prediction,''
\newblock in {\em Proceedings of the IEEE/CVF Conference on Computer Vision and Pattern Recognition}, 2020, pp. 6992--7001.

\bibitem{lee2021video}
Sangmin Lee, Hak~Gu Kim, Dae~Hwi Choi, Hyung-Il Kim, and Yong~Man Ro,
\newblock ``Video prediction recalling long-term motion context via memory alignment learning,''
\newblock in {\em Proceedings of the IEEE/CVF Conference on Computer Vision and Pattern Recognition}, 2021, pp. 3054--3063.

\bibitem{mao2019learning}
Wei Mao, Miaomiao Liu, Mathieu Salzmann, and Hongdong Li,
\newblock ``Learning trajectory dependencies for human motion prediction,''
\newblock in {\em Proceedings of the IEEE/CVF International Conference on Computer Vision}, 2019, pp. 9489--9497.

\bibitem{li2022skeleton}
Maosen Li, Siheng Chen, Zijing Zhang, Lingxi Xie, Qi~Tian, and Ya~Zhang,
\newblock ``Skeleton-parted graph scattering networks for 3d human motion prediction,''
\newblock in {\em Computer Vision--ECCV 2022: 17th European Conference, Tel Aviv, Israel, October 23--27, 2022, Proceedings, Part VI}. Springer, 2022, pp. 18--36.

\bibitem{xu2023eqmotion}
Chenxin Xu, Robby~T Tan, Yuhong Tan, Siheng Chen, Yu~Guang Wang, Xinchao Wang, and Yanfeng Wang,
\newblock ``Eqmotion: Equivariant multi-agent motion prediction with invariant interaction reasoning,''
\newblock in {\em Proceedings of the IEEE/CVF Conference on Computer Vision and Pattern Recognition}, 2023, pp. 1410--1420.

\bibitem{xu2023auxiliary}
Chenxin Xu, Robby~T Tan, Yuhong Tan, Siheng Chen, Xinchao Wang, and Yanfeng Wang,
\newblock ``Auxiliary tasks benefit 3d skeleton-based human motion prediction,''
\newblock in {\em Proceedings of the IEEE/CVF international conference on computer vision}, 2023, pp. 9509--9520.

\bibitem{ionescu2013human3}
Catalin Ionescu, Dragos Papava, Vlad Olaru, and Cristian Sminchisescu,
\newblock ``Human3. 6m: Large scale datasets and predictive methods for 3d human sensing in natural environments,''
\newblock {\em IEEE transactions on pattern analysis and machine intelligence}, vol. 36, no. 7, pp. 1325--1339, 2013.

\bibitem{sofianos2021space}
Theodoros Sofianos, Alessio Sampieri, Luca Franco, and Fabio Galasso,
\newblock ``Space-time-separable graph convolutional network for pose forecasting,''
\newblock in {\em Proceedings of the IEEE/CVF International Conference on Computer Vision}, 2021, pp. 11209--11218.

\bibitem{eigen2013learning}
David Eigen, Marc'Aurelio Ranzato, and Ilya Sutskever,
\newblock ``Learning factored representations in a deep mixture of experts,''
\newblock {\em arXiv preprint arXiv:1312.4314}, 2013.

\bibitem{shazeer2017outrageously}
Noam Shazeer, Azalia Mirhoseini, Krzysztof Maziarz, Andy Davis, Quoc Le, Geoffrey Hinton, and Jeff Dean,
\newblock ``Outrageously large neural networks: The sparsely-gated mixture-of-experts layer,''
\newblock {\em arXiv preprint arXiv:1701.06538}, 2017.

\bibitem{wang2023spatio}
Jiexin Wang, Yujie Zhou, Wenwen Qiang, Ying Ba, Bing Su, and Ji-Rong Wen,
\newblock ``Spatio-temporal branching for motion prediction using motion increments,''
\newblock {\em arXiv preprint arXiv:2308.01097}, 2023.

\bibitem{van2008visualizing}
Laurens Van~der Maaten and Geoffrey Hinton,
\newblock ``Visualizing data using t-sne.,''
\newblock {\em Journal of machine learning research}, vol. 9, no. 11, 2008.

\bibitem{martinez2017human}
Julieta Martinez, Michael~J Black, and Javier Romero,
\newblock ``On human motion prediction using recurrent neural networks,''
\newblock in {\em Proceedings of the IEEE conference on computer vision and pattern recognition}, 2017, pp. 2891--2900.

\bibitem{li2020dynamic}
Maosen Li, Siheng Chen, Yangheng Zhao, Ya~Zhang, Yanfeng Wang, and Qi~Tian,
\newblock ``Dynamic multiscale graph neural networks for 3d skeleton based human motion prediction,''
\newblock in {\em Proceedings of the IEEE/CVF conference on computer vision and pattern recognition}, 2020, pp. 214--223.

\bibitem{dang2021msr}
Lingwei Dang, Yongwei Nie, Chengjiang Long, Qing Zhang, and Guiqing Li,
\newblock ``Msr-gcn: Multi-scale residual graph convolution networks for human motion prediction,''
\newblock in {\em Proceedings of the IEEE/CVF International Conference on Computer Vision}, 2021, pp. 11467--11476.

\bibitem{vonMarcard2018}
Timo von Marcard, Roberto Henschel, Michael Black, Bodo Rosenhahn, and Gerard Pons-Moll,
\newblock ``Recovering accurate 3d human pose in the wild using imus and a moving camera,''
\newblock in {\em European Conference on Computer Vision (ECCV)}, sep 2018.

\bibitem{paszke2019pytorch}
Adam Paszke, Sam Gross, Francisco Massa, Adam Lerer, James Bradbury, Gregory Chanan, Trevor Killeen, Zeming Lin, Natalia Gimelshein, Luca Antiga, et~al.,
\newblock ``Pytorch: An imperative style, high-performance deep learning library,''
\newblock {\em Advances in neural information processing systems}, vol. 32, 2019.

\end{thebibliography}

\clearpage
\appendix

We report more experimental results and more technical details which are not included in the paper.


\subsection{Datasets}
\textbf{Human 3.6M (H3.6M).}  H3.6M~\cite{ionescu2013human3} consists of motion capture data of 11 actors performing 15 common actions, such as walking, running, and sitting. 
Following the standard paradigm~\cite{martinez2017human,li2020dynamic}, we train the models on the segmented clips in the 6 subjects and test the model on the specific clips in the 5th subject.

\textbf{CMU Motion Capture (CMU-Mocap).} CMU-Mocap, which contains motion capture data of various human activities, has been widely used in various applications. Following the evaluation protocol used in previous works~\cite{dang2021msr,li2022skeleton}, we evaluate the performance on 8 human action categories.

\textbf{3D Pose in the Wild (3DPW).} 3DPW ~\cite{vonMarcard2018} 
is the first dataset in the wild to provide accurate 3D pose annotations for evaluation. It contains more than 51,000 frames for challenging indoor and outdoor activities, and official organizers have defined data splits for training, validation, and testing, along with evaluation protocols. 

\subsection{Implementation Details}
Our experiment is implemented in PyTorch~\cite{paszke2019pytorch} and trained on a single NVIDIA A40 GPU. During our experiment, we follow the training setup of the original method without any modification. This approach allows us to accurately assess the performance of our proposed method. 
Since EqMotion was not evaluated on the CMU-Mocap and 3DPW datasets, we conducted experiments on H3.6M only for this specific baseline. Besides, EqMotion does not use outer residual connection and padding operation for input, therefore our method does not reconstruct historical information in EqMotion, and the remaining design remains consistent. 

\subsection{Advantages of Our Approach}
Compared to existing methods, our approach offers several advantages. Firstly, we harness multi-range decoupling decoding to synthesize motion representations by exploiting temporal correlations effectively. This not only refines learned features but also captures intricate dependencies spanning various future prediction horizons. Secondly, we introduce a gating-adjusting aggregation mechanism for the outputs of multi-range decoupling decoding, guided by the motion data. 
This adaptability ensures that the model dynamically combines diverse insights, contributing to improved prediction accuracy. Thirdly, the simplicity of our method facilitates easy integration into various prediction methods.
These strengths position our approach as a promising direction and easily integrable alternative for HMP.

\subsection{More Quantitative Results}
Here, we present one set of results from five experiments, showing the prediction results on all the contained actions in H3.6M and CMU Mocap for both short-term and long-term prediction. 
These results provide sufficient information for a detailed comparison of the algorithm development in future works.

First, we present the MPJPE of various models on H3.6M for short-term motion prediction, where the detailed results of any actions are shown in~\tablename~\ref{tab:h3.6M—shortterm-appendix}. We see that the proposed method multi-range decoupling decoding with gating-adjusting aggregation could effectively enhance the prediction performance of baselines at most actions, as well as achieve better results in terms of the average prediction errors. 
Also, we show the prediction MPJPEs on H3.6M for long-term motion prediction, which are presented in~\tablename~\ref{tab:h3.6M—longterm-appendix}. We see that our proposed method obtains an effective performance in long-term motion prediction, since those baselines with our method show lower MPJPEs on most actions as well as the average MPJPE over all actions.

Finally, we present the MPJPEs of various methods on all the actions of CMU Mocap in~\tablename~\ref{tab:CMU-Mocap-appendix} . The experiment results also verify the effectiveness of the proposed method.

\begin{figure*}[ht]
\centering
\subfigure[80ms.]{\includegraphics[width = 0.3\textwidth]{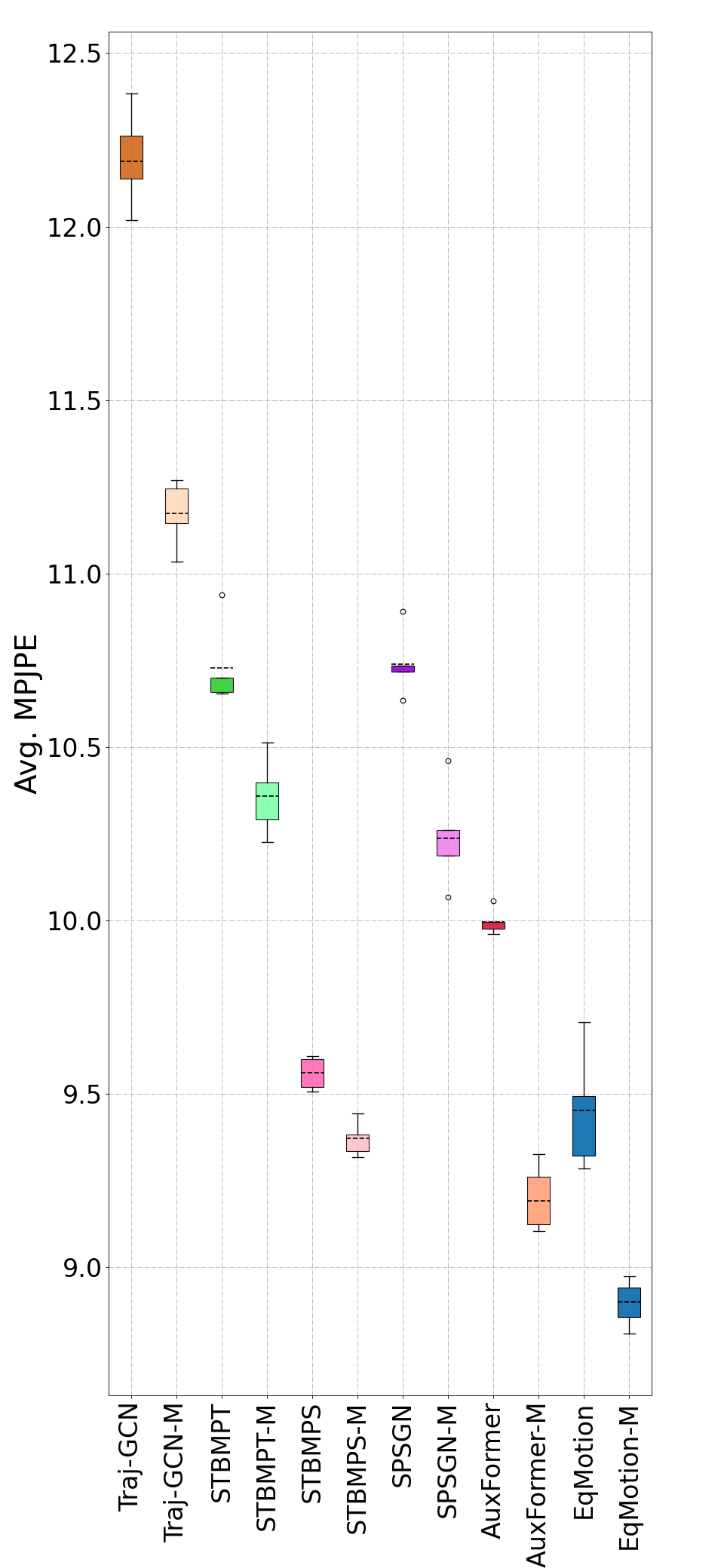}}
\subfigure[160ms.]{\includegraphics[width = 0.3\textwidth]{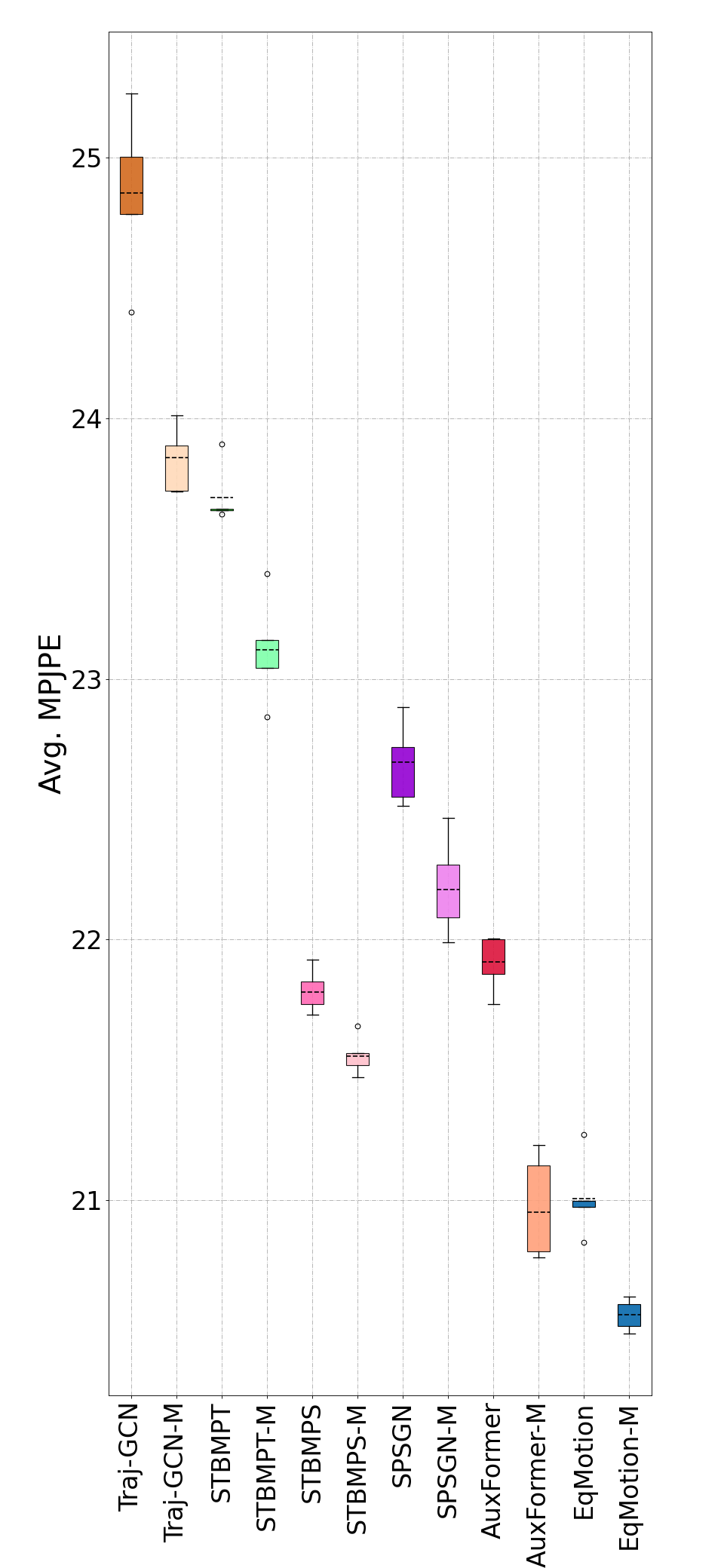}}
\subfigure[320ms.]{\includegraphics[width = 0.3\textwidth]{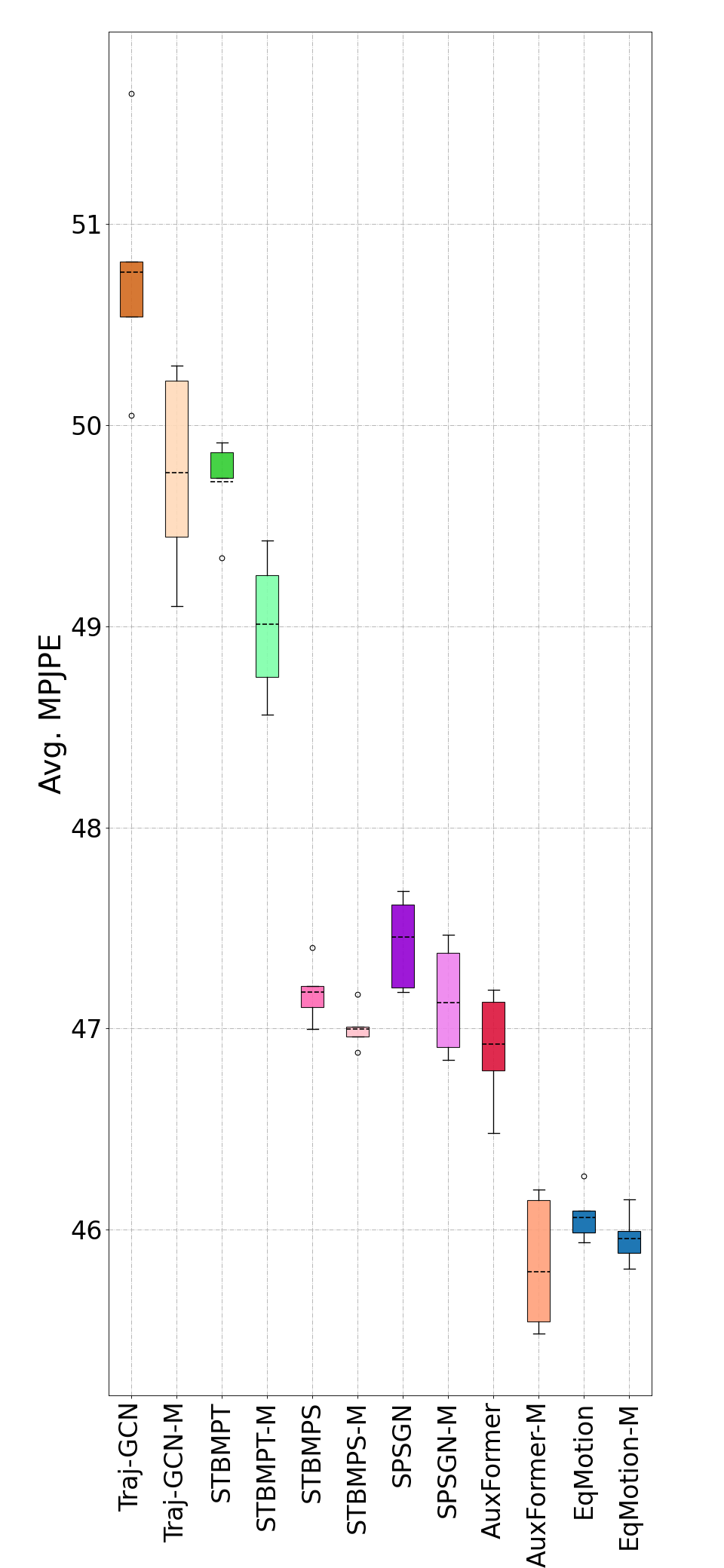}}\\
\subfigure[480ms.]{\includegraphics[width = 0.3\textwidth]{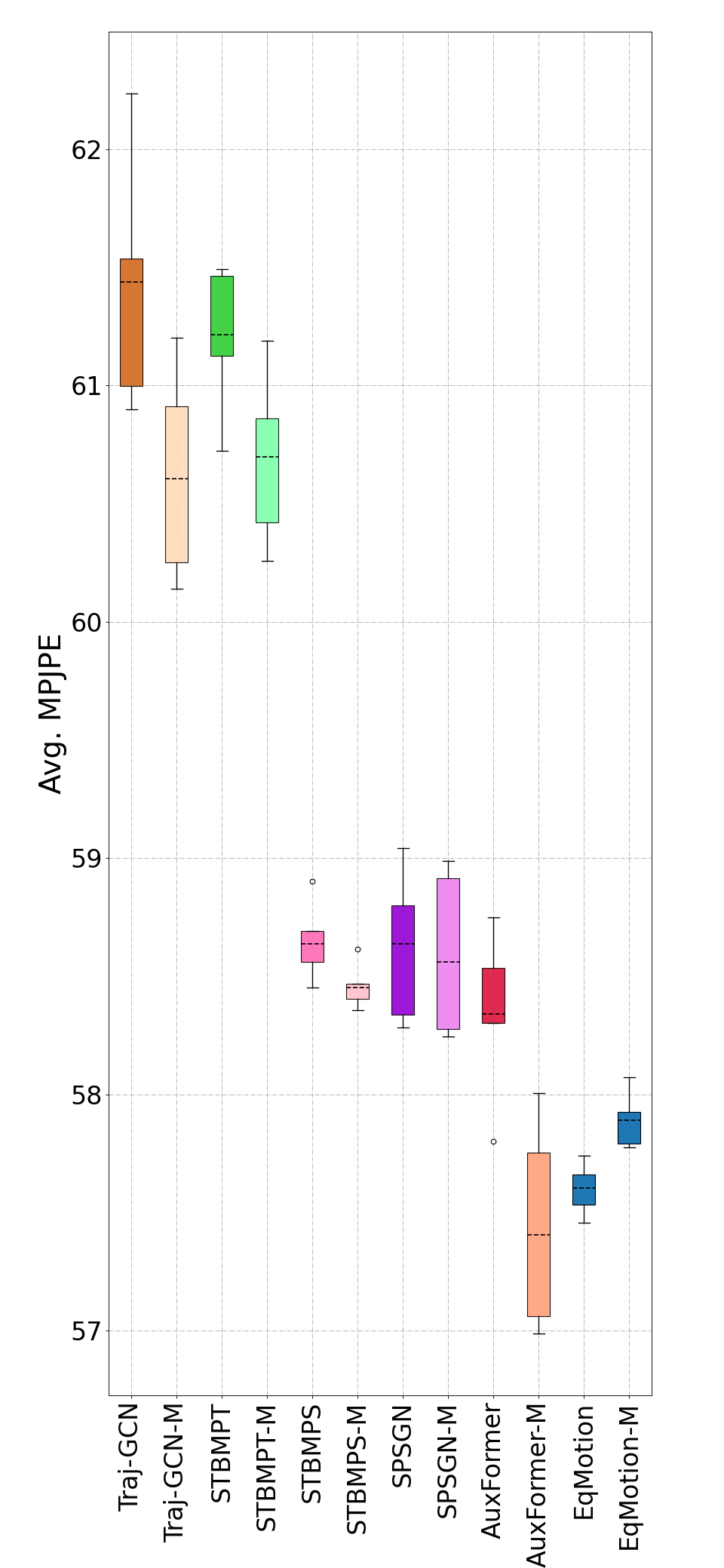}}
\subfigure[560ms.]{\includegraphics[width = 0.3\textwidth]{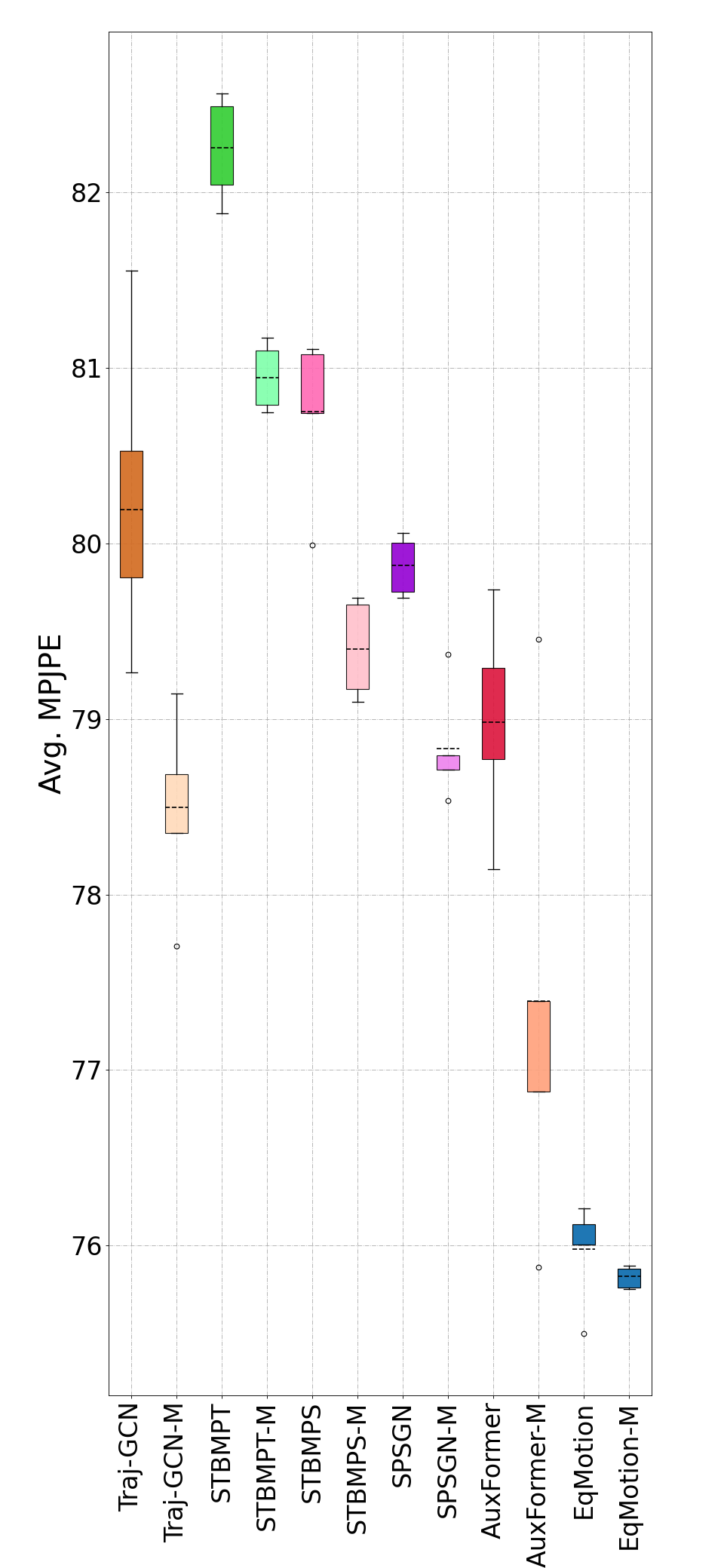}}
\subfigure[1000ms.]{\includegraphics[width = 0.3\textwidth]{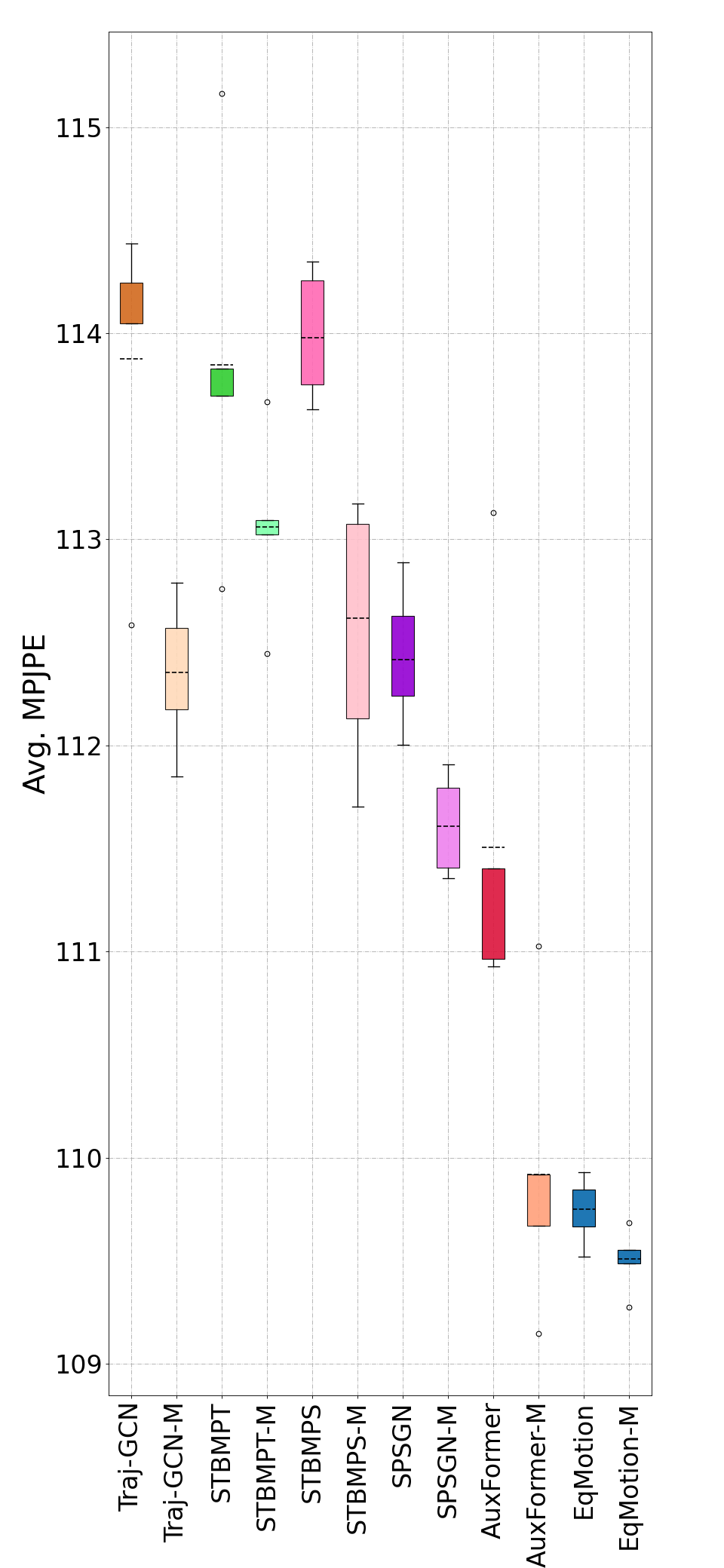}}
\caption{Performance comparison of prediction on H3.6M under various methods.}
\label{fig:rand_seed_h36m}
\end{figure*}

\begin{figure*}[ht]
\centering
\subfigure[80ms.]{\includegraphics[width = 0.3\textwidth]{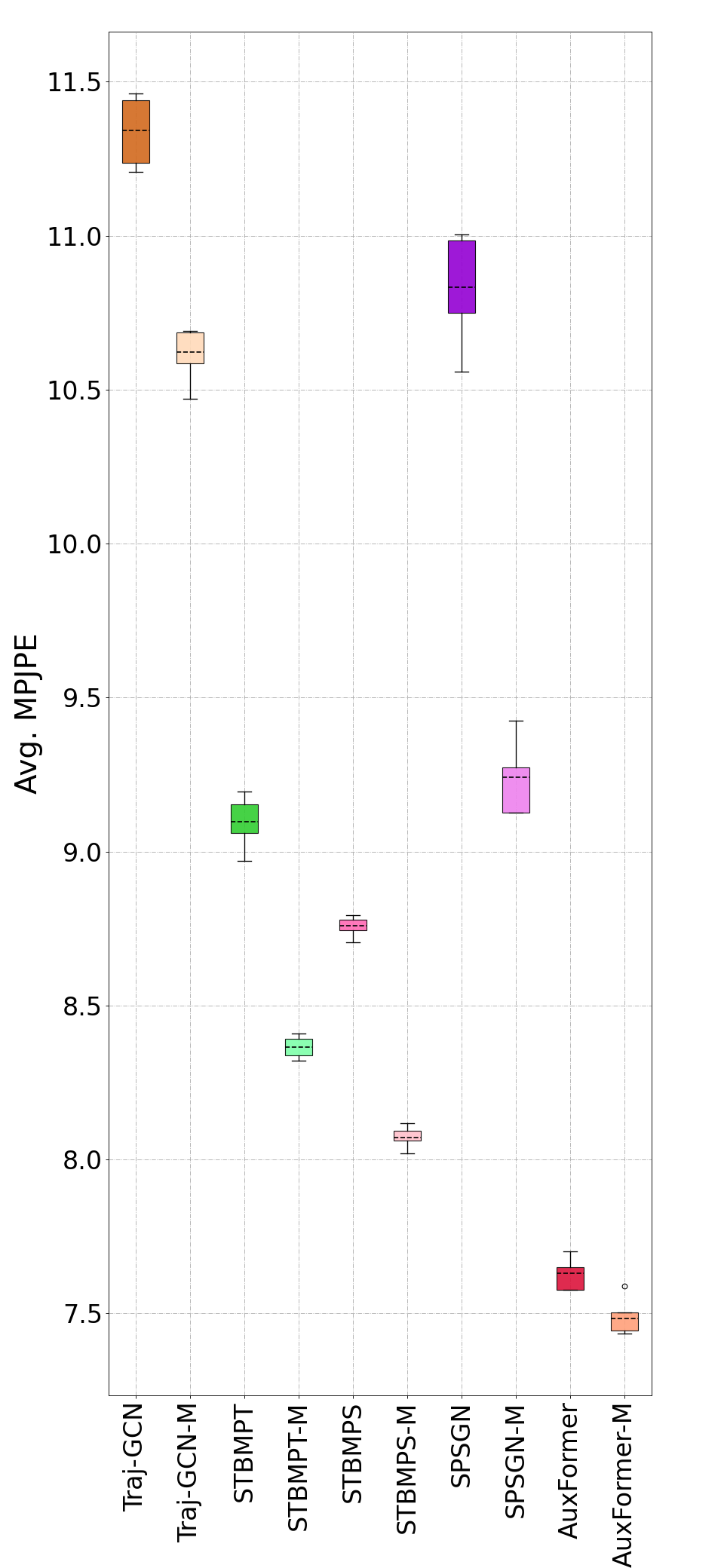}}
\subfigure[160ms.]{\includegraphics[width = 0.3\textwidth]{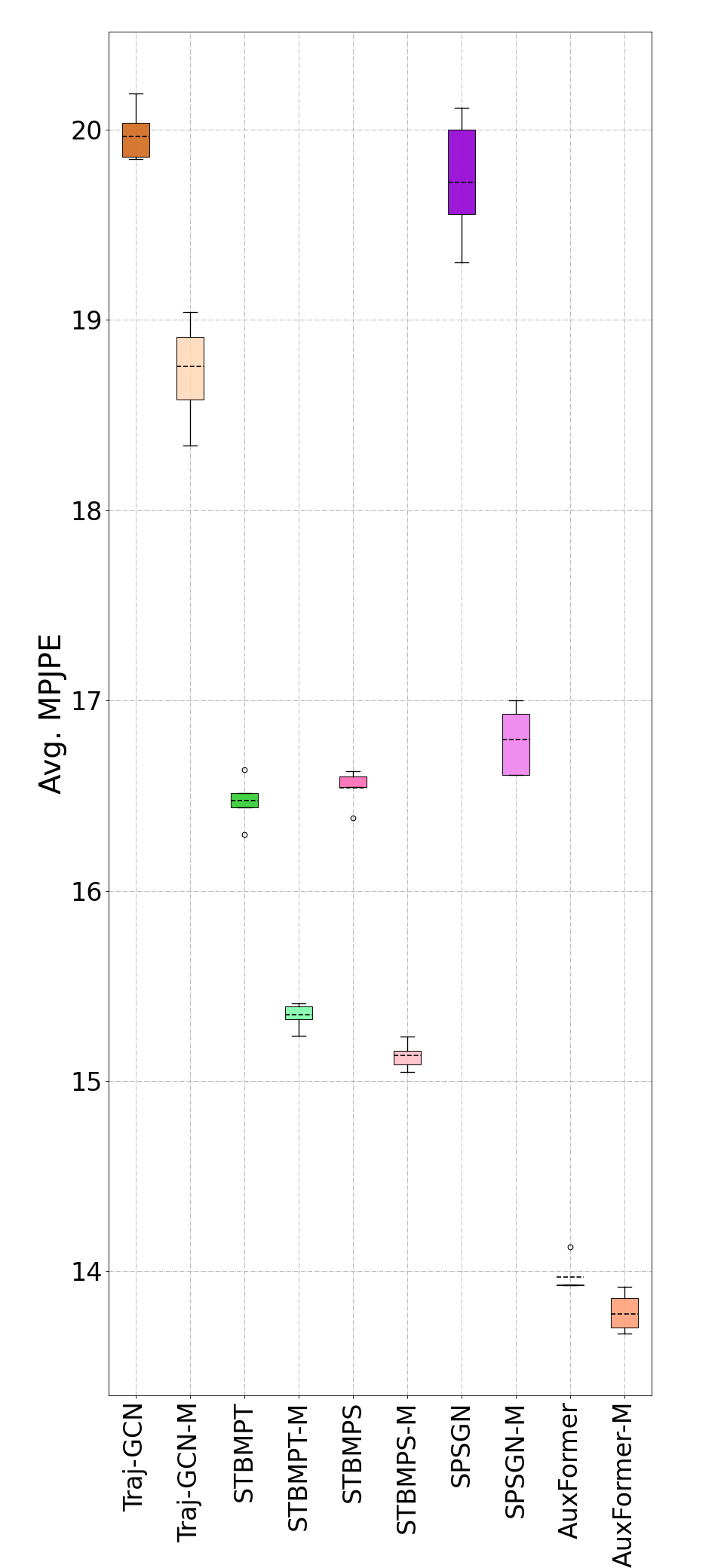}}
\subfigure[320ms.]{\includegraphics[width = 0.3\textwidth]{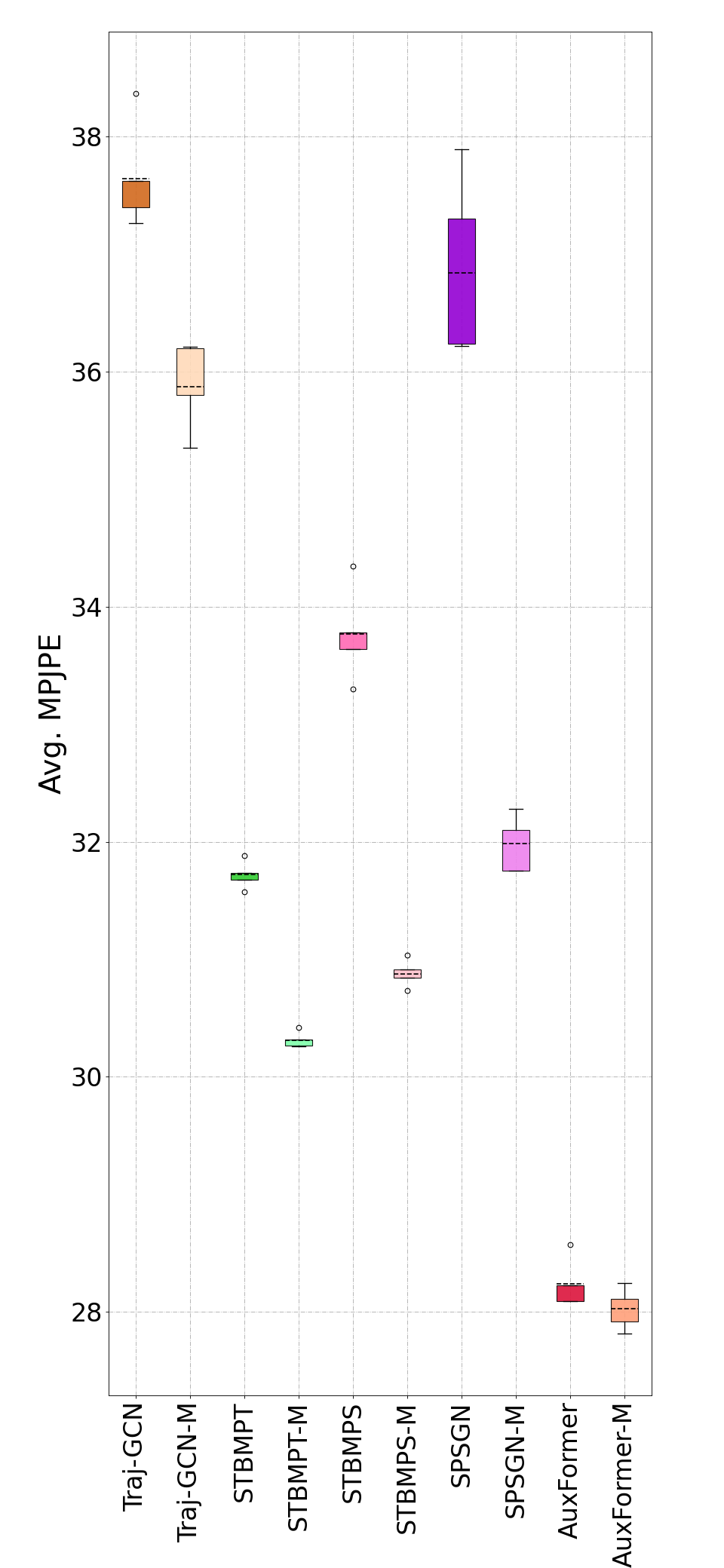}}\\
\subfigure[480ms.]{\includegraphics[width = 0.3\textwidth]{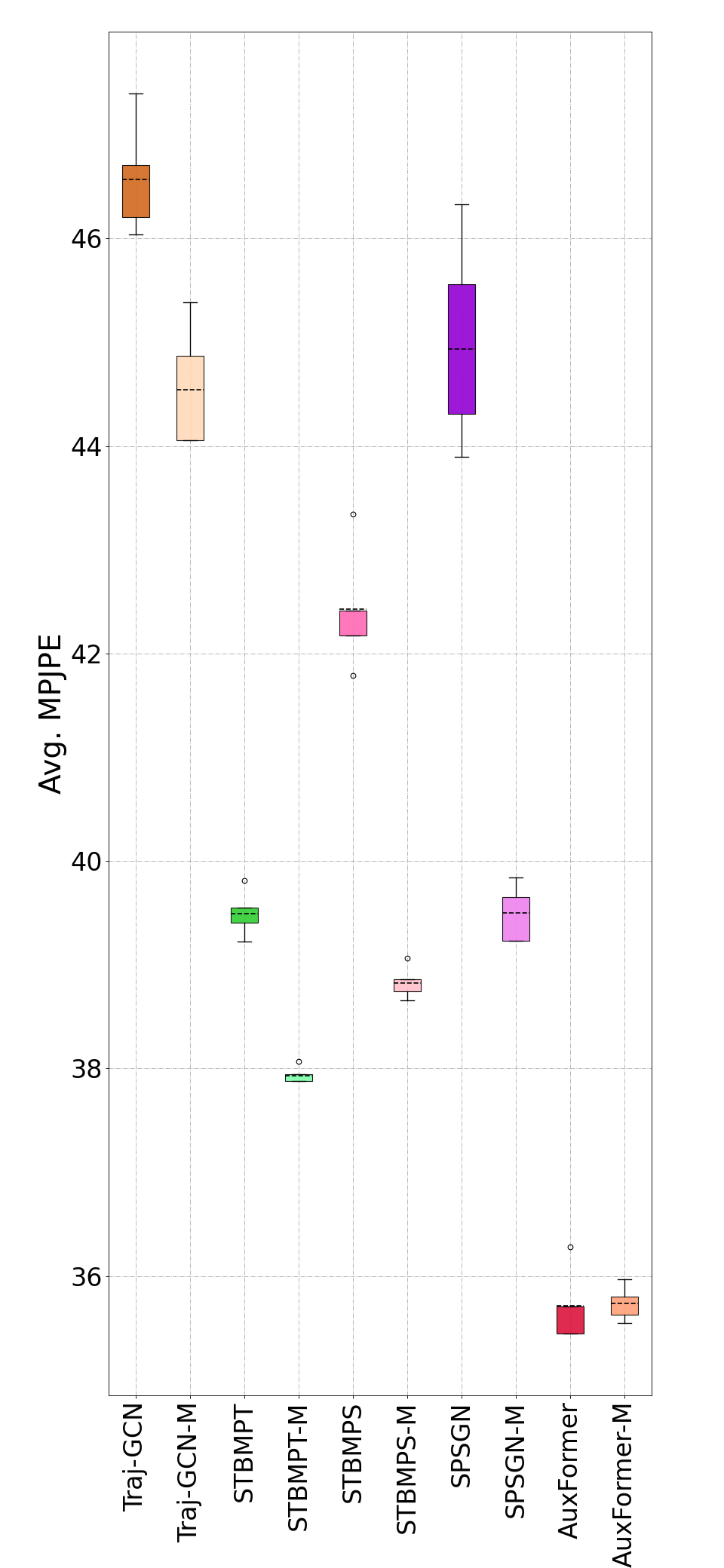}}
\subfigure[560ms.]{\includegraphics[width = 0.3\textwidth]{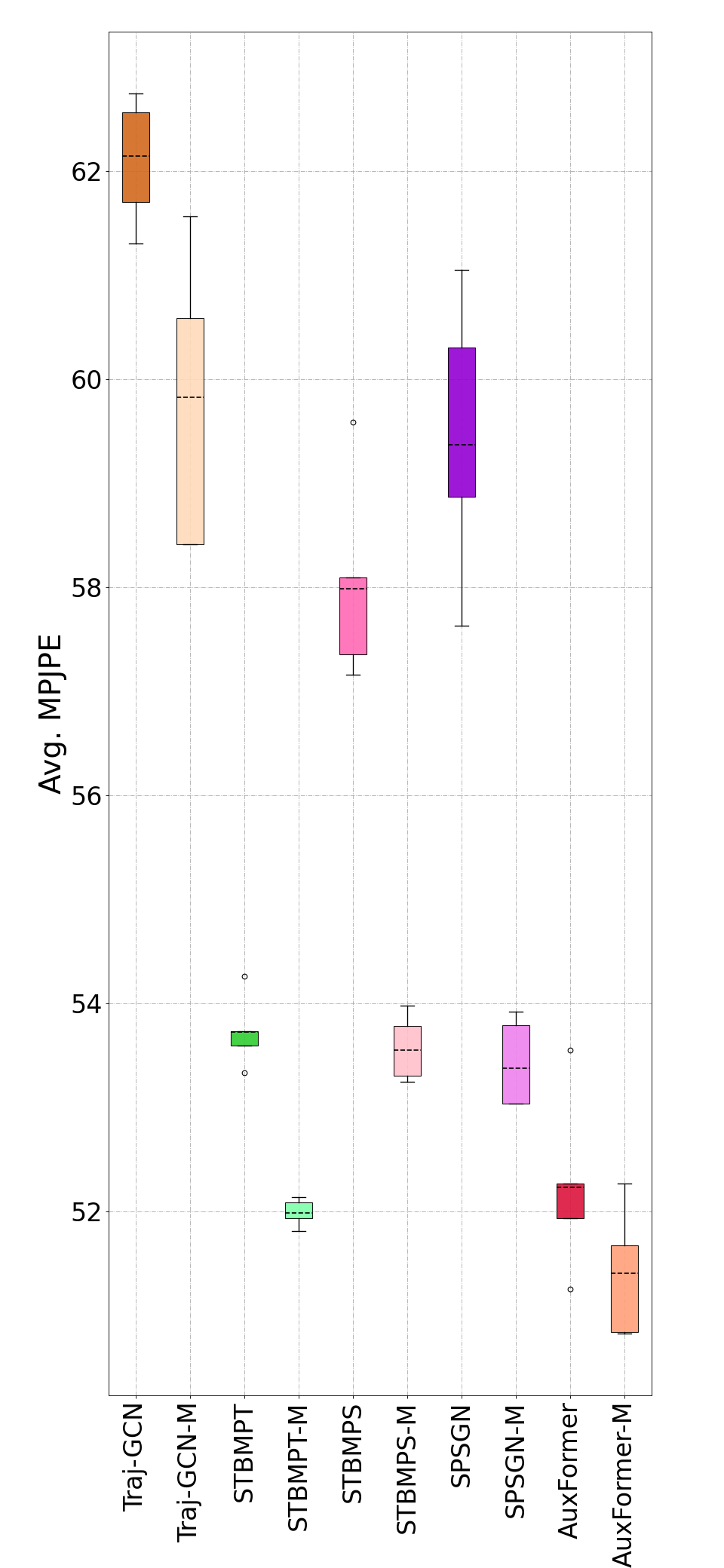}}
\subfigure[1000ms.]{\includegraphics[width = 0.3\textwidth]{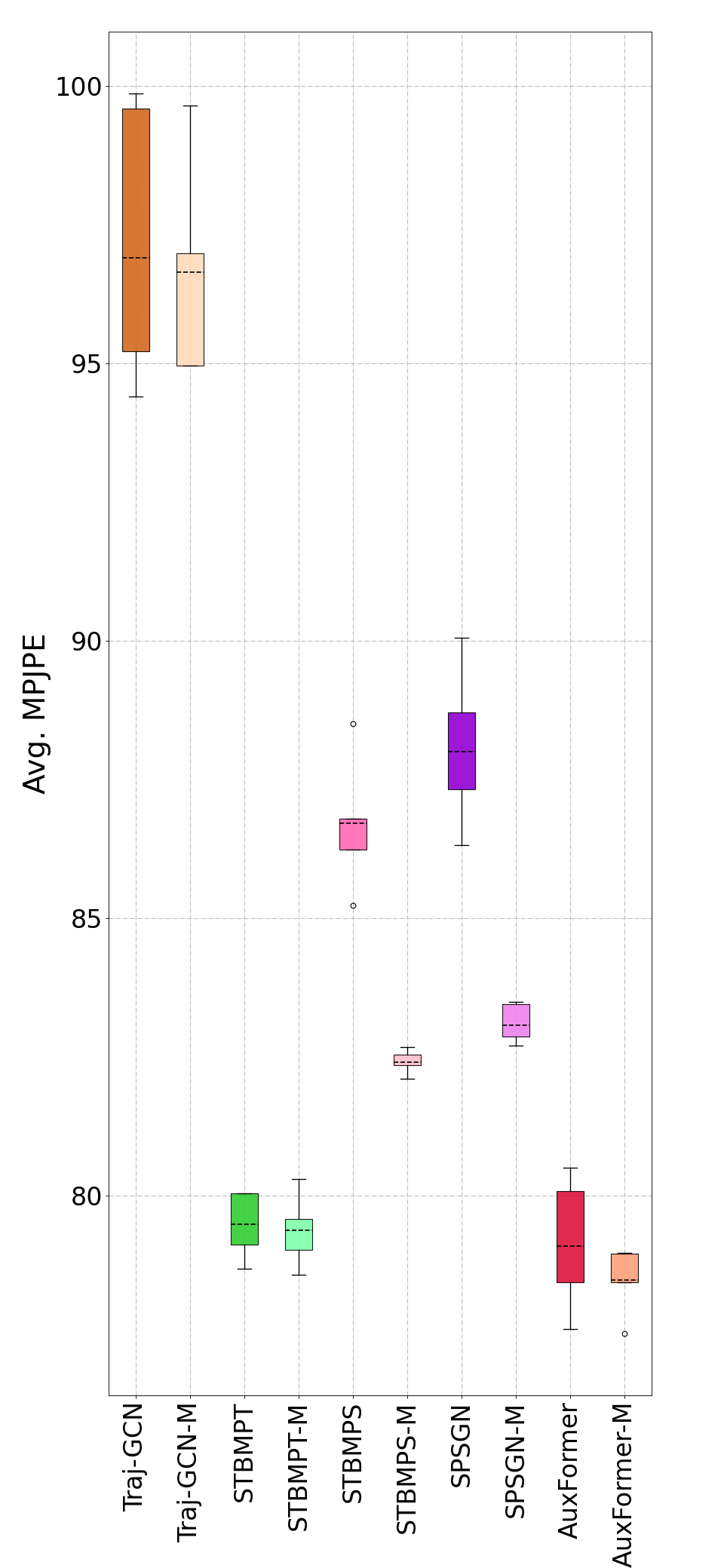}}
\caption{Performance comparison of prediction on CMU-Mocap under various methods.}
\label{fig:rand_seed_cmu}
\end{figure*}

\begin{figure*}[ht]
\centering
\subfigure[100ms.]{\includegraphics[width = 0.23\textwidth]{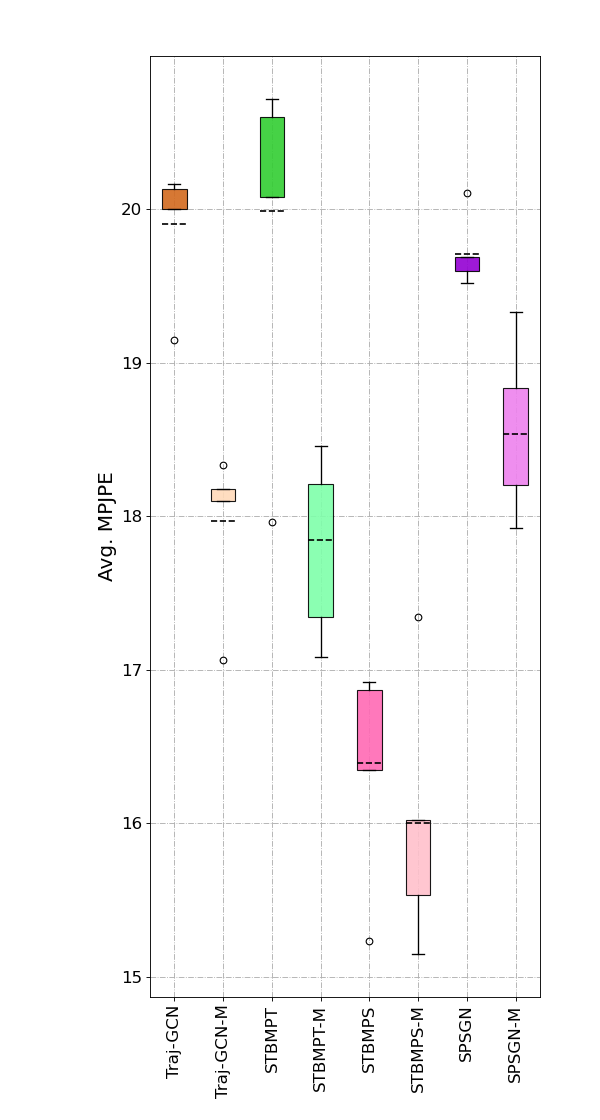}\label{fig:h36m_80}}
\subfigure[200ms.]{\includegraphics[width = 0.23\textwidth]{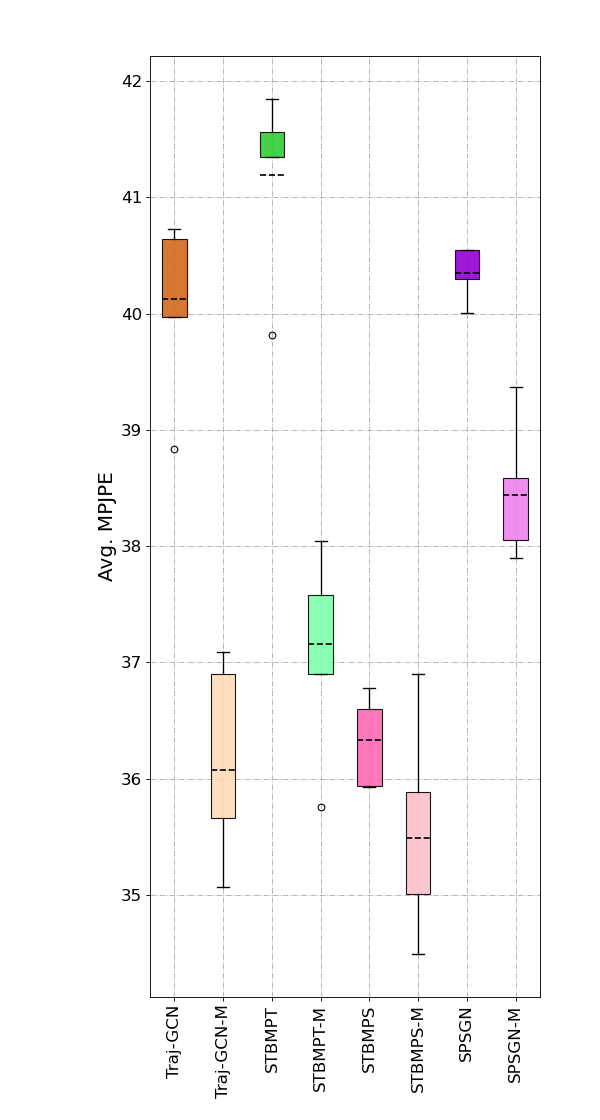}\label{fig:h36m_160}}
\subfigure[300ms.]{\includegraphics[width = 0.23\textwidth]{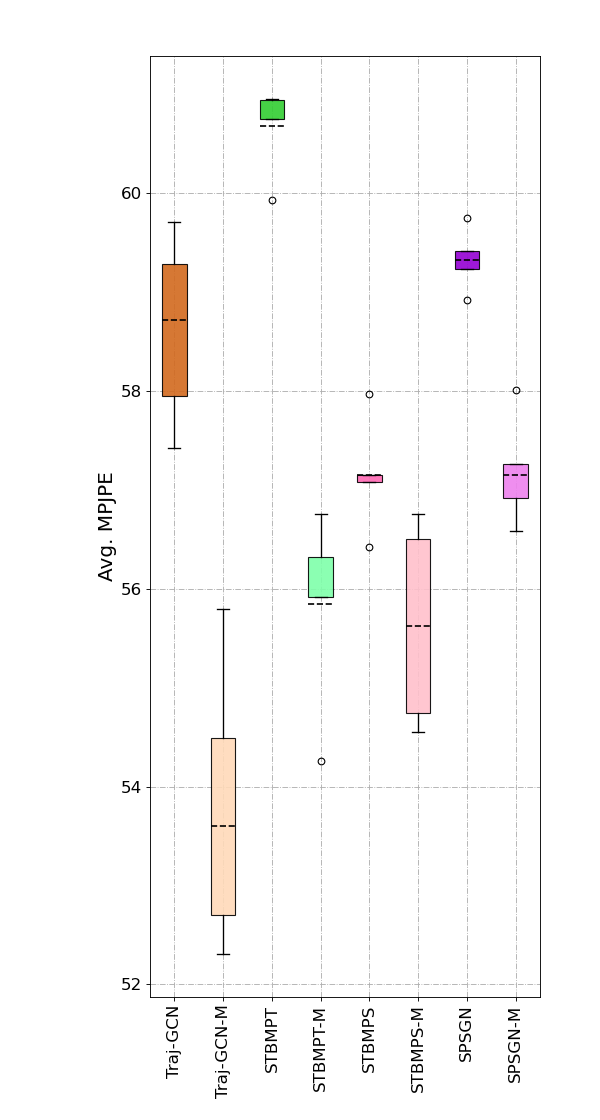}\label{fig:h36m_320}}
\subfigure[400ms.]{\includegraphics[width = 0.23\textwidth]{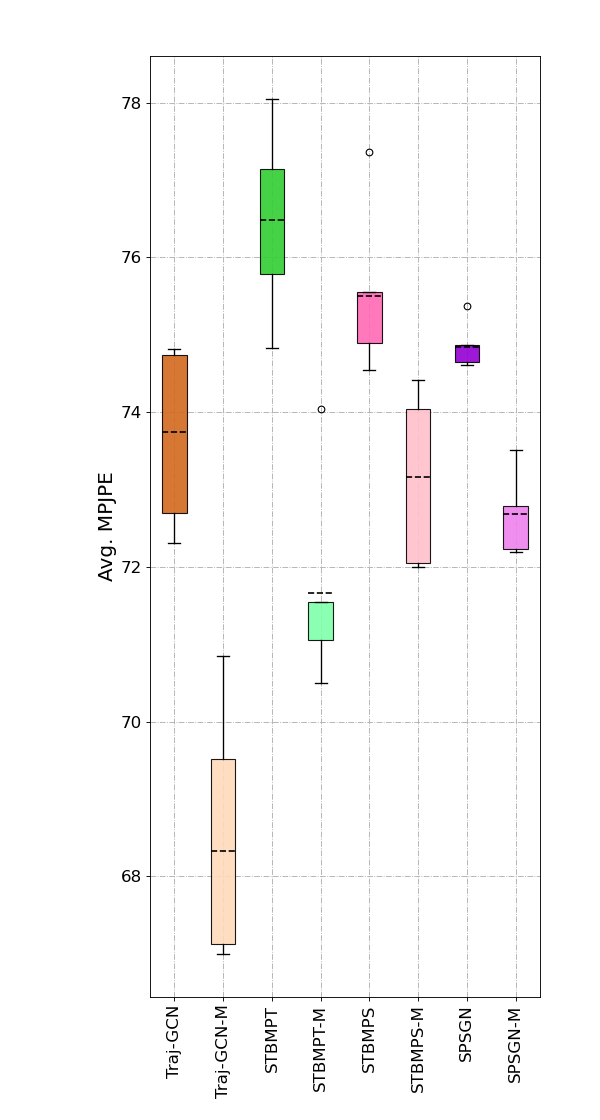}\label{fig:h36m_480}}
\subfigure[500ms.]{\includegraphics[width = 0.23\textwidth]{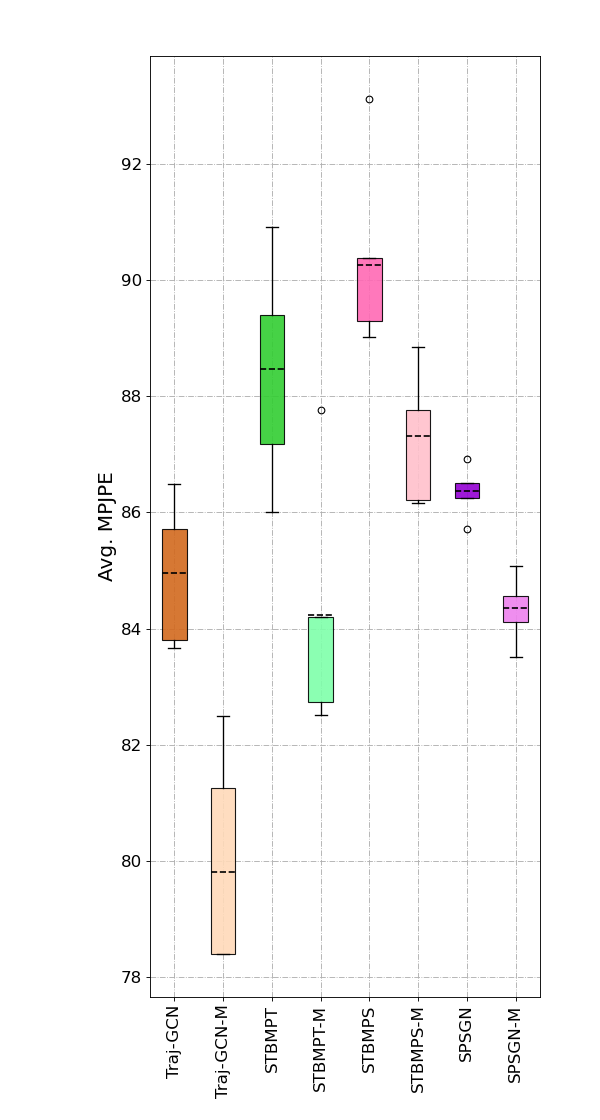}\label{fig:h36m_560}}
\subfigure[600ms.]{\includegraphics[width = 0.23\textwidth]{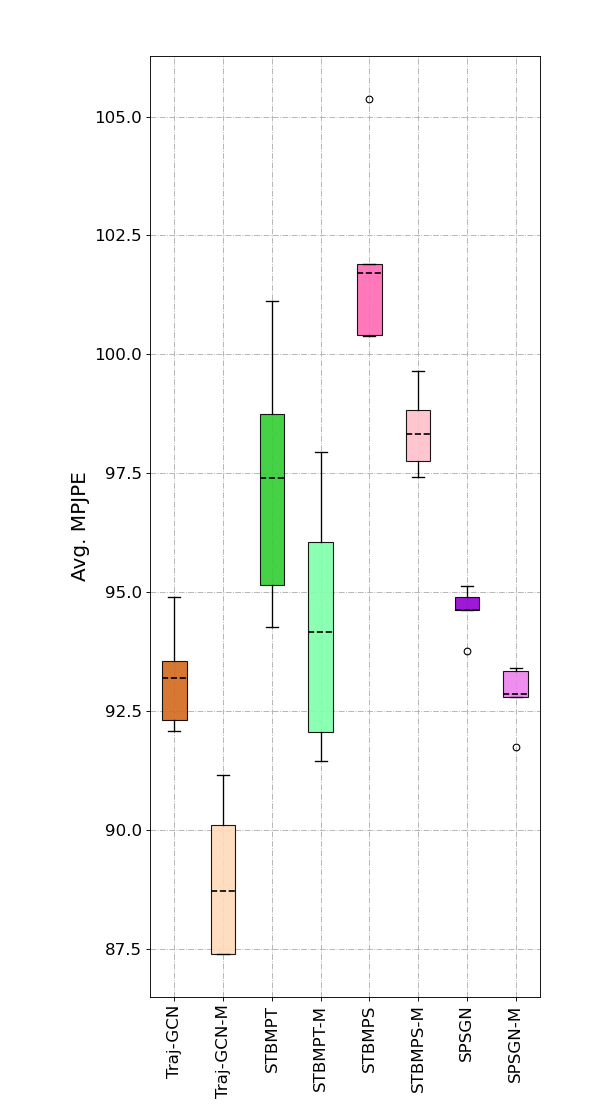}\label{fig:h36m_1000}}
\subfigure[700ms.]{\includegraphics[width = 0.23\textwidth]{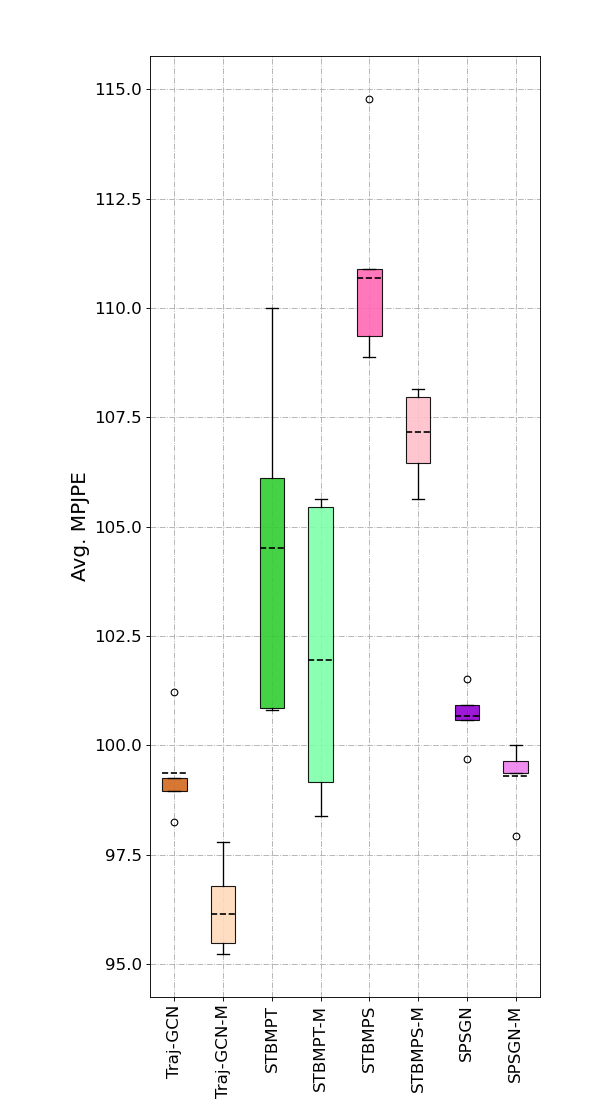}}
\subfigure[800ms.]{\includegraphics[width = 0.23\textwidth]{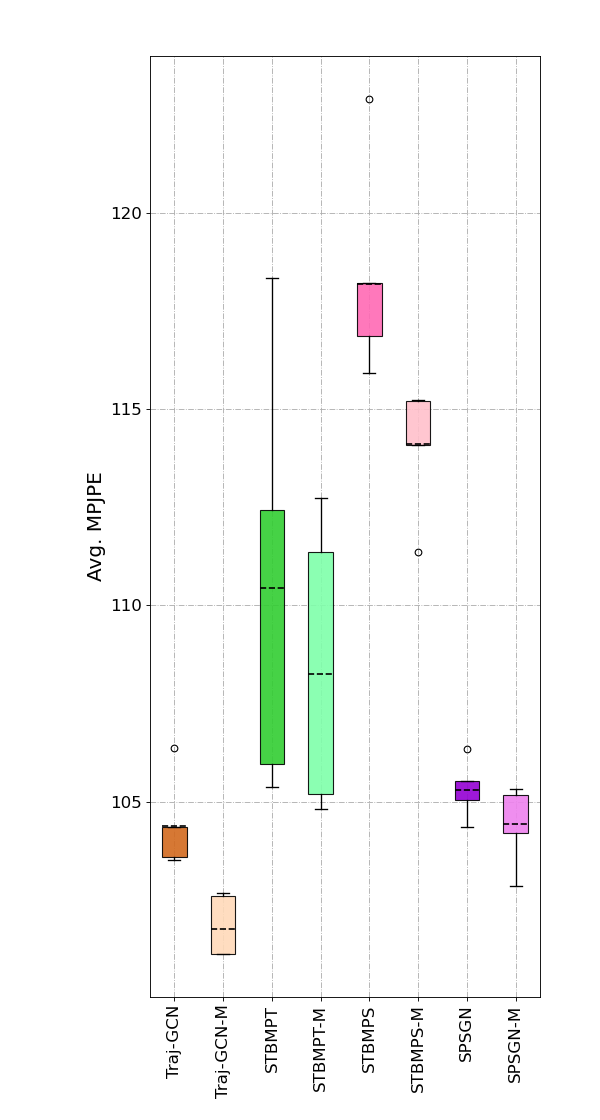}}
\subfigure[900ms.]{\includegraphics[width = 0.23\textwidth]{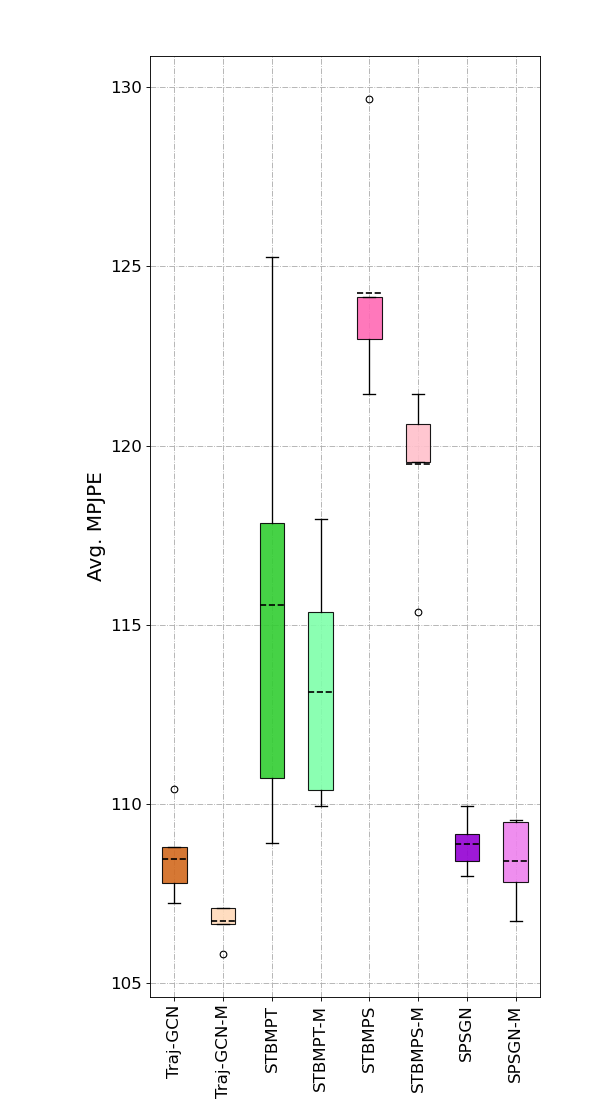}}
\subfigure[1000ms.]{\includegraphics[width = 0.23\textwidth]{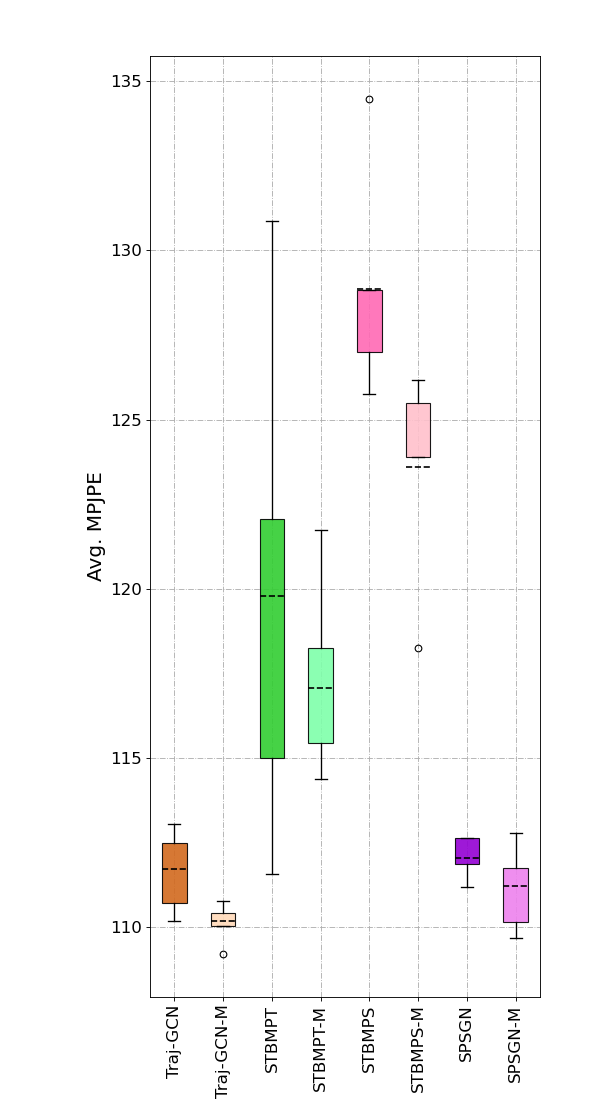}}
\subfigure[Average.]{\includegraphics[width = 0.23\textwidth]{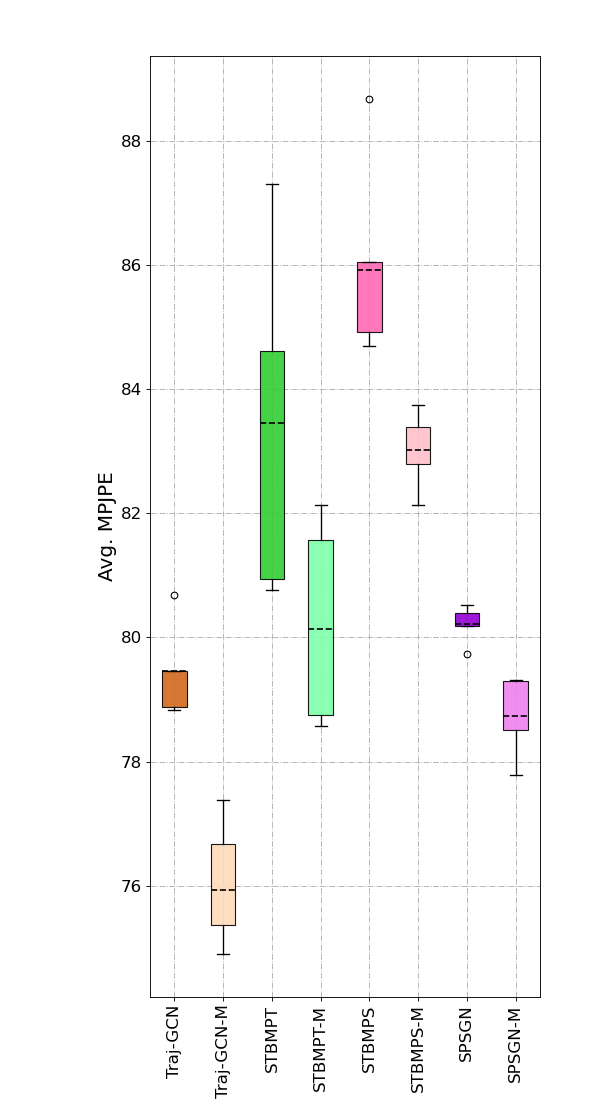}}
\caption{Performance comparison of prediction on 3DPW under various methods.}
\label{fig:rand_seed_3dpw}
\end{figure*}

For further comparison, we visualize the prediction results at each time step across 5 experiments for each method.  The results in \figurename{~\ref{fig:rand_seed_h36m}}, \figurename{~\ref{fig:rand_seed_cmu}}, and \figurename{~\ref{fig:rand_seed_3dpw}} show that the models with our method all outperform the performance of the original under various architectures, which highlights the superiority and stability of ours.

\subsection{More Qualitative Results}
\begin{figure*}[!t]
  \centering
    \subfigure[Sitting, waiting and directions in STBMPT.]{\includegraphics[width=0.33\linewidth]{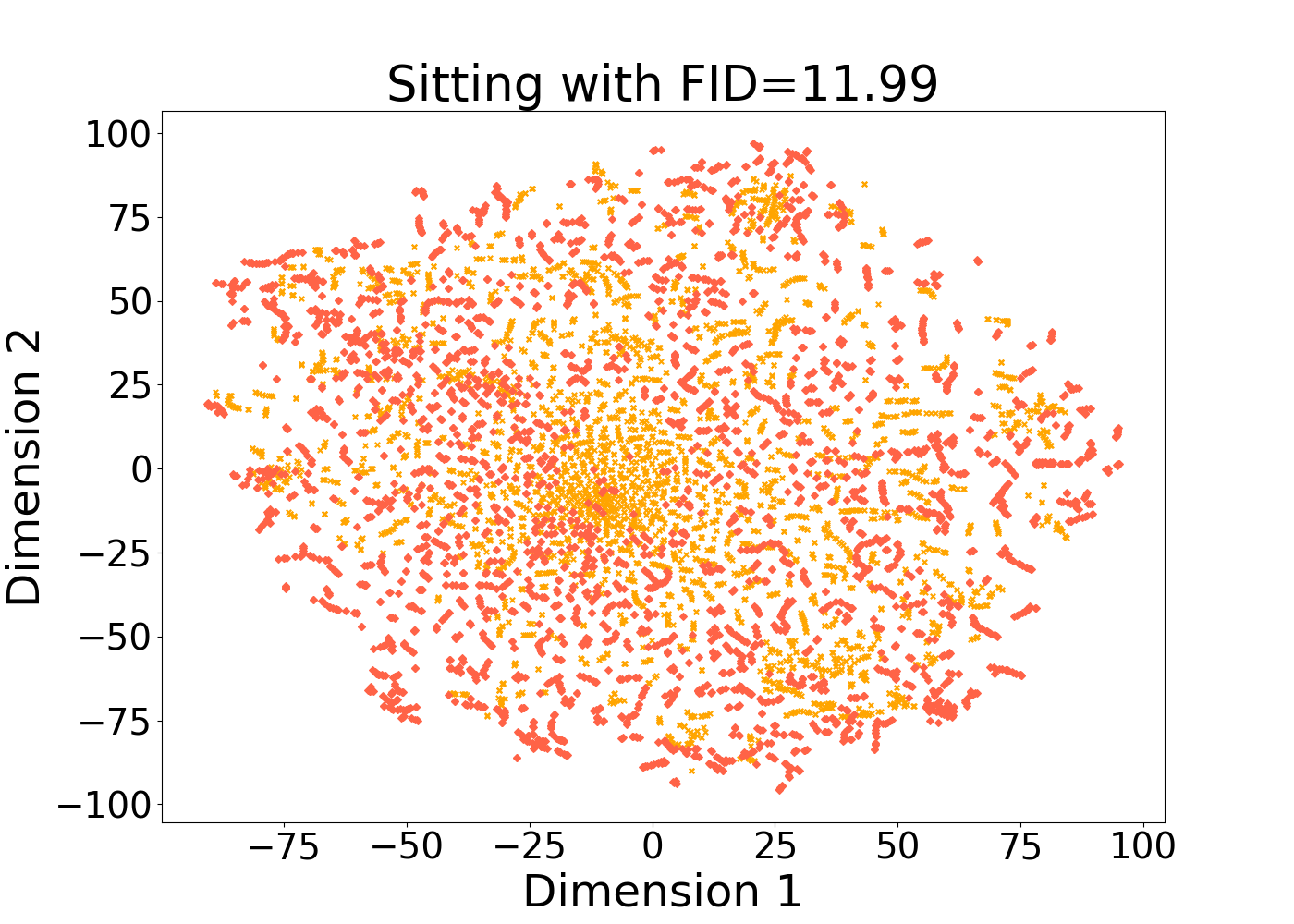}
    \includegraphics[width=0.33\linewidth]{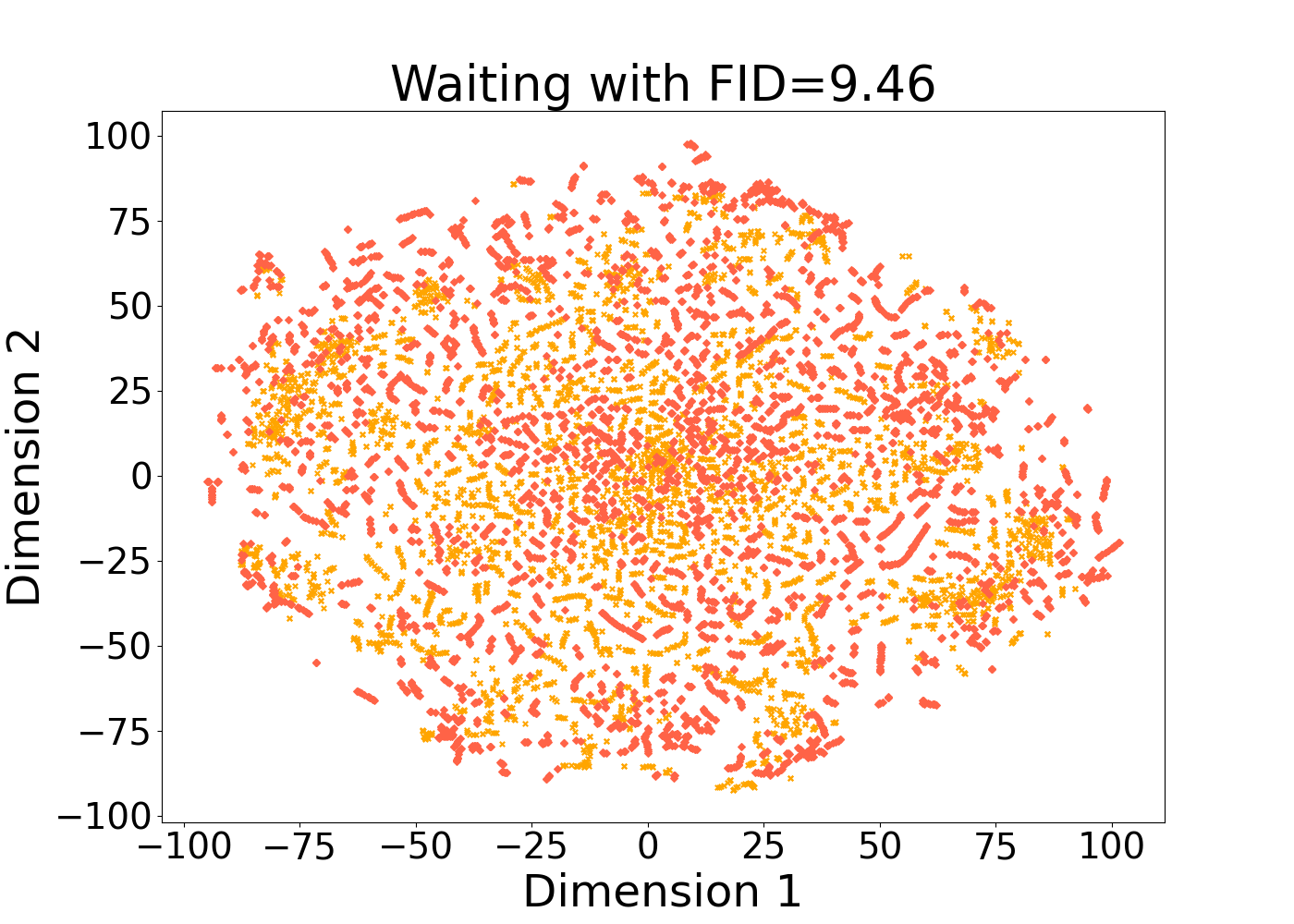}
    \includegraphics[width=0.33\linewidth]{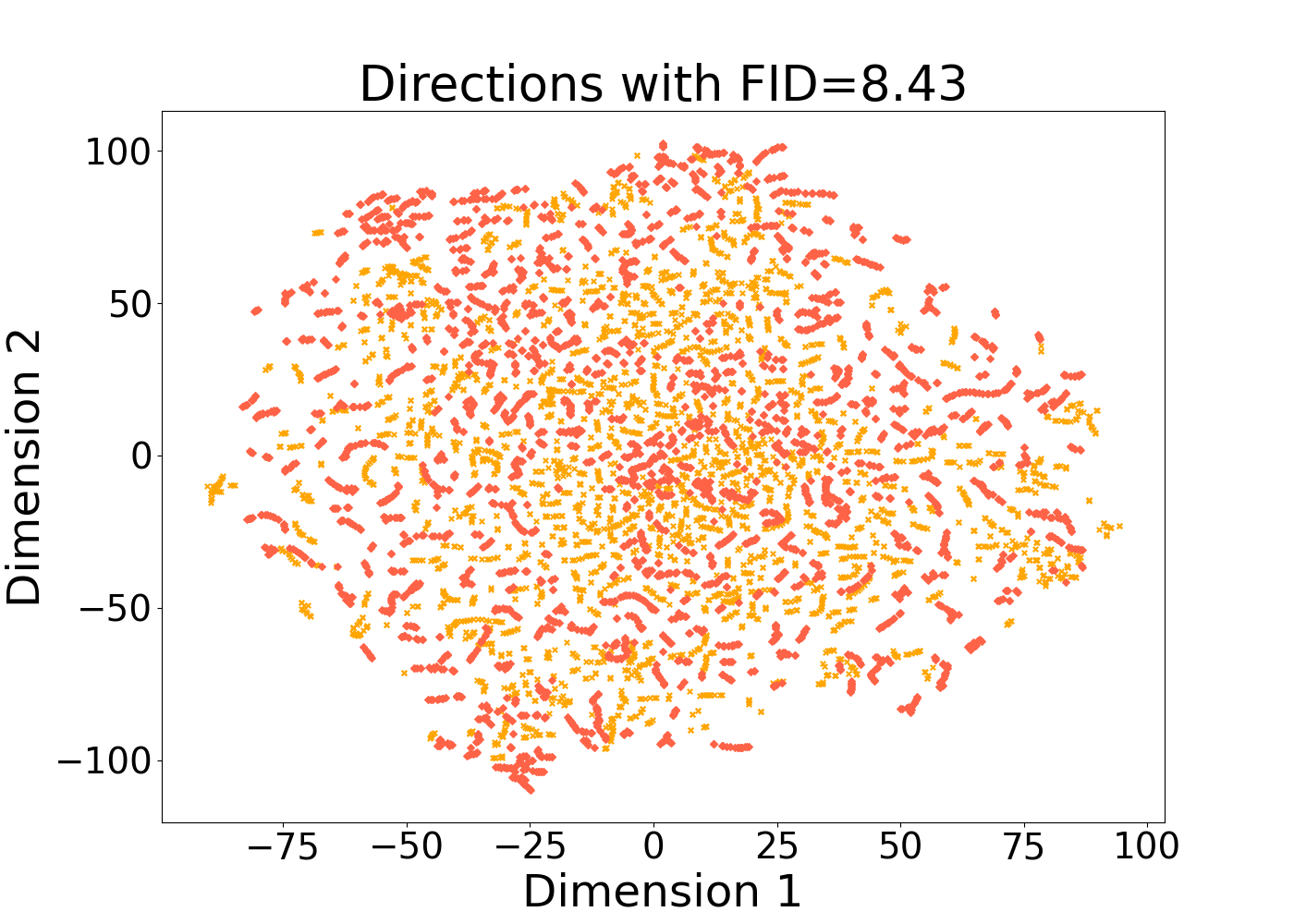}
    }
    \\
    \subfigure[Sitting, waiting and directions in STBMPT integrated with our method.]{
        \includegraphics[width=0.33\linewidth]{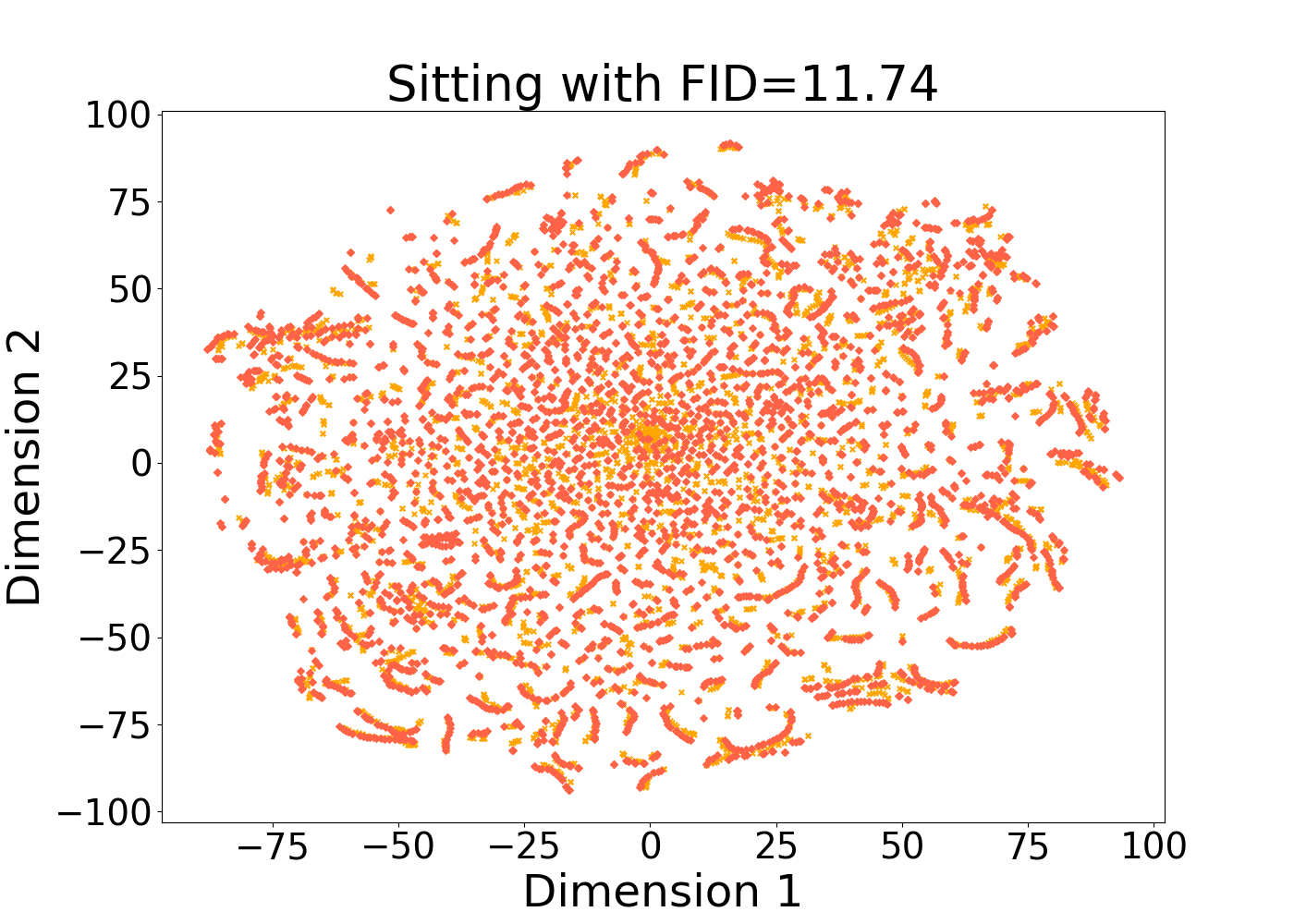}
        \includegraphics[width=0.33\linewidth]{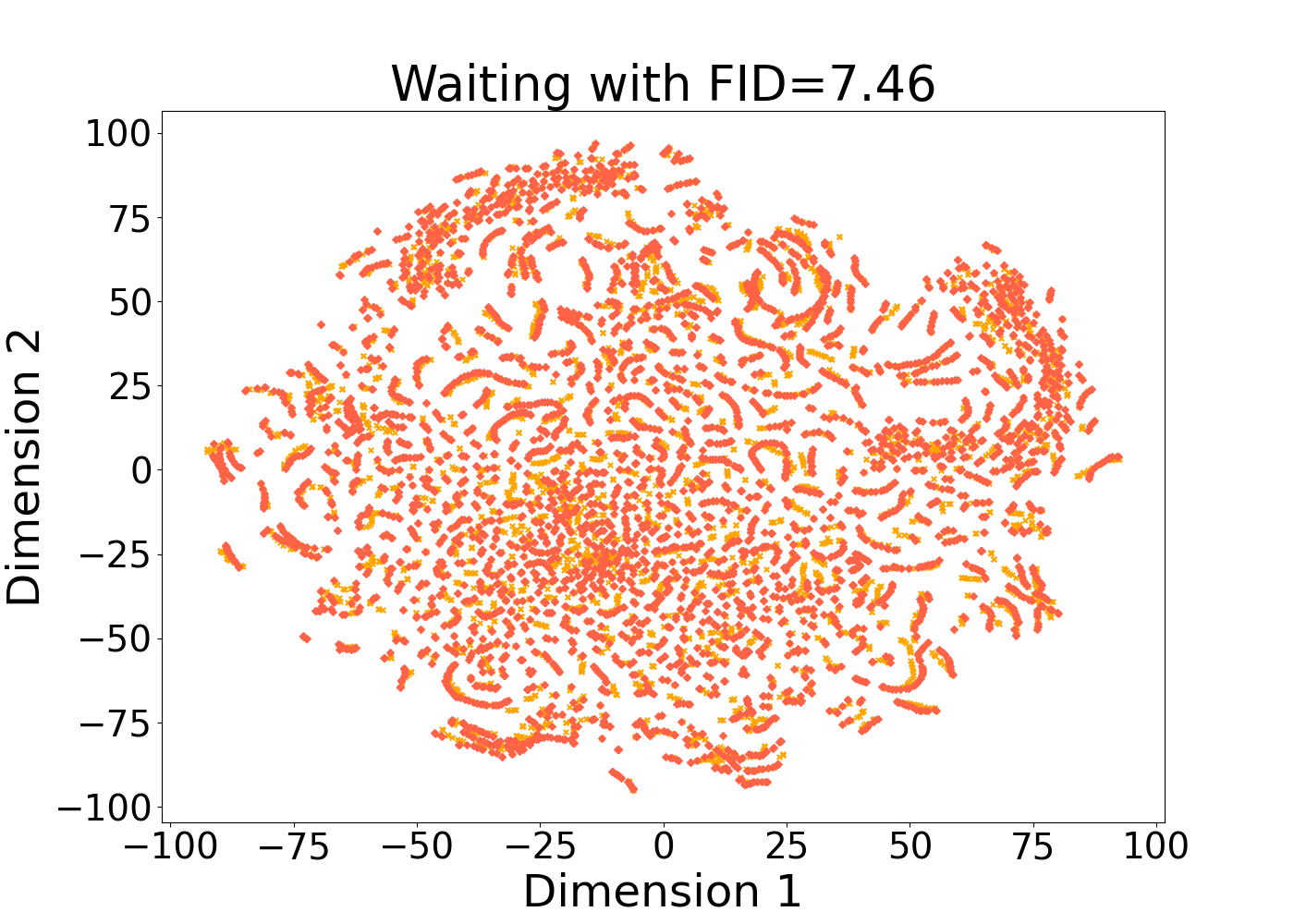}
        \includegraphics[width=0.33\linewidth]{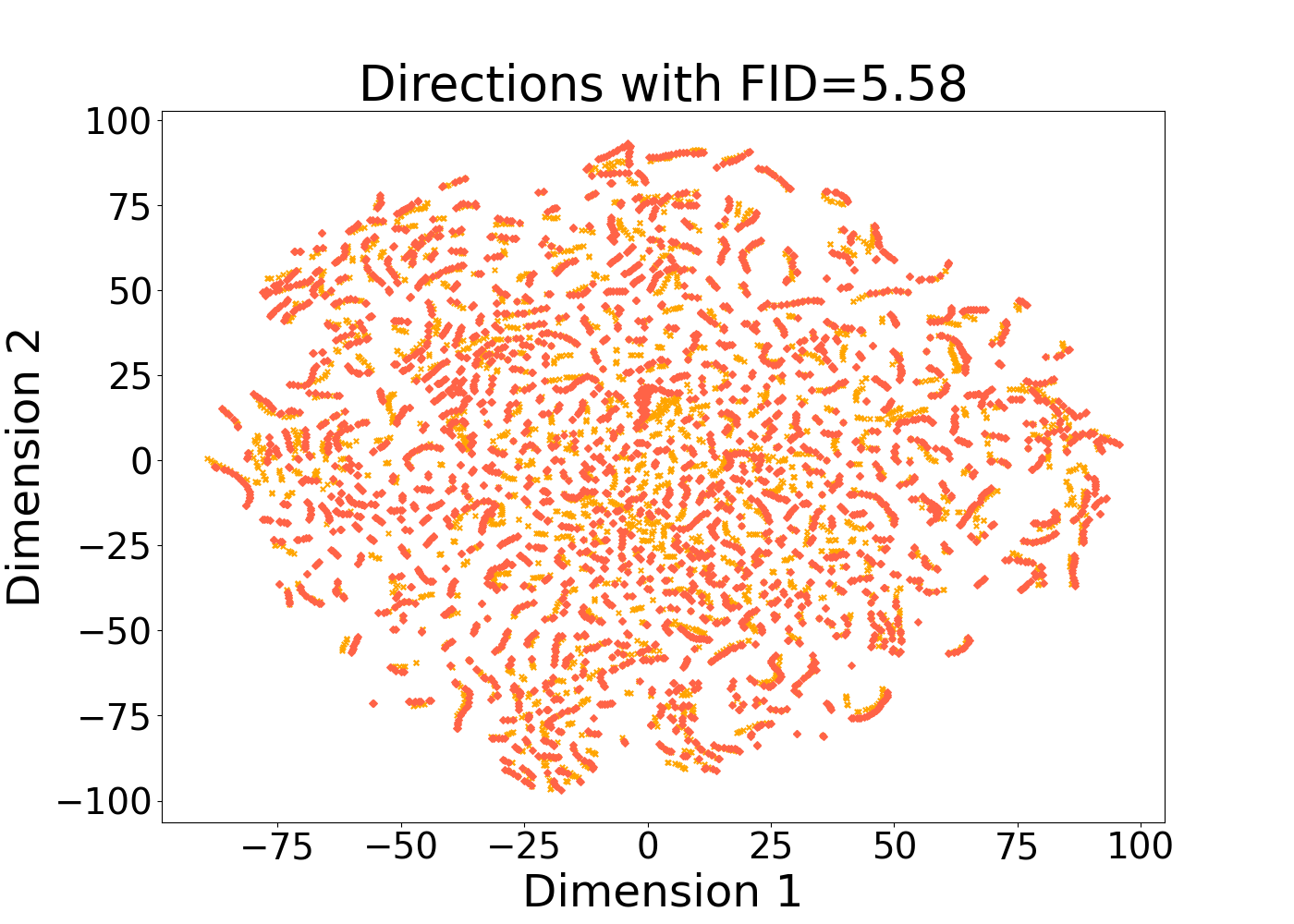}
        }
    \subfigure[Posing, eating, and smoking in STBMPT.]{
    \includegraphics[width=0.33\linewidth]{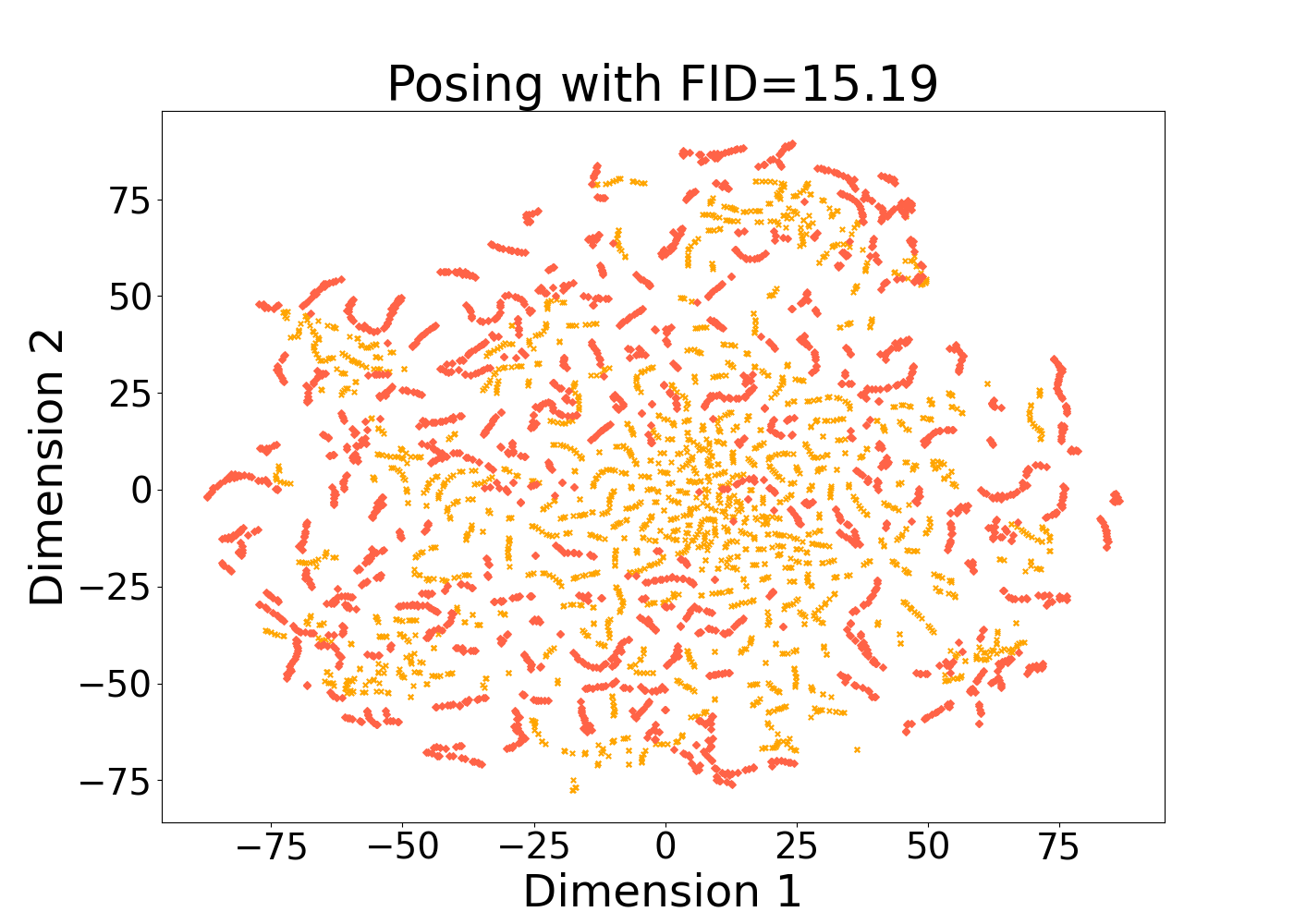}
     \includegraphics[width=0.33\linewidth]{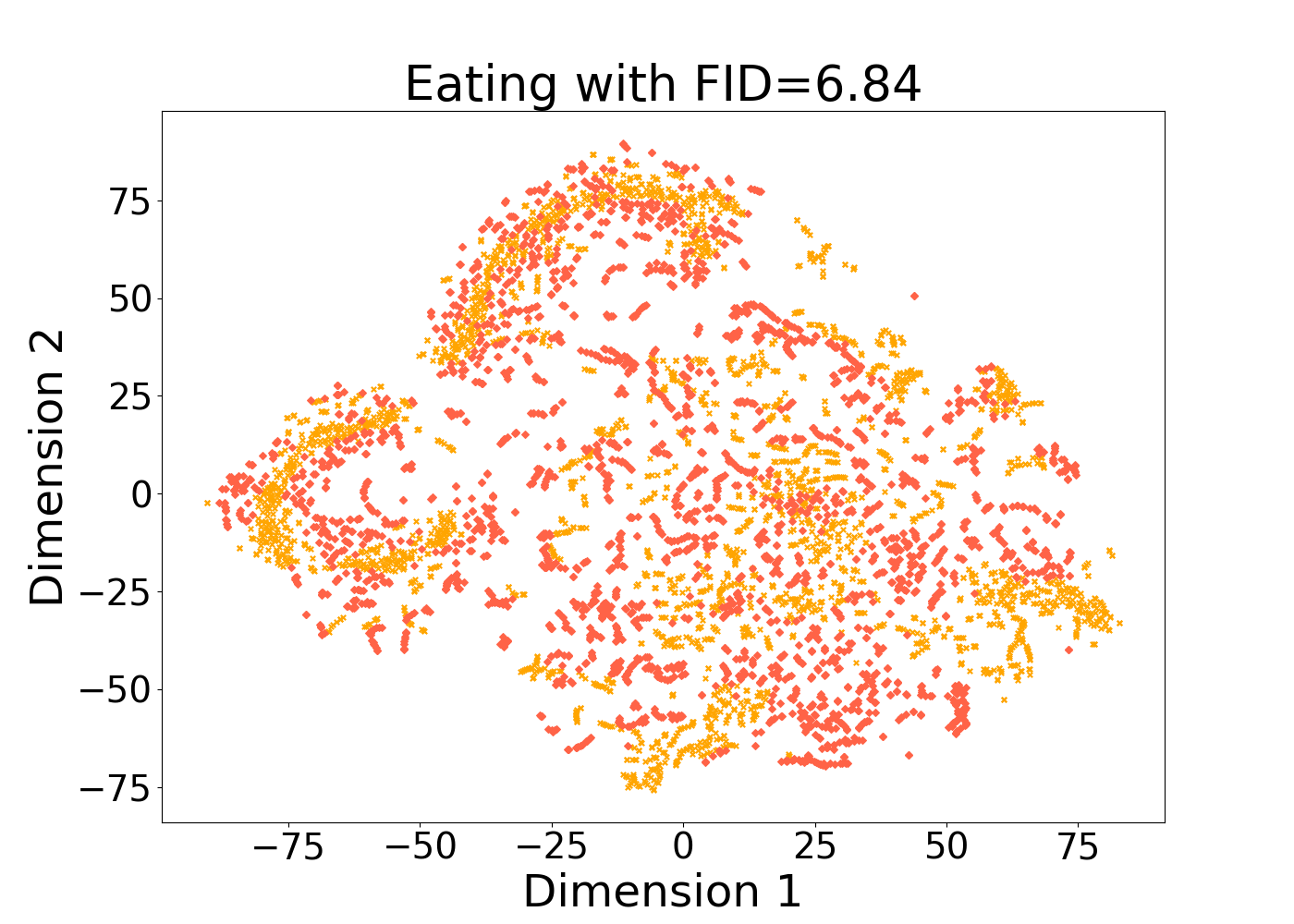}
    \includegraphics[width=0.33\linewidth]{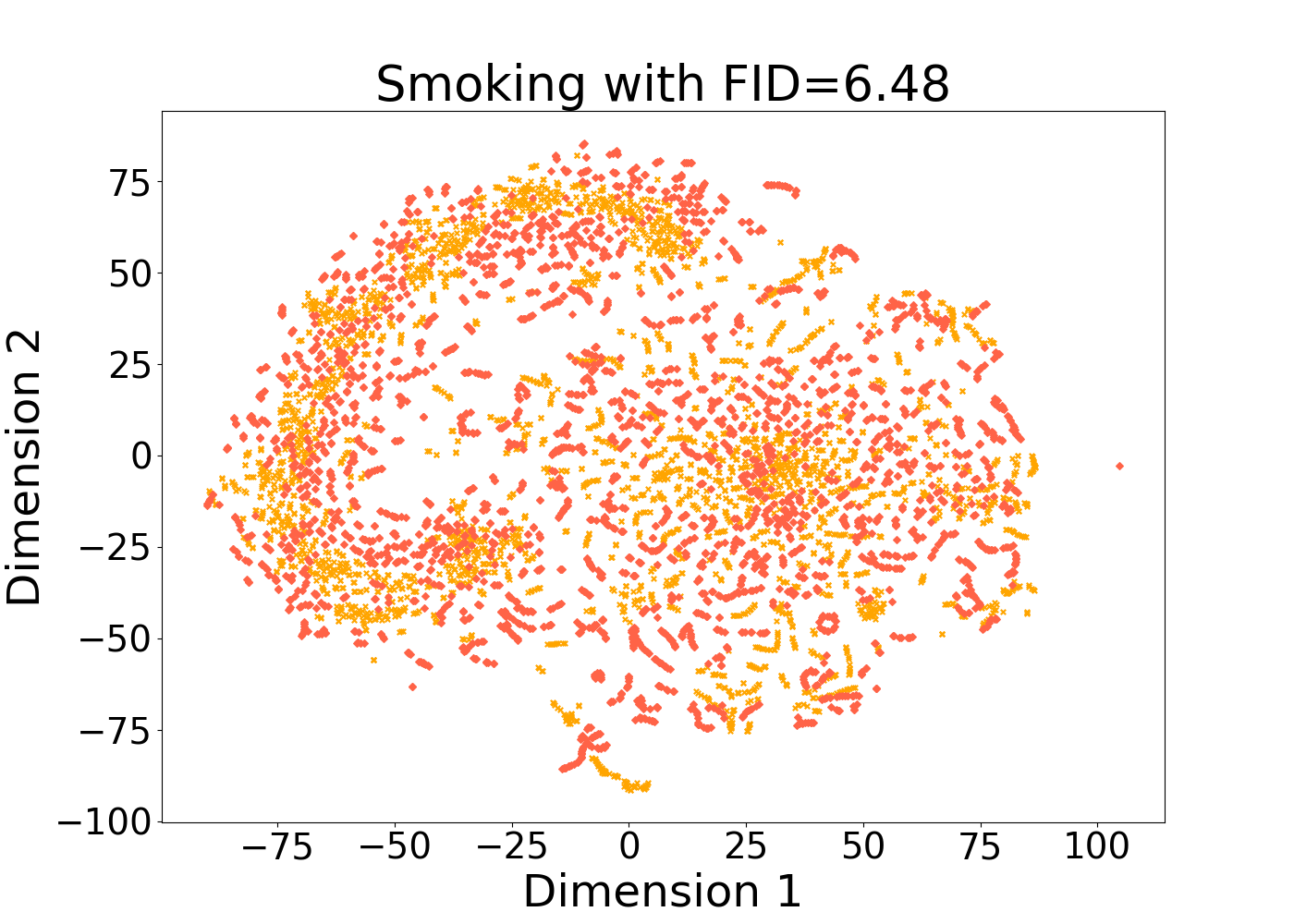}
    }
    \subfigure[Posing, eating, and smoking in STBMPT integrated with our method.]{
        \includegraphics[width=0.33\linewidth]{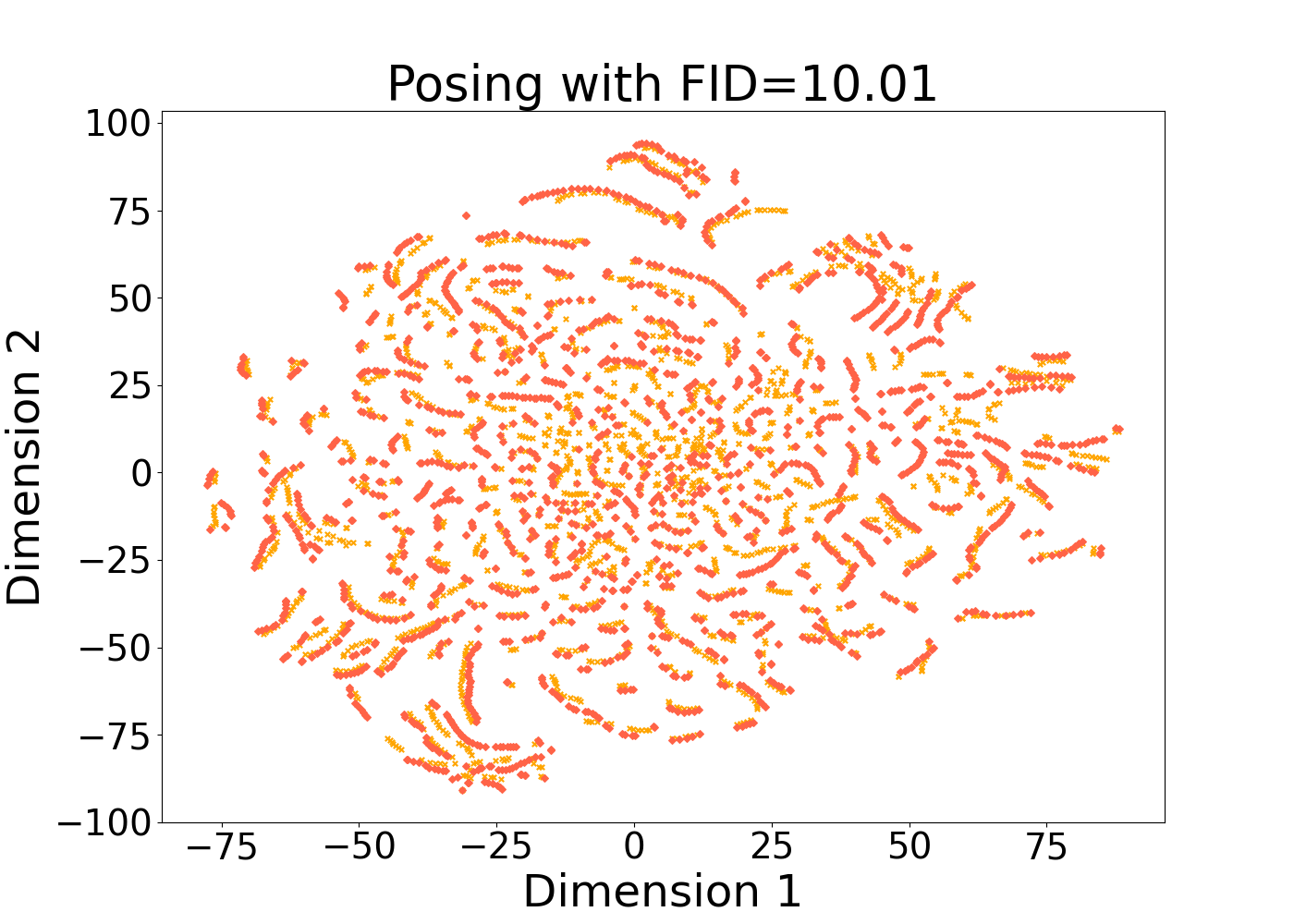} 
        \includegraphics[width=0.33\linewidth]{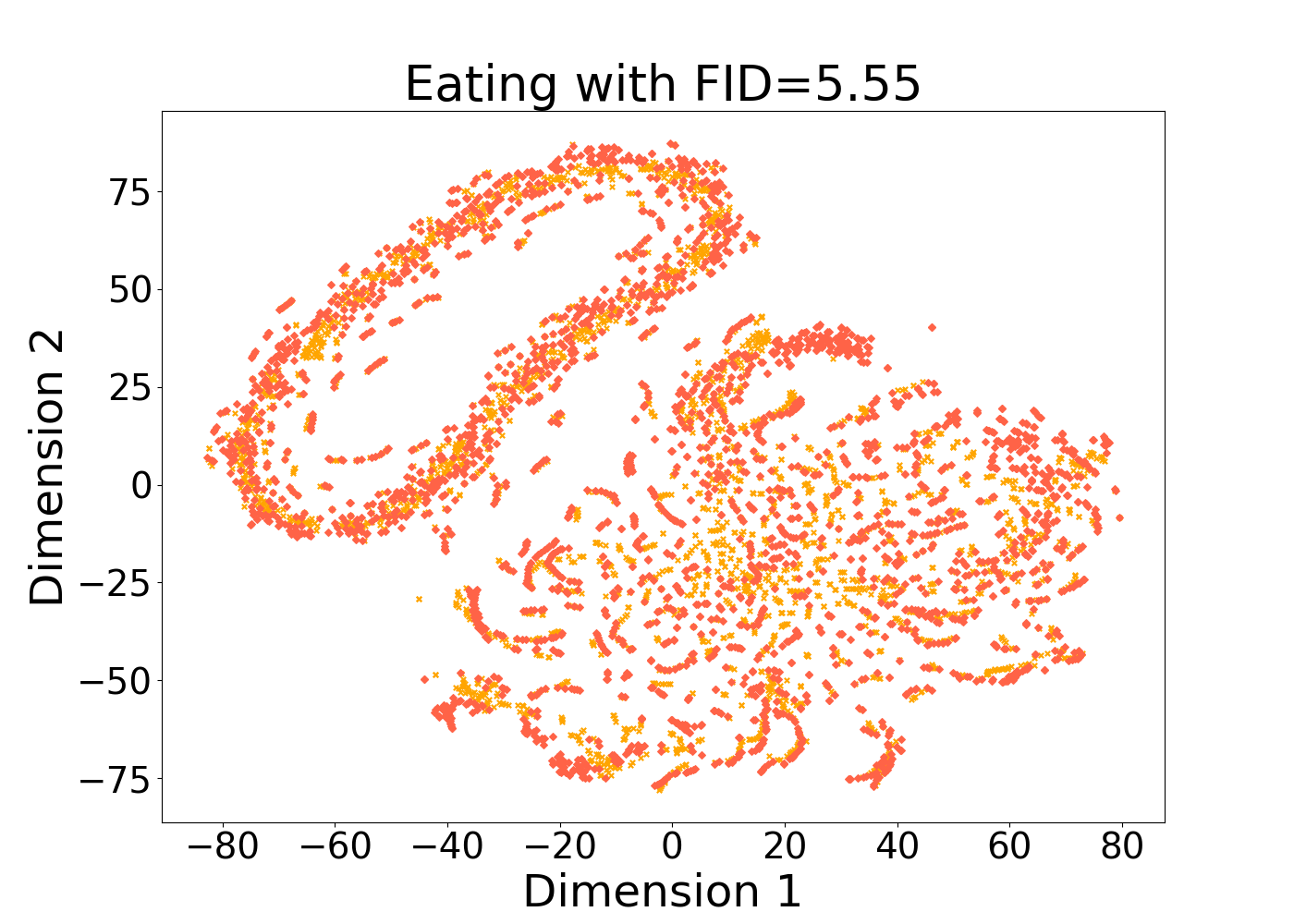}
        \includegraphics[width=0.33\linewidth]{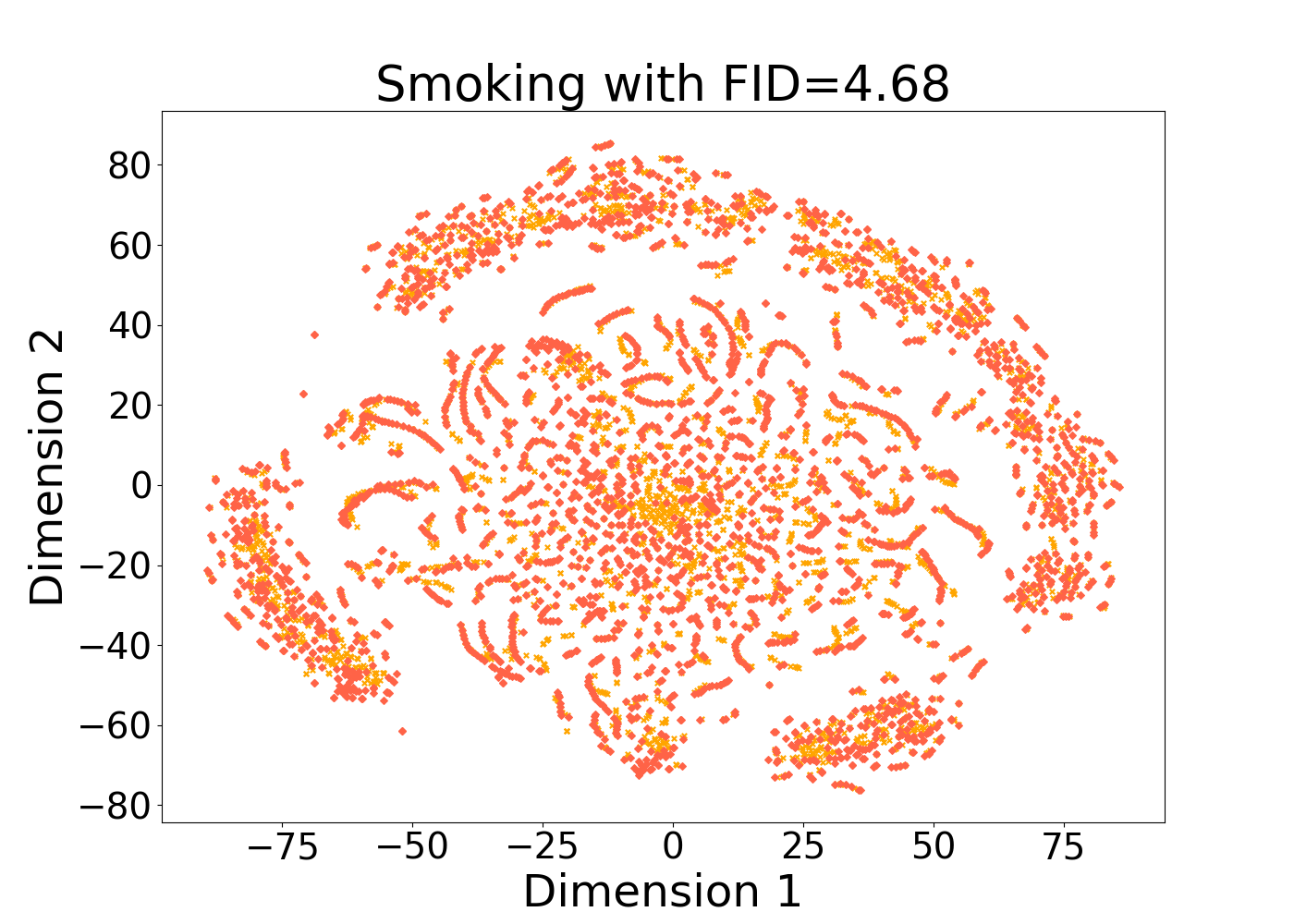}
        }
    \\
\caption{T-SNE visualization of predicted human motion. The red represent ground truth, while the yellow depict the motion features predicted by the model.}
\label{fig:feature_compare_spgsn_appendix}
\end{figure*}

To illustrate the effectiveness of our method, additional visual experimental results are presented in~\figurename~\ref{fig:feature_compare_spgsn_appendix}. Specifically, we use T-SNE to reduce dimensionality and visualize the predicted action features for "Sitting", "Waiting", "Directions", "Posing", "Eating", and "Smoking" actions. This visualization confirms that our method brings the predicted action features notably closer to the ground truth features. Additionally, we employ the Frechet Inception Distance (FID) to quantify the dissimilarity between the predicted and real action feature distributions, where a smaller FID value signifies a closer match in distribution. The results demonstrate a significant reduction in FID values for our predictions across all actions.

\subsection{Evaluation of Prediction Stability and Effectiveness}

\begin{figure}[ht]
    \centering
    \includegraphics[width=0.7\linewidth]{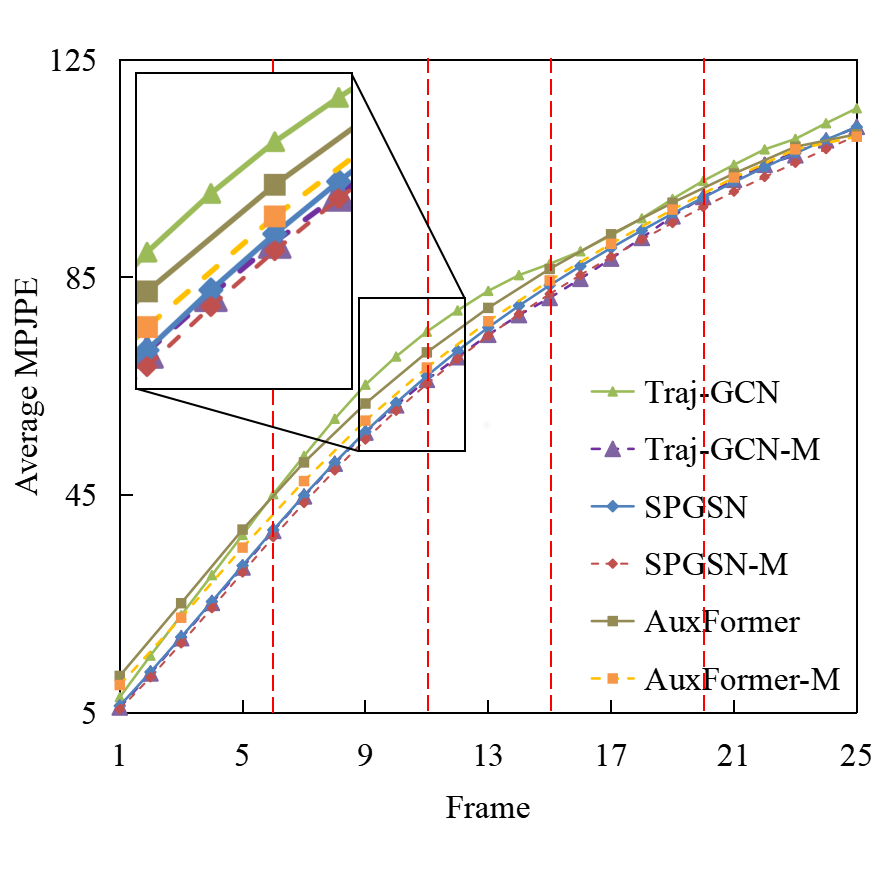}
    \vspace{-0.2in}
    \caption{Comparison of predictive performance in H3.6M. }
    \label{fig:result_curve}
\end{figure}

\textbf{Stability of prediction.} To further demonstrate the effectiveness of our method, we visualize the performance of all predictive frames in \figurename{~\ref{fig:result_curve}}. It is clear that with the integration of our method: 1) prediction errors consistently decrease at every time step, indicating superior performance compared to the original model; and 2) error variations per frame become more linear (e.g., Traj-GCN and Traj-GCN-M), implying greater stability in predictions and reduced likelihood of sudden changes. These findings further highlight the benefits of our approach.

\textbf{Effect of Varying Input Lengths.}
We conducted experiments to evaluate the performance of our method with varying input lengths (output: 10 frames) on the H3.6M dataset. The results, shown in \tablename{~\ref{tab:h3.6m_varyinput}}, demonstrate that integrating our method consistently improves prediction performance across different input lengths. This confirms the effectiveness of our approach under longer inputs.

\begin{table}[ht]
\centering
\caption{Average prediction MPJPEs on H3.6M under varying input lengths (output is 10 frames).}
\setlength{\tabcolsep}{13pt} 
\label{tab:h3.6m_varyinput}
\begin{tabular}{ccccc}
\toprule
Input Frames & 10 &20 & 25 & 50 \\
\hline
Traj-GCN &37.31 & 37.01 & 37.26  & 35.81 \\
Traj-GCN-M &\textbf{36.35} & \textbf{36.16} & \textbf{36.45}  & \textbf{35.75} \\ \hline
STBMPT & 36.34 & 36.36 & 36.13 & 35.69 \\
STBMPT-M & \textbf{35.79} & \textbf{35.36} & \textbf{35.54} & \textbf{35.05} \\ \hline
STBMPS & 34.29 & 34.00 & 34.20 & 34.00 \\
STBMPS-M & \textbf{34.09} & \textbf{33.95} & \textbf{33.77}  & \textbf{33.76} \\
\bottomrule
\end{tabular}
\end{table}

\textbf{Maximum Prediction Errors in Joint-level Prediction.}
To further validate the smoothness of predictions, we conducted experiments on the H3.6M dataset and measured the maximum MPJPE in the joint-level prediction of the prediction frame. The results, shown in \tablename{~\ref{tab:h3.6m_max}}, demonstrate that after integrating our method, the model exhibits reduced maximum joint prediction errors across various time steps, further confirming the smoothness of the predictions.

\begin{table}[ht]
\centering
\caption{Maximum prediction errors (MPJPE) in joint-level prediction at 80ms, 160ms, 320ms, and 400ms in the future are shown.}
\setlength{\tabcolsep}{12pt} 
\begin{tabular}{ccccc}
\toprule
Method          & 80ms   & 160ms  & 320ms  & 400ms  \\
\hline
Traj-GCN  & 82.55  & 162.85 & 384.05 & 461.12 \\
Traj-GCN-M & \textbf{76.05} & \textbf{162.14} & \textbf{373.09} & \textbf{439.23} \\
\hline
STBMPT    & 116.80 & 227.06 & 440.47 & 523.44 \\
STBMPT-M  & \textbf{116.29} & \textbf{224.45} & \textbf{438.75} & \textbf{521.61} \\
\hline
STBMPS    & 111.02 & 212.48 & 433.00 & 522.87 \\
STBMPS-M  & \textbf{109.60} & \textbf{210.78} & \textbf{431.27} & \textbf{521.47} \\
\bottomrule
\end{tabular}
\label{tab:h3.6m_max}
\label{tab:comparison}
\end{table}

\subsection{Consistency of Different Decoders} 
To illustrate the consistency of prediction in $MDD$ intuitively, we provide a quantitative analysis in \tablename{~\ref{tab:h3.6m_mdd}}, where we can find the following observations. 
Firstly, we can find that as the shared prediction horizon increases, the prediction differences between different decoders also increase in most cases. 
Secondly, it is worth noting that the prediction differences under different decoders in the shared prediction horizons are somewhat stable, with differences within the range of 1.1\%. Therefore, the results confirm that $MDD$ ensures the consistency of the predictions across multiple decoders, thereby avoiding unwanted behaviors arising from conflicting predictions.

\begin{table}[ht]
\centering
\setlength{\tabcolsep}{10pt} 
\caption{Average MPJPEs of the shared prediction horizons between different decoders (De-1$\sim$De-6) of SPGSN-M on H3.6M for long-term prediction. 
The final output, including 25 frames, has an average MPJPE of 67.60.
}
\label{tab:h3.6m_mdd}
\begin{tabular}{lcccccc}
\toprule
& De-1 & De-2 & De-3 & De-4 & De-5 & De-6 \\ \hline 
De-1 & - &0.09 &0.12 &0.12 &0.14 &0.13\\
De-2 & 0.09 & - &0.41 &0.55 &0.66 &0.79\\
De-3 & 0.12 &0.41 & - &0.42 &0.72 &1.09\\
De-4 & 0.12 &0.55 &0.42 &- &0.53 &1.09\\
De-5 & 0.13 &0.66 &0.72 &0.53 &- &0.84\\
De-6 & 0.13 &0.79 &1.09 &1.09 &0.84 &-\\
\bottomrule
\end{tabular}
\end{table}

\subsection{Robustness under Variable Input Lengths} 
To further evaluate the effectiveness of our method, we conducted experiments on H3.6M with varying input lengths (output fixed at 10 frames) and reported the average MPJPEs in \tablename{~\ref{tab:frame_comparison}}. Compared to the  prediction results in \tablename{~\ref{tab:h3.6m}} (input and output are 10 frames), our method consistently improves performance across different input lengths, validating its stability and adaptability.

\begin{table}[htbp]
\setlength{\tabcolsep}{6.5pt}
    \centering
    \caption{Prediction performance with varying input frame lengths. The average MPJPEs across all actions at 80ms, 160ms, 320ms and 400ms in H3.6M are reported}
    \begin{tabular}{lcccc}
        \toprule
        Input Frames & 20 frames & 25 frames & 30 frames & 50 frames \\
        \midrule
        Traj-GCN    & 37.01  & 37.26  & 34.30  & 35.81  \\
        Traj-GCN-M  & \textbf{36.16}  & \textbf{36.45}  & \textbf{34.10}  & \textbf{35.75}  \\ \hline
        STBMPT      & 36.36  & 36.13  & -      & 35.69  \\
        STBMPT-M    & \textbf{35.36}  & \textbf{35.54}  & -      & \textbf{35.05}  \\ \hline
        STBMPS      & 34.00  & 34.20  & -      & 34.00  \\
        STBMPS-M    & \textbf{33.95}  & \textbf{33.77}  & -      & \textbf{33.76}  \\
        \bottomrule
    \end{tabular}
    \label{tab:frame_comparison}
\end{table}

\subsection{Long-term Performance under Different $K$ Setting} 
\begin{figure}[ht]
  \centering
  \subfigure[ 560ms.]{\includegraphics[width=0.47\linewidth]{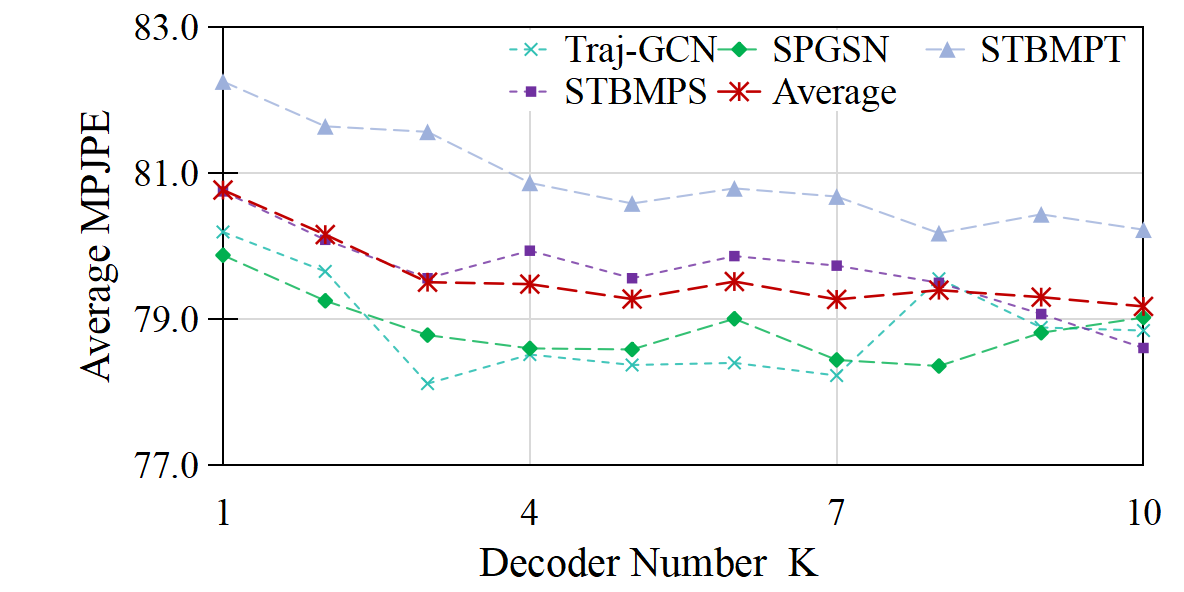}\label{fig:1}}
  \subfigure[1000ms.]{\includegraphics[width=0.47\linewidth]{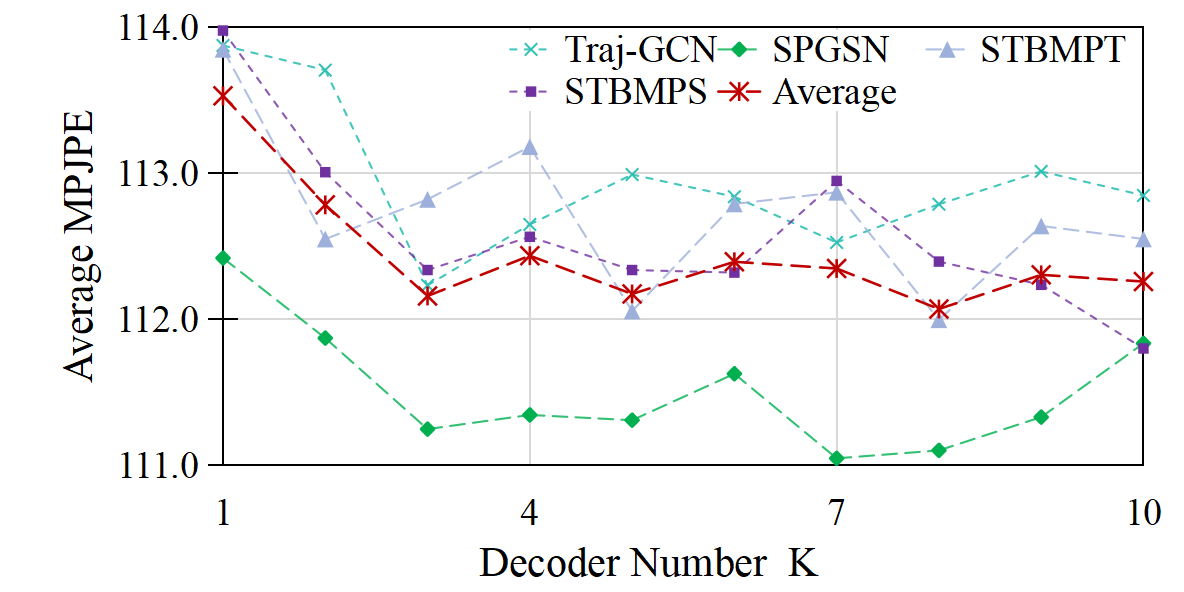}\label{fig:2} }
  \vspace{-0.1in}
   \caption{Ablation on the number of decoding. Average MPJPE across all actions at 560ms and 1000ms in H3.6M is reported.}
   \vspace{-0.2in}
   \label{fig:result_decodernumber_appendix}
\end{figure}

To provide a more comprehensive hyperparameter sensitivity analysis, we extend our study to long-term prediction performance under different $K$ settings. The results, shown in \figurename{~\ref{fig:1}} and \figurename{~\ref{fig:2}}, supplement the findings in \figurename{~\ref{fig:result_decodernumber}}.

\begin{table*}[ht]\small
\centering
\setlength{\tabcolsep}{3.5pt} 
\caption{Comparisons of short-term prediction MPJPEs on H3.6M. Results for motion prediction at 80ms, 160ms, 320ms, and 400ms in the future are shown, along with the average MPJPEs across all actions. The better results are highlighted in bold.}
\label{tab:h3.6M—shortterm-appendix}
\begin{tabular}{lcccccccccccccccc}
\toprule
 Motion      & \multicolumn{4}{c}{Walking} & \multicolumn{4}{c}{Eating} & \multicolumn{4}{c}{Smoking} & \multicolumn{4}{c}{Discussion}\\ \cmidrule(r){1-1} \cmidrule(r){2-5} \cmidrule(r){6-9} \cmidrule(r){10-13} \cmidrule(r){14-17}
 millisecond & 80 & 160 & 320 & 400 & 80 & 160 & 320 & 400 & 80 & 160 & 320 & 400 & 80 & 160 & 320 & 400 \\ \hline
 Traj-GCN&   9.06 &\textbf{15.97} &30.09 &34.49 &8.75 &18.74 &39.07 &47.66 &7.96 &15.22 &25.51 &30.08 &9.87 &22.74 &41.22 &45.36\\
 Traj-GCN-M&  \textbf{8.59} & 16.10 & \textbf{28.82} & \textbf{32.43} & \textbf{8.21} & \textbf{17.97} & \textbf{38.05} & \textbf{46.18} & \textbf{7.11} & \textbf{13.39} & \textbf{24.04} & \textbf{29.39} & \textbf{9.64} & \textbf{22.42} & \textbf{40.56} & \textbf{44.55}\\ \hline
 SPGSN   & 10.38 &19.95 &36.00 &42.82 &7.27& 15.26 &31.24 &38.81 &6.85 &14.05 &28.12 &34.79 &10.39 &23.86 &53.69 &66.85\\
 SPGSN-M & \textbf{9.98} & \textbf{19.54} & \textbf{35.50} & \textbf{42.15} & \textbf{6.97} & \textbf{14.94} & \textbf{30.83} & \textbf{38.41} & \textbf{6.53} & \textbf{13.78} & \textbf{27.82} & \textbf{34.43} & \textbf{10.02} & \textbf{23.52} & \textbf{53.36} & \textbf{66.63}\\   \hline
 EqMotion  & 9.44 &\textbf{18.62} &\textbf{34.73} &\textbf{41.70} &6.48 &14.22 &29.85 &\textbf{37.18} &6.04 &12.98 &\textbf{26.82} &\textbf{33.46} &9.24 &22.58 &52.20 &\textbf{65.35}\\ 
 EqMotion-M & \textbf{9.22} & 18.82 & 36.01 & 43.81 & \textbf{6.16} & \textbf{14.03} & \textbf{30.22} & 38.13 & \textbf{5.71} & \textbf{12.86} & 27.15 &\textbf{34.17} & \textbf{8.65} & \textbf{21.87} & \textbf{52.10} & 65.77\\ \hline

 STBMPT   & 10.52 & 20.69 & 36.21 & 42.84 &7.22 & 15.82 & \textbf{32.01} & \textbf{39.73} &6.93 & 14.75 & \textbf{29.63} & \textbf{36.46} &10.21 & 24.62 & 55.30 & 68.56\\
 STBMPT-M & \textbf{10.01} &\textbf{19.41} &\textbf{35.39} &\textbf{42.64} &\textbf{7.05} &\textbf{15.51} &32.05 &40.26 &\textbf{6.7}5 &\textbf{14.43} &29.71 &36.87 &\textbf{9.91} &\textbf{23.94} &\textbf{54.17} &\textbf{67.43}\\ \hline
 STBMPS   & 9.38 & 18.72 & \textbf{34.21} & \textbf{40.47} & 6.40 & 14.39 & \textbf{30.14} & \textbf{37.57} & 6.08 & 13.49 & 27.97 & 34.65 & 9.19 & 22.98 & 53.71 & 67.15\\
 STBMPS-M &  \textbf{9.05} &\textbf{18.30} &34.27 &41.10 &\textbf{6.25} &\textbf{14.25} &30.35 &37.95 &\textbf{5.95} &\textbf{13.25} &\textbf{27.80} &\textbf{34.63} &\textbf{8.99} &\textbf{22.64} &\textbf{53.17} &\textbf{66.56}\\ 
 \bottomrule\toprule

 Motion      & \multicolumn{4}{c}{Directions} & \multicolumn{4}{c}{Greeting} & \multicolumn{4}{c}{Phoning} & \multicolumn{4}{c}{Posing}\\ \cmidrule(r){1-1} \cmidrule(r){2-5} \cmidrule(r){6-9} \cmidrule(r){10-13} \cmidrule(r){14-17}
 millisecond & 80 & 160 & 320 & 400 & 80 & 160 & 320 & 400 & 80 & 160 & 320 & 400 & 80 & 160 & 320 & 400 \\ \hline
 Traj-GCN&   12.88 &24.38 &49.79 &60.81 &13.83 &28.05 &\textbf{65.03} &\textbf{79.79} &11.75 &19.91 &37.72 &\textbf{43.17} &\textbf{8.85} &\textbf{23.37} &67.24 &83.77\\
 Traj-GCN-M & \textbf{11.71} & \textbf{22.94} & \textbf{46.92} & \textbf{57.26} & \textbf{13.41} & \textbf{27.90} & 68.45 & 84.17 & \textbf{11.03} & \textbf{19.01} & \textbf{36.40} & 43.51 & 8.87 & 23.69 & \textbf{66.73} & \textbf{83.14}\\ \hline
 SPGSN   & 7.52 &17.56 &40.43 &50.92 &14.91 &32.83 &\textbf{70.55} &\textbf{86.87} &8.77 &18.67 &39.13 &49.01 &11.02 &25.51 &60.04 &77.01\\
 SPGSN-M & \textbf{7.19} & \textbf{17.24} & \textbf{40.04} & \textbf{50.48} & \textbf{14.22} & \textbf{32.45} & 70.64 & 86.91 & \textbf{8.45} & \textbf{18.36} & \textbf{38.89} & \textbf{48.81} & \textbf{10.58} & \textbf{25.07} & \textbf{59.38} & \textbf{76.32}\\ \hline
 EqMotion  & 6.48 &16.20 &39.24 &50.09 &13.27 &31.05 &69.73 &\textbf{86.31} &7.64 &17.31 &\textbf{37.81} &\textbf{47.70} &9.60 &24.54 &60.51 &78.38\\ 
 EqMotion-M & \textbf{6.01} &\textbf{15.76} &\textbf{39.07} &\textbf{50.07} &\textbf{12.13} &\textbf{29.97} &\textbf{69.30} &86.57 &\textbf{7.35} &\textbf{17.10} &38.16 &48.50 &\textbf{8.85} &\textbf{23.28} &\textbf{59.31} &\textbf{77.84}\\ \hline
 STBMPT   & 7.23 & 18.06 & 42.15 & 52.92 & 14.57 & 34.80 & 74.18 & 90.39 & 8.81 & 19.32 & 40.57 & 50.73 & 10.70 & 26.35 & 62.35 & 79.57\\
 STBMPT-M & \textbf{6.97} &\textbf{17.56} &\textbf{41.38} &\textbf{52.16} &\textbf{14.37} &\textbf{34.19} &\textbf{72.88} &\textbf{89.31} &\textbf{8.55} &\textbf{18.91} &\textbf{40.34} &\textbf{50.72} &\textbf{10.33} &\textbf{25.61} &\textbf{61.68} &\textbf{79.14}\\ \hline
 STBMPS   & 6.44 & 16.69 & 40.49 & 51.30 &13.63 & 32.48 & 71.88 & 88.55 &7.75 & 17.77 & 38.71 & 48.66 &9.40 & 24.15 & 59.20 & 76.07 \\
 STBMPS-M & \textbf{6.32} &\textbf{16.46} &\textbf{39.84} &\textbf{50.43} &\textbf{13.20} &\textbf{31.77} &\textbf{70.31} &\textbf{86.58} &\textbf{7.62} &\textbf{17.61} &\textbf{38.53} &\textbf{48.53} &\textbf{9.20} &\textbf{23.76} &\textbf{58.50} &\textbf{75.44}\\ 
 \bottomrule \toprule

 Motion      & \multicolumn{4}{c}{Purchases} & \multicolumn{4}{c}{Sitting} & \multicolumn{4}{c}{Sitting Down} & \multicolumn{4}{c}{Taking Photo}\\ \cmidrule(r){1-1} \cmidrule(r){2-5} \cmidrule(r){6-9} \cmidrule(r){10-13} \cmidrule(r){14-17}
 millisecond & 80 & 160 & 320 & 400 & 80 & 160 & 320 & 400 & 80 & 160 & 320 & 400 & 80 & 160 & 320 & 400 \\ \hline
 Traj-GCN& 19.24 &\textbf{38.04} &61.11 &\textbf{70.36} &10.64 &25.00 &50.63 &61.30 &11.54 &27.57 &55.29 &\textbf{65.18}  &6.86 &16.19 &42.24 &54.27 \\
 Traj-GCN-M & \textbf{18.93} & 38.18 & \textbf{60.86} & 71.54 & \textbf{10.33} & \textbf{23.56} & \textbf{48.73} & \textbf{59.22} & \textbf{11.33} & \textbf{26.53} & \textbf{55.07} & 66.53 & \textbf{6.61} & \textbf{16.11} & \textbf{41.11} & \textbf{53.05} \\ \hline
 SPGSN   & 12.67 &\textbf{28.20} &\textbf{60.05} &\textbf{73.70} &9.28 &19.50 &42.53 &53.88 &14.34 &28.08 &57.20 &71.28 &8.90 &19.12 &41.83 &53.10\\
 SPGSN-M & \textbf{12.43} & 28.29 & 60.58 & 74.10 & \textbf{9.06} & \textbf{19.33} & \textbf{42.43} & \textbf{53.82} & \textbf{13.87} & \textbf{27.40} & \textbf{56.36} & \textbf{70.42} & \textbf{8.50} & \textbf{18.62} & \textbf{41.28} & \textbf{52.53} \\ \hline
 EqMotion  & 11.43 &26.88 &59.14 &73.12 &8.09 &17.94 &41.04 &52.57 &12.97 &26.06 &54.86 &68.95 &8.04 &18.04 &41.25 &52.90\\ 
 EqMotion-M& \textbf{10.77} &\textbf{26.37} &\textbf{59.46} &\textbf{73.71} &\textbf{7.83} &\textbf{17.81} &\textbf{41.32} &\textbf{52.96} &\textbf{12.53} &\textbf{25.72} &\textbf{55.23} &\textbf{69.84} &\textbf{7.60 }&\textbf{17.57} &\textbf{41.02} &\textbf{52.81}\\  \hline
 STBMPT   & 13.06 & 30.30 & 63.81 & 77.25 & 10.01 & 21.10 & 45.36 & 56.96 & 15.47 & 29.86 & 59.36 & 73.46 & 8.99 & 19.92 & 44.17 & 55.79\\
 STBMPT-M & \textbf{12.56} &\textbf{29.20} &\textbf{62.12} &\textbf{75.86} &\textbf{9.73} &\textbf{20.57} &\textbf{44.85} &\textbf{56.60} &\textbf{15.33} &\textbf{29.59} &\textbf{59.10} &\textbf{73.22} &\textbf{8.71} &\textbf{19.37} &\textbf{43.56} &\textbf{55.33}\\ \hline
 STBMPS   & 11.48 & 27.84 & 61.63 & 75.75 & 8.40 & \textbf{18.69} & \textbf{42.48} & \textbf{54.13} &13.49 & 27.06 & 56.35 & 70.37 & 7.99 & 18.26 & 41.74 & 53.25\\
 STBMPS-M & \textbf{11.25} &\textbf{27.47} &\textbf{61.33} &\textbf{75.32} &\textbf{8.33} &18.81 &42.99 &54.68 &\textbf{13.26} &\textbf{26.89} &\textbf{56.30} &\textbf{70.30} &\textbf{7.80} &\textbf{17.90} &\textbf{41.05} &\textbf{52.37}\\ 
 \bottomrule \toprule

 Motion      & \multicolumn{4}{c}{Waiting} & \multicolumn{4}{c}{Walking Dog} & \multicolumn{4}{c}{Walking Together} & \multicolumn{4}{c}{Average}\\ \cmidrule(r){1-1} \cmidrule(r){2-5} \cmidrule(r){6-9} \cmidrule(r){10-13} \cmidrule(r){14-17}
 millisecond & 80 & 160 & 320 & 400 & 80 & 160 & 320 & 400 & 80 & 160 & 320 & 400 & 80 & 160 & 320 & 400 \\ \hline
 Traj-GCN&  9.49 &\textbf{21.83} &\textbf{58.25} &\textbf{75.28} &31.97 &\textbf{55.57} &\textbf{98.48} &\textbf{117.94} &9.27 &18.86 &35.49 &44.10 &12.13 &24.76 &50.48 &60.90\\
 Traj-GCN-M& \textbf{9.05}& 22.42 & 59.42 & 76.33 & \textbf{30.49} & 55.75 & 99.47 & 120.70 & \textbf{8.14} & \textbf{16.99} & \textbf{33.69} & \textbf{43.08} & \textbf{11.56} & \textbf{24.20} & \textbf{49.89} & \textbf{60.75}\\  \hline
 SPGSN   & 9.20 &19.98 &43.44 &\textbf{54.25} &17.94 &37.12 &\textbf{71.43} &\textbf{84.91} &9.00 &18.37 &\textbf{33.91} &\textbf{40.54} &10.56 &22.54 &47.31 &58.58\\
 SPGSN-M & \textbf{8.98} & \textbf{19.85} & \textbf{43.28} & 54.33 & \textbf{17.47} & \textbf{36.99} &71.69 & 85.17 & \textbf{8.57} & \textbf{18.00} & 34.04 & 40.84 & \textbf{10.19} & \textbf{22.22} & \textbf{47.08} & \textbf{58.36}\\  \hline
 EqMotion  & 8.03 &18.32 &41.57 &53.07 &16.84 &36.48 &72.24 &85.79  &8.33 &17.58 &33.02 &39.57 &9.46 &21.25 &46.27 &57.74\\ 
 EqMotion-M& \textbf{7.48} &\textbf{17.82} &\textbf{42.13} &\textbf{54.17 }&\textbf{16.12} &\textbf{35.88} &\textbf{72.64} &\textbf{86.93} &\textbf{8.06} &\textbf{17.58} &\textbf{33.73} &\textbf{40.78} &\textbf{8.96}  &\textbf{20.83} &\textbf{46.46} &\textbf{58.40}\\ \hline
 
 STBMPT   & 8.97 & 20.43 & 44.71 & 55.86 & 18.53 & 39.60 & \textbf{75.02} & \textbf{88.33} &9.07 & 19.18 & 35.26 & 41.98 & 10.69 & 23.65 & 49.34 & 60.72 \\
 STBMPT-M & \textbf{8.69} &\textbf{19.84} &\textbf{44.20} &\textbf{55.62} &\textbf{18.36} &\textbf{39.41} &75.26 &89.24 &\textbf{8.65} &\textbf{18.09} &\textbf{34.54} &\textbf{41.89} &\textbf{10.40} &\textbf{23.04} &\textbf{48.75} &\textbf{60.42}\\ \hline
 
 STBMPS   &7.99 &18.80 &42.43 &53.54 &17.26 &37.67 &\textbf{73.13} &\textbf{87.10} &8.19 &17.84 &34.17 &41.08 &9.54 &21.79 &47.22 &58.64\\
 STBMPS-M & \textbf{7.85} &\textbf{18.58} &\textbf{42.24} &\textbf{53.38} &\textbf{16.97} &\textbf{37.64} &74.16 &88.07 &\textbf{8.00} &\textbf{17.42} &\textbf{33.82} &\textbf{40.90} &\textbf{9.34} &\textbf{21.52} &\textbf{46.98} &\textbf{58.42}\\ 
\bottomrule
\end{tabular}
\end{table*}

\begin{table*}[ht]\small
\centering
\setlength{\tabcolsep}{2.1pt} 
\caption{Comparison of long-term prediction performance on H3.6M. The table presents the performance at 560ms and 1000ms in the future.}
\label{tab:h3.6M—longterm-appendix}
\begin{tabular}{lcccccccccccccccc}
\toprule
 Motion      & \multicolumn{2}{c}{Walking} & \multicolumn{2}{c}{Eating} & \multicolumn{2}{c}{Smoking} & \multicolumn{2}{c}{Discussion}  & \multicolumn{2}{c}{Directions} & \multicolumn{2}{c}{Greeting} & \multicolumn{2}{c}{Phoning} & \multicolumn{2}{c}{Posing} \\ \cmidrule(r){1-1} \cmidrule(r){2-3} \cmidrule(r){4-5} \cmidrule(r){6-7} \cmidrule(r){8-9} \cmidrule(r){10-11} \cmidrule(r){12-13} \cmidrule(r){14-15} \cmidrule(r){16-17}
 millisecond & 560 & 1K & 560 & 1K & 560 & 1K & 560 & 1K & 560 & 1K & 560 & 1K  & 560 & 1K & 560 & 1K\\ \hline 
 Traj-GCN&  \textbf{40.54} &\textbf{46.34} &59.03 &71.08 &\textbf{34.10} &63.03 &68.17 &102.59 &\textbf{81.06} &\textbf{105.88} &101.26 &\textbf{93.20} &64.58 &115.41 &110.12 &211.41 \\
 Traj-GCN-M & 42.99 & 50.82 & \textbf{57.76} & \textbf{70.33} & 34.83 & \textbf{62.95} & \textbf{73.41} & \textbf{106.80} & 81.35 & 106.25 & \textbf{100.56} & 95.39 & \textbf{63.82} & \textbf{112.17} & \textbf{107.62} & \textbf{209.55}\\ \hline
 SPGSN   &  52.09 &58.62 &52.06 &76.52 &48.79 &71.08 &90.51 &121.88 &71.18 &102.67 &114.16 &147.00 &66.84 &102.89 &\textbf{108.14} &\textbf{164.25}\\
 SPGSN-M & \textbf{50.70} & \textbf{58.62} & \textbf{51.18} & \textbf{75.56} & \textbf{47.39} & \textbf{69.67} & \textbf{89.27} & \textbf{119.82} & \textbf{69.66} & \textbf{100.73} & \textbf{111.31} & \textbf{145.59} & \textbf{66.64} & \textbf{102.15} & 109.09 & 166.95\\ \hline
 EqMotion  & 43.87 &53.72 &49.49 &73.88 &44.70 &67.00 &86.80 &116.98 &68.90 &\textbf{99.82} &108.99 &143.97 &\textbf{65.14} &\textbf{101.51} &109.43 &169.92\\ 
 EqMotion-M& \textbf{43.34} & \textbf{53.45} & \textbf{48.24} & \textbf{72.46} & \textbf{43.73} & \textbf{66.46} & \textbf{85.01} & \textbf{114.51} & \textbf{68.69} & 99.94 & \textbf{108.04} & \textbf{140.34} & 65.20 & 101.60 & \textbf{106.67} & \textbf{165.12}\\ \hline
 STBMPT   & \textbf{48.97} &\textbf{57.32} &\textbf{54.78} &\textbf{78.81} &50.83 &74.17 &92.45 &121.42 &75.28 &107.64 &115.53 &\textbf{146.50} &71.04 &107.58 &115.72 &171.21\\
 STBMPT-M & 51.19 &61.84 &55.02 &79.87 &\textbf{50.09} &\textbf{72.82} &\textbf{90.32} &\textbf{120.60} &\textbf{72.96} &\textbf{103.71} &\textbf{114.52} &147.21 &\textbf{70.58} &\textbf{106.63} &\textbf{114.14} &\textbf{169.40}\\ \hline
 STBMPS   & 51.51 &57.29 &53.00 &\textbf{77.43} &49.58& 72.56& 90.36 &121.58 &72.19 &\textbf{103.19} &\textbf{114.71} &148.39 &68.49 &105.66 &112.55 &168.35\\
 STBMPS-M & \textbf{50.08} &\textbf{57.04} &\textbf{52.22} &77.50 &\textbf{48.77} &\textbf{71.49} &\textbf{89.98} &\textbf{121.33} &\textbf{71.69} &104.47 &115.03 &\textbf{148.31} &\textbf{67.94} &\textbf{105.47} &\textbf{108.07} &\textbf{164.00}\\
 \bottomrule \toprule

 Motion      & \multicolumn{2}{c}{Purchases} & \multicolumn{2}{c}{Sitting} & \multicolumn{2}{c}{Sitting Down} & \multicolumn{2}{c}{Taking Photo} & \multicolumn{2}{c}{Waiting} & \multicolumn{2}{c}{Walking Dog} & \multicolumn{2}{c}{\scriptsize{Walking Together}} & \multicolumn{2}{c}{Average}\\ \cmidrule(r){1-1} \cmidrule(r){2-3} \cmidrule(r){4-5} \cmidrule(r){6-7} \cmidrule(r){8-9} \cmidrule(r){10-11} \cmidrule(r){12-13} \cmidrule(r){14-15} \cmidrule(r){16-17}
 millisecond & 560 & 1K & 560 & 1K & 560 & 1K & 560 & 1K & 560 & 1K & 560 & 1K  & 560 & 1K & 560 & 1K\\ \hline
 Traj-GCN&  \textbf{93.17} &\textbf{125.98} &84.76 &116.15 &83.69 &135.09  &\textbf{77.36} &97.67 &108.41 &\textbf{170.30} &140.33 &172.30 &\textbf{61.31} &84.30 &80.53 &114.05\\
 Traj-GCN-M& 94.14 & 126.47 & \textbf{82.35} & \textbf{112.74} & \textbf{80.10} & \textbf{133.29} & 78.33 & \textbf{95.46} & \textbf{103.04} & 174.55 & \textbf{132.21} & \textbf{166.63} & 61.43 & \textbf{83.60} & \textbf{79.60} & \textbf{113.80}\\
\hline
 SPGSN   &  100.35 &141.41 &75.18 &\textbf{116.51} &98.52 &150.00 &77.47 &121.99 &74.47 &105.53 &\textbf{103.51} &\textbf{138.72} &53.29 &63.73 &79.10 &112.19\\
 SPGSN-M & \textbf{97.76} & \textbf{136.99} & \textbf{75.10} & 116.56 & \textbf{98.16} & \textbf{149.40} & \textbf{75.07} & \textbf{118.43} & \textbf{73.14} & \textbf{103.94} & 105.73 & 142.12 & \textbf{51.80} & \textbf{63.34} & \textbf{78.13} & \textbf{111.33} \\
\hline 
 EqMotion  & 93.98 &135.56 &74.26 &116.03 &96.29 &148.31  &75.02 &\textbf{118.52} &73.09 &105.21 &104.62 &139.95 &47.27 &\textbf{57.34} &76.12 &109.85\\ 
 EqMotion-M& \textbf{93.54} & \textbf{134.54} & \textbf{72.86} & \textbf{115.07} & \textbf{94.07} & \textbf{145.63} & \textbf{74.80} & 120.04 & \textbf{71.04} & \textbf{104.01} & \textbf{103.93} & \textbf{139.32} & \textbf{46.20} & 57.73 & \textbf{75.02} & \textbf{108.68} \\
\hline
 STBMPT   & \textbf{102.71} &\textbf{139.20} &82.80 &126.14 &102.25 &152.39 &82.41 &127.20 &79.22 &110.17 &109.43 &141.91 &\textbf{53.91} &\textbf{65.80} &82.49 &115.16\\
 STBMPT-M & 104.13 &142.44 &\textbf{80.04} &\textbf{121.85} &\textbf{100.56} &\textbf{150.27} &\textbf{79.38} &\textbf{121.95} &\textbf{76.73} &\textbf{107.26} &\textbf{107.63} &\textbf{141.09} &54.49 &66.51 &\textbf{81.45} &\textbf{114.23}\\ \hline
 STBMPS   & 103.97 &145.29 &\textbf{78.46} &\textbf{121.05} &99.02 &149.95 &78.27 &122.17 &77.02 &107.09 &\textbf{108.11} &144.02 &55.32 &64.49 &80.84 &113.90\\
 STBMPS-M & \textbf{101.58} &\textbf{141.03} &78.63 &121.79 &\textbf{98.10} &\textbf{149.43} &\textbf{77.29} &\textbf{121.41} &\textbf{74.14} &\textbf{105.46} &108.38 &\textbf{144.00} &\textbf{52.86} &\textbf{63.40} &\textbf{79.65} &\textbf{113.08}\\
 \bottomrule
\end{tabular}
\end{table*}

\begin{table*}[ht]\scriptsize
\centering
\setlength{\tabcolsep}{3.5pt} 
\caption{
Comparison of prediction on 8 action categories from CMU-Mocap. Results at 80ms, 160ms, 320ms, 400ms and 1000ms in the future are shown.
}
\label{tab:CMU-Mocap-appendix}
\begin{tabular}{lcccccccccccccccccccc}\toprule
 Motion      & \multicolumn{5}{c}{basketball} & \multicolumn{5}{c}{basketball signal} & \multicolumn{5}{c}{directing traffic} & \multicolumn{5}{c}{jumping}\\ \cmidrule(r){1-1} \cmidrule(r){2-6} \cmidrule(r){7-11} \cmidrule(r){12-16} \cmidrule(r){17-21}
 millisecond & 80 & 160 & 320 & 400 &1K  & 80 & 160 & 320 & 400 &1K & 80 & 160 & 320 & 400  &1K & 80 & 160 & 320 & 400 &1K \\ \hline 
 
 Traj-GCN& \textbf{13.44 }&\textbf{23.58} &\textbf{46.58} &\textbf{58.00} &\textbf{106.84} &3.68 &6.42 &13.24 &17.28 &58.41 &7.70 &15.75 &35.42 &47.07  &162.58 &16.98 &\textbf{33.48} &\textbf{74.05} &\textbf{94.26} &174.41\\ 
 Traj-GCN-M & 13.64 & 24.28 & 47.11 & 59.00 & 108.82 & \textbf{2.95} & \textbf{5.25} & \textbf{11.95} & \textbf{16.17} & \textbf{55.71} & \textbf{6.63} & \textbf{13.86} & \textbf{29.76} & \textbf{39.30} & \textbf{140.94} & \textbf{16.34} & 33.65 & 76.50 &96.31 & \textbf{164.11} \\  \hline

 SPGSN   &  11.97 &21.58 &41.11 &51.08 &102.87 &3.55 &6.70 &14.36 &18.80 &52.21& 7.19 &14.08 &30.43 &39.74 &110.10  &17.31 &32.55 &61.32 &74.71 &136.25\\
 SPGSN-M &\textbf{10.55}& \textbf{18.94}& \textbf{37.19}& \textbf{46.81}& \textbf{99.06}& \textbf{2.84}& \textbf{5.11}& \textbf{10.80}& \textbf{14.25}& \textbf{45.22}& \textbf{5.79}& \textbf{11.27}& \textbf{25.11}& \textbf{33.38}& \textbf{101.69}& \textbf{15.57}& \textbf{29.73}& \textbf{58.13}& \textbf{71.75}& \textbf{131.45} \\ \hline
 
 STBMPT   & 10.69 &19.64 &38.93 &48.42 &90.89 &2.89 &5.76 &14.21 &19.46 &\textbf{57.20} &5.87 &12.19 &29.50 &39.82 &113.60 &15.22 &29.09 &56.70 &69.82 &125.04\\
 STBMPT-M & \textbf{10.38} &\textbf{18.93} &\textbf{37.39} &\textbf{46.22} &\textbf{84.55} &\textbf{2.80} &\textbf{5.54} &\textbf{13.43} &\textbf{18.35} &57.95 &\textbf{5.70} &\textbf{11.87} &\textbf{29.01} &\textbf{39.32} &\textbf{112.93} &\textbf{14.97} &\textbf{28.58} &\textbf{55.92} &\textbf{69.03} &\textbf{122.52}\\ \hline
 
 STBMPS   & 10.17 &19.22 &40.21 &51.10 &100.75 &3.07 &6.46 &16.81 &23.47 &74.12 &5.52 &11.60 &29.23 &40.10 &118.44 &14.91 &28.97 &57.66 &71.21 &127.27\\
 STBMPS-M & \textbf{9.43} &\textbf{17.52} &\textbf{37.18} &\textbf{47.31} &\textbf{90.91} &\textbf{2.66} &\textbf{5.27} &\textbf{13.16} &\textbf{18.23} &\textbf{62.52} &\textbf{4.95} &\textbf{10.19} &\textbf{25.47} &\textbf{34.89} &\textbf{108.04} &\textbf{13.75} &\textbf{27.23} &\textbf{56.37} &\textbf{70.43} &\textbf{125.13}\\
\bottomrule \toprule
 
Motion      & \multicolumn{5}{c}{running} & \multicolumn{5}{c}{soccer} & \multicolumn{5}{c}{walking} & \multicolumn{5}{c}{washing window}\\ \cmidrule(r){1-1} \cmidrule(r){2-6} \cmidrule(r){7-11} \cmidrule(r){12-16} \cmidrule(r){17-21}
 millisecond & 80 & 160 & 320 & 400 &1K & 80 & 160 & 320 & 400 &1K & 80 & 160 & 320 & 400 &1K & 80 & 160 & 320 & 400 &1K \\ \hline 
 Traj-GCN& 23.87 &34.54 &37.00 &37.85 &\textbf{63.57} &12.49 &23.05 &44.18 &53.77 &\textbf{111.41}  &7.36& 11.42 &20.17& 23.54 &39.33 &6.00 &12.01 &30.35 &40.17 &82.39\\
 Traj-GCN-M & \textbf{22.10} & \textbf{31.30} & \textbf{31.84} & \textbf{31.94} & 63.99 & \textbf{10.68} & \textbf{20.04} & \textbf{40.86} & \textbf{51.73} & 117.54 & \textbf{6.55} & \textbf{10.19} & \textbf{17.43} & \textbf{20.26} & \textbf{33.45} & \textbf{5.20} & \textbf{11.01} & \textbf{29.23} & \textbf{38.67} & \textbf{78.46} \\ \hline
 
 SPGSN   &  17.00 &29.68 &48.25 &52.96 &66.41 &14.43 &25.92 &46.48 &55.96 &110.13 &7.83 &13.12 &21.69 &25.11 &42.73 &6.71 &12.83 &26.13 &32.82 &\textbf{69.82}\\
 SPGSN-M & \textbf{14.49} & \textbf{24.66} & \textbf{39.44} & \textbf{43.75} & \textbf{60.91} & \textbf{12.48} & \textbf{22.39} & \textbf{41.33} & \textbf{50.86} & \textbf{108.26} & \textbf{6.90} & \textbf{11.42} & \textbf{18.73} & \textbf{22.12} & \textbf{41.97} & \textbf{5.23} & \textbf{10.16} & \textbf{22.44} & \textbf{29.38} & 70.27 \\ \hline
 
 STBMPT   & 14.07 &22.04 &31.89 &35.49 &43.74 &12.59 &22.21 &40.79 &50.79 &106.33 &6.52 &10.38 &16.71 &19.77 &\textbf{35.12} &5.36 &10.99 &25.29 &32.81 &\textbf{68.11}\\
 STBMPT-M & \textbf{12.67} &\textbf{20.02} &\textbf{31.12} &\textbf{34.68} &\textbf{42.80} &\textbf{11.37} &\textbf{19.92} &\textbf{37.81} &\textbf{47.88} &\textbf{105.15} &\textbf{6.26} &\textbf{9.77} &\textbf{15.76} &\textbf{18.89} &36.45 &\textbf{5.15} &\textbf{10.49} &\textbf{24.19} &\textbf{31.51} &69.45\\ \hline
 
 STBMPS   & 13.50 &24.27 &41.90 &46.62 &53.28 &11.29 &20.73 &41.61 &52.98 &112.02 &6.40 &10.66 &18.28 &21.65 &38.61 &5.09 &10.44 &24.57 &32.16 &69.55\\
 STBMPS-M & \textbf{12.12} &\textbf{21.25} &\textbf{35.04} &\textbf{39.05} &\textbf{52.05} &\textbf{10.59} &\textbf{19.49} &\textbf{39.71} &\textbf{50.54} &\textbf{111.97} &\textbf{6.09} &\textbf{10.15} &\textbf{17.45} &\textbf{20.72} &\textbf{39.42} &\textbf{4.57} &\textbf{9.28} &\textbf{22.35} &\textbf{29.69} &\textbf{68.86}\\
\bottomrule
\end{tabular}
\end{table*}

\end{document}